\documentclass[11pt]{article}

\usepackage[]{acl}
\usepackage{adjustbox}
\usepackage{microtype}
\usepackage{xspace}
\usepackage{mathtools}
\usepackage{amssymb}
\usepackage{rotating}
\usepackage{footnote}
\usepackage{tablefootnote}
\usepackage{graphicx}
\usepackage{color, colortbl}
\usepackage{xstring}
\usepackage{comment}
\usepackage{subcaption}
\usepackage{float}
\restylefloat{table}
\usepackage{array,booktabs,makecell}
\usepackage{appendix}
\usepackage[normalem]{ulem}
\usepackage{enumitem}
\usepackage{scalerel,xparse}
\usepackage{siunitx}
\usepackage{times}
\usepackage{latexsym}
\usepackage{multirow}
\usepackage{amsmath}

\usepackage[T1]{fontenc}

\usepackage[utf8]{inputenc}
\usepackage{arydshln}

\usepackage{caption}
\usepackage{bbm}
\newcolumntype{H}{>{\setbox0=\hbox\bgroup}c<{\egroup}@{}}

\NewDocumentCommand\grinningface{}{
    \scalerel*{
        \includegraphics{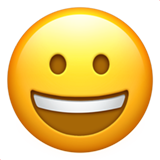}
    }{X}
}
\NewDocumentCommand\grinningsquintingface{}{
    \scalerel*{
        \includegraphics{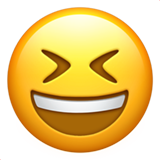}
    }{X}
}
\NewDocumentCommand\loudlycryingface{}{
    \scalerel*{
        \includegraphics{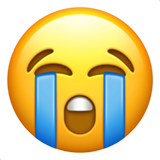}
    }{X}
}
\NewDocumentCommand\grinningfacewithsmilingeyes{}{
    \scalerel*{
        \includegraphics{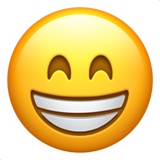}
    }{X}
}
\NewDocumentCommand\smilingfacewithsmilingeyes{}{
    \scalerel*{
        \includegraphics{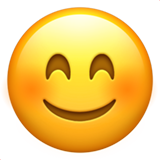}
    }{X}
}
\NewDocumentCommand\grinningfacewithbigeyes{}{
    \scalerel*{
        \includegraphics{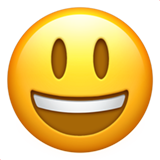}
    }{X}
}
\NewDocumentCommand\thumbsup{}{
    \scalerel*{
        \includegraphics{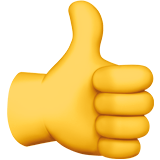}
    }{X}
}
\NewDocumentCommand\confettiball{}{
    \scalerel*{
        \includegraphics{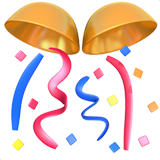}
    }{X}
}
\NewDocumentCommand\partypopper{}{
    \scalerel*{
        \includegraphics{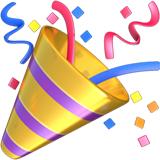}
    }{X}
}
\NewDocumentCommand\squintingfacewithtongue{}{
    \scalerel*{
        \includegraphics{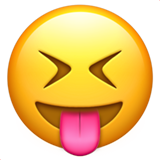}
    }{X}
}
\NewDocumentCommand\winkingfacewithtongue{}{
    \scalerel*{
        \includegraphics{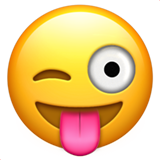}
    }{X}
}
\NewDocumentCommand\wrappedgift{}{
    \scalerel*{
        \includegraphics{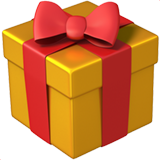}
    }{X}
}
\NewDocumentCommand\grimacingface{}{
    \scalerel*{
        \includegraphics{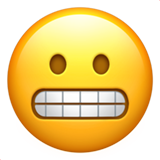}
    }{X}
}
\NewDocumentCommand\facewithtearsofjoy{}{
    \scalerel*{
        \includegraphics{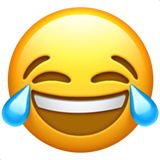}
    }{X}
}
\NewDocumentCommand\grinningfacewithsweat{}{
    \scalerel*{
        \includegraphics{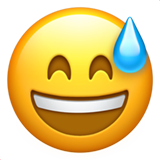}
    }{X}
}
\NewDocumentCommand\relievedface{}{
    \scalerel*{
        \includegraphics{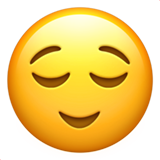}
    }{X}
}
\NewDocumentCommand\flaggermany{}{
    \scalerel*{
        \includegraphics{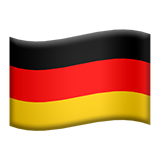}
    }{X}
}
\NewDocumentCommand\flagchina{}{
    \scalerel*{
        \includegraphics{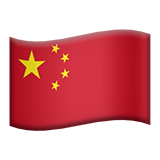}
    }{X}
}
\NewDocumentCommand\flagrussia{}{
    \scalerel*{
        \includegraphics{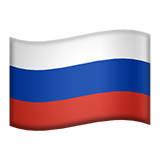}
    }{X}
}
\NewDocumentCommand\flagsouthkorea{}{
    \scalerel*{
        \includegraphics{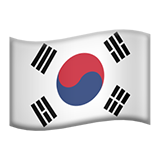}
    }{X}
}
\NewDocumentCommand\flagspain{}{
    \scalerel*{
        \includegraphics{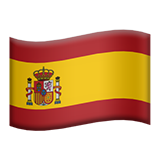}
    }{X}
}
\NewDocumentCommand\flagfrance{}{
    \scalerel*{
        \includegraphics{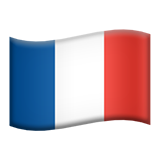}
    }{X}
}
\NewDocumentCommand\flagjapan{}{
    \scalerel*{
        \includegraphics{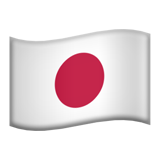}
    }{X}
}
\NewDocumentCommand\flagitaly{}{
    \scalerel*{
        \includegraphics{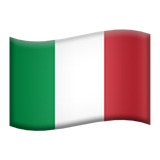}
    }{X}
}
\NewDocumentCommand\flagunitedkingdom{}{
    \scalerel*{
        \includegraphics{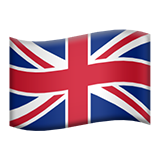}
    }{X}
}
\NewDocumentCommand\flagportugal{}{
    \scalerel*{
        \includegraphics{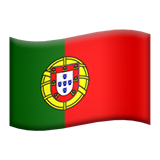}
    }{X}
}
\NewDocumentCommand\flagaustralia{}{
    \scalerel*{
        \includegraphics{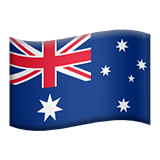}
    }{X}
}
\NewDocumentCommand\flagireland{}{
    \scalerel*{
        \includegraphics{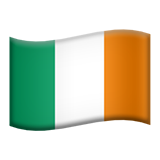}
    }{X}
}
\NewDocumentCommand\shamrock{}{
    \scalerel*{
        \includegraphics{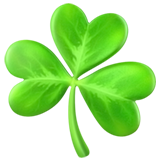}
    }{X}
}
\NewDocumentCommand\flaginhole{}{
    \scalerel*{
        \includegraphics{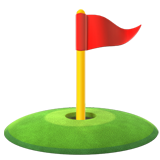}
    }{X}
}
\NewDocumentCommand\icehockey{}{
    \scalerel*{
        \includegraphics{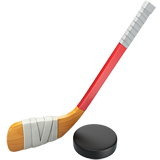}
    }{X}
}
\NewDocumentCommand\flagpakistan{}{
    \scalerel*{
        \includegraphics{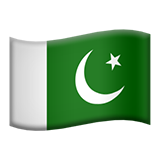}
    }{X}
}
\NewDocumentCommand\flagphilippines{}{
    \scalerel*{
        \includegraphics{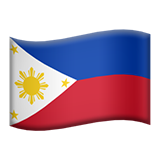}
    }{X}
}
\NewDocumentCommand\cloudwithsnow{}{
    \scalerel*{
        \includegraphics{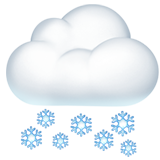}
    }{X}
}
\NewDocumentCommand\snowmanwithoutsnow{}{
    \scalerel*{
        \includegraphics{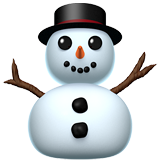}
    }{X}
}
\NewDocumentCommand\windface{}{
    \scalerel*{
        \includegraphics{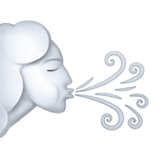}
    }{X}
}
\NewDocumentCommand\snowflake{}{
    \scalerel*{
        \includegraphics{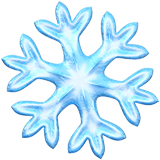}
    }{X}
}
\NewDocumentCommand\sunbehindraincloud{}{
    \scalerel*{
        \includegraphics{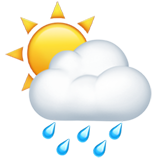}
    }{X}
}
\NewDocumentCommand\cloudwithlightning{}{
    \scalerel*{
        \includegraphics{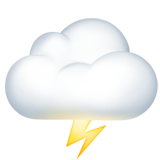}
    }{X}
}
\NewDocumentCommand\fog{}{
    \scalerel*{
        \includegraphics{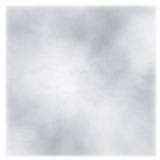}
    }{X}
}
\NewDocumentCommand\cloudwithrain{}{
    \scalerel*{
        \includegraphics{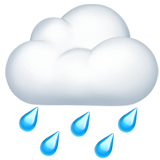}
    }{X}
}
\NewDocumentCommand\sunbehindlargecloud{}{
    \scalerel*{
        \includegraphics{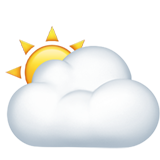}
    }{X}
}
\NewDocumentCommand\cloudwithlightningandrain{}{
    \scalerel*{
        \includegraphics{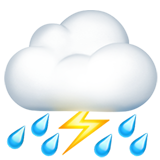}
    }{X}
}
\NewDocumentCommand\tornado{}{
    \scalerel*{
        \includegraphics{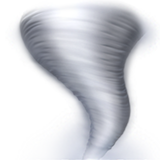}
    }{X}
}
\NewDocumentCommand\foggy{}{
    \scalerel*{
        \includegraphics{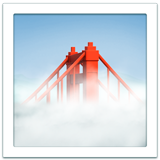}
    }{X}
}
\NewDocumentCommand\helicopter{}{
    \scalerel*{
        \includegraphics{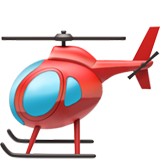}
    }{X}
}
\NewDocumentCommand\closedumbrella{}{
    \scalerel*{
        \includegraphics{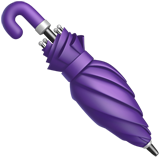}
    }{X}
}
\NewDocumentCommand\mountain{}{
    \scalerel*{
        \includegraphics{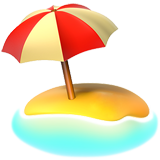}
    }{X}
}
\NewDocumentCommand\bus{}{
    \scalerel*{
        \includegraphics{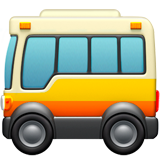}
    }{X}
}

\title{Contrastive Learning of Sociopragmatic Meaning in Social Media}

\author{Chiyu Zhang$^{1}$  ~~~~Muhammad Abdul-Mageed$^{1,2}$ ~~~~Ganesh Jawahar$^{1}$  \\ 
  $^{1}$Deep Learning \& Natural Language Processing Group,
  The University of British Columbia\\\normalsize  $^{2}$Department of Natural Language Processing \& Department of Machine Learning, MBZUAI\\ 
  \tt chiyuzh@mail.ubc.ca, \tt muhammad.mageed@ubc.ca, \\ \tt ganeshjwhr@gmail.com}

\begin{document}
\maketitle
\begin{abstract}

Recent progress in representation and contrastive learning in NLP has not widely considered the class of \textit{sociopragmatic meaning} (i.e., meaning in interaction within different language communities). To bridge this gap, we propose a novel framework for learning task-agnostic representations transferable to a wide range of sociopragmatic tasks (e.g., emotion, hate speech, humor, sarcasm). Our framework outperforms other contrastive learning frameworks for both in-domain and out-of-domain data, across both the general and few-shot settings. For example, compared to two popular pre-trained language models, our model obtains an improvement of $11.66$ average $F_1$ on $16$ datasets when fine-tuned on only $20$ training samples per dataset. We also show that our framework improves uniformity and preserves the semantic structure of representations. Our code is available at: \url{https://github.com/UBC-NLP/infodcl}

\end{abstract}
\section{Introduction}
\label{sec:intro}

Meaning emerging through human interaction such as on social media is deeply contextualized. It extends beyond referential meaning of utterances to involve both information about language users and their identity (the domain of \textit{sociolinguistics}~\cite{tagliamonte2015making}) and the communication goals of these users (the domain of \textit{pragmatics}~\cite{thomas2014meaning}). From a sociolinguistics perspective, a message can be expressed in various linguistic forms, depending on user background. For example, someone might say `let's watch the soccer game', but they can also call the game `football'. In real world, the game is the same thing. While the two expressions are different ways of saying the same thing~\cite{labov1972sociolinguistic}, they do carry information about the user such as their region (i.e., where they could be coming from). From a pragmatics perspective, the meaning of an utterance depends on its interactive context. For example, while the utterance `it's really hot here' (said in a physical meeting) could be a polite way of asking someone to open the window, it could mean `it's not a good idea for you to visit at this time' (said in a phone conversation discussing travel plans). We refer to the meaning communicated through this type of socially embedded interaction as \textit{sociopragmatic meaning} (SM). 

\begin{figure}[t]
\centering
\includegraphics[width=\linewidth]{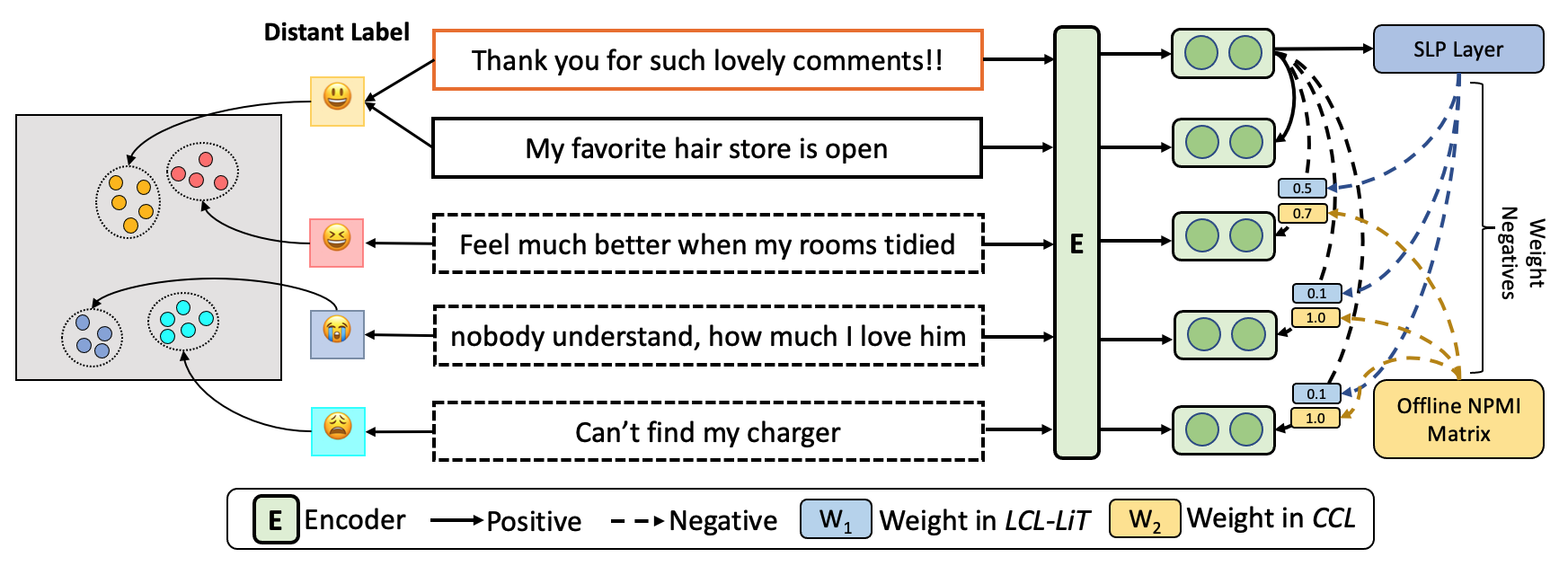} 
\caption{Illustration of our proposed InfoDCL framework. We exploit distant/surrogate labels (i.e., emojis) to supervise two contrastive losses, $\mathcal{L}_{CCL}$ and $\mathcal{L}_{LCL-LiT}$ (see text). Sequence representations from our model should keep the cluster of each class distinguishable and preserve semantic relationships between classes. }\label{fig:illustration}
\end{figure} 

While SM is an established concept in linguistics~\cite{leech1983principles}, NLP work still lags behind. This issue is starting to be acknowledged in the NLP community~\cite{dong-2021-on-learning},  and there has been calls to include social aspects in \textit{representation learning} of language~\cite{bisk2020experience}. Arguably, pre-trained language models (PLMs) such as BERT~\cite{devlin-2019-bert} learn representations relevant to SM tasks. While this is true to some extent, PLMs are usually pre-trained on standard forms of language (e.g. BookCorpus) and hence miss \textbf{(i)} \textit{variation in language use among different language communities} (social aspects of meaning) in \textbf{(ii)} \textit{interactive} settings (pragmatic aspects). In spite of recent efforts to rectify some of these limitations by PLMs such as BERTweet on casual language~\cite{nguyen-etal-2020-bertweet}, it is not clear whether the masked language modeling (MLM) objective employed in PLMs is sufficient for capturing the rich representations needed for sociopragmatics. 

Another common issue with PLMs is that their sequence-level embeddings suffer from the \textit{anisotropy} problem~\cite{ethayarajh-2019-contextual,li-etal-2020-sentence}. That is, these representations tend to occupy a narrow cone on the multidimensional space. This makes it hard for effectively teasing apart sequences belonging to different classes without use of large amounts of labeled data. Work on \textit{contrastive learning} (CL) has targeted this issue of anisotropy by attempting to bring semantic representations of instances of a given class (e.g., positive pairs of the same objects in images or same topics in text) closer and representations of negative class(es) instances farther away~\cite{liu2021fast,gao2021simcse}. A particularly effective type of CL is supervised CL~\cite{khosla2020supervised, khondaker-2022-benchmark}, but it \textbf{(i)} requires labeled data \textbf{(ii)} for each downstream task. Again, acquiring labeled data is expensive and resulting models are task-specific (i.e., cannot be generalized to all SM tasks).    

In this work, our goal is to \textit{learn effective task-agnostic representations for SM from social data without a need for labels}. To achieve this goal, we introduce a novel framework situated in CL that we call \textbf{InfoDCL}. InfoDCL leverages sociopragmatic signals such as emojis or hashtags naturally occurring in social media, treating these as distant/surrogate labels.\footnote{We use distant label and surrogate label interchangeably.} Since surrogate labels are abundant (e.g., hashtags on images or videos), our framework can be extended beyond language. 
To illustrate the superiority of our proposed framework, we evaluate representations by our InfoDCL on $24$ SM datasets (such as emotion recognition~\cite{mohammad-2018-semeval} and irony detection~\cite{ptavcek2014sarcasm}) and compare against $11$ competitive baselines. Our proposed framework outperforms all baselines on $14$ (out of $16$) in-domain datasets and seven (out of eight) out-of-domain datasets (Sec.~\ref{sec:main_results}). Furthermore, our framework is \textit{strikingly successful in few-shot learning}: it consistently outperforms baselines by a large margin for different sizes of training data (Sec.~\ref{sec:main_results}). Our framework is also \textit{language-independent}, as demonstrated on several tasks from three languages other than English (Sec.~\ref{subsec:app_multi-ling}).  

Our major contributions are as follows:
\textbf{(1)} We introduce InfoDCL, a novel CL framework for learning sociopragmatics exploiting surrogate labels. To the best of our knowledge, this is the first work to utilize surrogate labels in language CL to improve PLMs. \textbf{(2)} We propose a new CL loss, Corpus-Aware Contrastive Loss (CCL), to preserve the semantic structure of representations exploiting corpus-level information (Sec.~\ref{subsec:ccl}).  
\textbf{(3)} Our framework outperforms several competitive methods on a wide range of SM tasks (both \textit{in-domain} and \textit{out-of-domain}, across \textit{general} and \textit{few-shot} settings).
\textbf{(4)} Our framework is language-independent, as demonstrated by its utility on various SM tasks in four languages. 
\textbf{(5)} We offer an extensive number of ablation studies that show the contribution of each component in our framework and qualitative analyses that demonstrate superiority of representation from our models (Sec.~\ref{sec:ablation_study_main}). 

\section{Related Work}\label{sec:rel}
Our work combines advances in representation learning and contrastive learning. 

\paragraph{\textbf{Representation Learning.}} PLMs encode discrete language symbols into a continuous representation space that can capture the syntactic and the semantic information underlying the text. 
Since BERT is pre-trained on standard text that is not ideal for social media,~\citet{nguyen-etal-2020-bertweet} propose BERTweet, a model pre-trained on tweets with MLM objective and without intentionally learning SM from social media data. Previous studies~\cite{felbo2017using,corazza-2020-hybrid} have also utilized distant supervision (e.g., use of emoji) to obtain better representations for a limited number of tasks. 
Our work differs in that we make use of distant supervision \textit{in the context of CL} to acquire rich representations \textit{suited to the whole class of SM tasks}. In addition, our methods excel not only in the full data setting but also for \textit{few-shot learning} and diverse domains. 

\paragraph{\textbf{Contrastive Learning.}} There has been a flurry of recent CL frameworks introducing self-supervised~\cite{liu2021fast, gao2021simcse,cao-etal-2022-exploring}, semi-supervised~\cite{yu2021fine}, weakly-supervised~\cite{zheng2021weakly}, and strongly supervised~\cite{gunel2020supervised,suresh2021not,zhou-etal-2022-debiased} learning objectives.\footnote{These frameworks differ across a number of dimensions that we summarize in Table~\ref{fig:related_work} in Sec.~\ref{sec:app_related} in Appendix.} 
Although effective, existing supervised CL (SCL) frameworks~\cite{gunel2020supervised, suresh2021not, pan-2022-improved} suffer from \textbf{two major drawbacks}. The \textbf{first drawback} is SCL's dependence on task-specific labeled data (which is required to identify positive samples in a batch). Recently,~\citet{zheng2021weakly} introduced a weakly-supervised CL (WCL) objective for computer vision, 
which generates a similarity-based \textit{1}-nearest neighbor graph in each batch and assigns weak labels for samples of the batch (thus clustering vertices in the graph). 
It is not clear, however, how much an WCL method with augmentations akin to language would fare for NLP. We propose a framework that does not require model-derived weak labels, which outperforms a clustering-based WCL approach. The \textbf{second drawback} with SCL is related to how negative samples are treated.~\newcite{khosla2020supervised,gunel2020supervised} treat all the negatives equally, which is sub-optimal since hard negatives should be more informative~\cite{robinson2020contrastive}. \newcite{suresh2021not} attempt to rectify this by introducing a \textit{label-aware contrastive loss} (LCL) where they feed the anchor sample to a task-specific model and assign higher weights to confusable negatives based on this model's confidence on the class corresponding to the negative sample. LCL, however, is both \textbf{narrow} and \textbf{costly}. It is narrow since it exploits \textit{task-specific} labels. We fix this by employing surrogate labels generalizable to \textit{all} SM tasks. In addition, LCL is costly since it requires an auxiliary task-specific model to be trained with the main model. Again, we fix this issue by introducing a \textit{light} LCL framework (\textbf{LCL-LiT}) where we use our main model, rather than an auxiliary model, to derive the weight vector $w_{i}$ from our main model through an additional loss (i.e., weighting is performed end-to-end in our main model). Also, LCL \textbf{\textit{only} considers instance-level information} to capture relationships between individual sample and classes. In comparison, we introduce a novel corpus-aware contrastive loss (CCL) that overcomes this limitation (Sec.~\ref{subsec:ccl}). 




\section{Proposed Framework}\label{sec:proposed_method}
Our goal is to learn rich and diverse representations suited for a wide host of SM tasks. 
To this end, we introduce our novel \textbf{InfoDCL} framework. 
InfoDCL is a \textit{distantly supervised} CL (DCL) framework that exploits distant/surrogate label (e.g., emoji) as a proxy for supervision and incorporates corpus-level information to capture inter-class relationships.
\subsection{Contrastive Losses}\label{subsec:self-cl}
CL aims to learn efficient representations by pulling samples from the same class together and pushing samples from other classes apart~\cite{hadsell2006dimensionality}. We formalize the framework now. Let $C$ denote the set of class labels. Let $D = \{(x_i, y_i)\}_{i=1}^{m}$ denote a randomly sampled batch of size $m$, where $x_i$ and $y_i \in C$ denote a sample and its label respectively.
Many CL frameworks construct the similar (\textit{a.k.a}., positive) sample ($x_{m+i}$) for an anchor sample ($x_i$) by applying a data augmentation technique ($\mathcal{T}$) such as back-translation~\cite{fang2020cert}, token masking~\cite{liu2021fast}, and dropout masking~\cite{gao2021simcse} on the anchor sample ($x_i$). Let $B=\{(x_i, y_i)\}_{i=1}^{2m}$ denote an augmented batch, where $x_{m+i} = \mathcal{T}(x_i)$ and $y_{m+i}$ = $y_i$ ($i=\{1,\dots, m\}$).

\paragraph{\textbf{Self-supervised Contrastive Loss.}} We consider $|C| = N$, where $N$ is the total number of training samples. 
Hence, the representation of the anchor sample $x_i$ is pulled closer to that of its augmented (positive) sample $x_{m+i}$ and pushed away from the representations of other $2m-2$ (negative) samples in the batch. The semantic representation $h_i \in \mathbbm{R}^d$ for each sample $x_i$ is computed by an encoder, $\Phi$, where $h_i = \Phi(x_i)$.
\citet{chen2017sampling} calculate the contrastive loss in a batch as follows: 
\begin{equation}
\small
    \mathcal{L}_{SSCL} = \sum_{i=1}^{2m} - \log \frac{e^{sim(h_i, h_{p(i)})/\tau}}{\sum_{a=1}^{2m} \mathbbm{1}_{[a\neq i]} e^{sim(h_i, h_a)/ \tau}},\label{eq:selfcl}
\end{equation}
where $p(i)$ is the index of positive sample of $x_i$,\footnote{If $i\leq m$, $p(i)=i+m$, otherwise $p(i)=i-m$.} $\tau \in R^+$ is a scalar temperature parameter, and $sim(h_i, h_j)$ is the cosine similarity $\frac{h_i^\top h_j}{\|h_i\|\cdot\|h_j\|}$. 

\paragraph{\textbf{Supervised Contrastive Loss.}} The CL loss in Eq.~\ref{eq:selfcl} is unable to handle the case of multiple samples belonging to the same class when utilizing a supervised dataset ($|C| < N$). Positive samples in SCL~\cite{khosla2020supervised} is a set composed of not only the augmented sample but also the samples belonging to the same class as $x_i$. The positive samples of $x_i$ are denoted by $P_i = \{\rho \in B : y_\rho = y_i \land \rho \neq i\}$, and $|P_i|$ is its cardinality. The SCL is formulated as: 
\begin{equation}
\scriptsize
    \mathcal{L}_{SCL} =  \sum_{i=1}^{2m} \frac{-1}{|P_i|} \sum_{p\in P_i} \log \frac{e^{sim(h_i, h_p)/\tau}}{\sum_{a=1}^{2m} \mathbbm{1}_{[a\neq i]} e^{sim(h_i, h_a)/ \tau}}.
\end{equation}

In our novel framework, we make use of SCL but employ surrogate labels instead of gold labels to construct the positive set. 

\subsection{Label-Aware Contrastive Loss}\label{subsec:lcl}
\citet{suresh2021not} extend the SCL to capture relations between negative samples. They hypothesize that not all negatives are equally difficult for an anchor and that the more confusable negatives should be emphasized in the loss. 
They propose LCL, which introduces a weight $w_{i,y_a}$ to indicate the confusability of class label $y_a$ w.r.t. anchor $x_i$:
\begin{equation}
\scriptsize
    \mathcal{L}_{LCL} = \sum_{i=1}^{2m} \frac{-1}{|P_i|} \sum_{p\in P_i} \log \frac{w_{i,y_i} \cdot e^{sim(h_i, h_p)/\tau}}{\sum_{a=1}^{2m} \mathbbm{1}_{[a\neq i]} w_{i,y_a} \cdot e^{sim(h_i, h_a)/ \tau}}.\label{eq:lcl}
\end{equation}

The weight vector $w_{i} \in \mathbbm{R}^{|C|}$ comes from the class-specific probabilities (or confidence score) outputted by an auxiliary task-specific supervised model after consuming the anchor $x_i$. LCL assumes that the highly confusable classes w.r.t anchor receive higher confidence scores, while the lesser confusable classes w.r.t anchor receive lower confidence scores. As stated earlier, limitations of LCL include \textbf{(i)} its dependence on gold annotations, \textbf{(ii)} its inability to generalize to all SM tasks due to its use of task-specific labels, and \textbf{(iii)} its ignoring of corpus-level and inter-class information. As explained in Sec.~\ref{sec:rel}, we fix all these issues. 

\subsection{Corpus-Aware Contrastive Loss}
\label{subsec:ccl} 
In spite of the utility of existing CL methods for text representation, a uniformity-tolerance dilemma has been identified in vision representation model by~\newcite{wang-2021-understanding}: pursuing excessive uniformity makes a model intolerant to semantically similar samples, thereby breaking its underlying semantic structure (and thus causing harm to downstream performance).\footnote{For details see Sec.~\ref{sec:uni-tole} in Appendix.} Our learning objective is to obtain representations suited to all SM tasks, thus we hypothesize that preserving the semantic relationships between surrogate labels during pre-training can benefit many of downstream SM tasks. Since we have a large number of fine-grained classes (i.e., surrogate labels), each class will not be equally distant from all other classes. For example, the class `\grinningface' shares similar semantics with the class `\grinningsquintingface', but is largely distant to the class `\loudlycryingface'. The texts with `\grinningface' and `\grinningsquintingface' belong to same class of `joy' in downstream emotion detection task. We thus propose a new CL method that relies on distant supervision to learn general knowledge of all SM tasks and incorporates corpus-level information to capture inter-class relationships, while improving uniformity of PLM and preserving the underlying semantic structure. 
Concretely, our proposed corpus-aware contrastive loss (CCL) exploits a simple yet effective corpus-level measure based on pointwise mutual information (PMI)~\cite{bouma2009normalized}  to extract relations between surrogate labels (e.g., emojis) from a large amount of unlabeled tweets.\footnote{We experiment with a relatively sophisticated approach that learns class embeddings to capture the inter-class relations in Sec.~\ref{sec:ablation_study_main}, but find it to be sub-optimal.} 
The PMI method is cheap to compute as it requires neither labeled data nor model training: PMI is based only on the co-occurrence of emoji pairs. We hypothesize that PMI scores of emoji pairs could provide globally useful semantic relations between emojis. Our CCL based on PMI 
can be formulated as:
\begin{equation}
\scriptsize
     \mathcal{L}_{CCL} =  \sum_{i=1}^{2m} \frac{-1}{|P_i|}  \sum_{p\in P_i} \log \frac{e^{sim(h_i, h_p)/\tau}}{\sum_{a=1}^{2m} \mathbbm{1}_{[a\neq i]} w_{y_i, y_a}\cdot e^{sim(h_i, h_a)/ \tau}}, \label{eq:iwscl}
\end{equation}
where the weight $w_{y_i, y_a}  = 1 - max(0, npmi(y_i, y_a))$, and $npmi(y_i, y_a) \in [-1, 1]$ is normalized point-wise mutual information~\cite{bouma2009normalized} between $y_a$ and $y_i$.\footnote{Equation for NPMI is in Appendix~\ref{subsec:append_npmi}.} 

\subsection{Overall Objective}\label{subsec:overall}
To steer the encoder to learn representations that recognize corpus-level inter-class relations while distinguishing between classes, we combine our $\mathcal{L}_{CCL}$ and $\mathcal{L}_{LCL}$.\footnote{Note that $\mathcal{L}_{LCL}$ operates over surrogate labels rather than task-specific downstream labels as in~\cite{suresh2021not}, thereby allowing us to learn broad SM representations.}
The resulting loss, which we collectively refer to as \textit{distantly-supervised contrastive loss} $\mathcal{L}_{DCL}$ is given by:
\begin{equation}    
    \mathcal{L}_{DCL} = \gamma \mathcal{L}_{LCL} + (1-\gamma) \mathcal{L}_{CCL},\label{eq:dcl}
\end{equation}
where $\gamma \in [0,1]$ is a hyper-parameter that controls the relative importance of each of the contrastive losses. 
Our results show that a model trained with $\mathcal{L}_{DCL}$ can achieve sizeable improvements over baselines (Table~\ref{tab:main_result}). For a more enhanced representation, our proposed framework also exploits a \textit{surrogate label prediction} (SLP) objective $\mathcal{L}_{SLP}$ where the encoder $\Phi$ is jointly optimized for the emoji prediction task using cross entropy loss. Our employment of an SLP objective now allows us to weight the negatives in $\mathcal{L}_{LCL}$ using classification probabilities from our main model rather than training an additional weighting model, another divergence from~\citet{suresh2021not}. This new LCL framework is our \textbf{LCL-LiT} (for \textit{light} LCL),\footnote{The formula of LCL-LiT is the same as Eq. 3 (i.e., Loss of LCL).} giving us a lighter DCL loss that we call \textbf{DCL-LiT}: 
\begin{equation}
    \mathcal{L}_{DCL-LiT} = \gamma \mathcal{L}_{LCL-LiT} + (1-\gamma) \mathcal{L}_{CCL}.\label{eq:dcl-lit}
\end{equation}

Our sharing strategy where a single model is trained end-to-end on an overall objective incorporating negative class weighting should also improve our model efficiency (e.g., training speed, energy efficiency). Our ablation study in Sec.~\ref{sec:ablation_study_main} confirms that using the main model as the weighing network is effective for overall performance. To mitigate effect of any catastrophic forgetting of token-level knowledge, the proposed framework includes an MLM objective defined by $\mathcal{L}_{MLM}$.\footnote{The Equations of $\mathcal{L}_{SLP}$ and $\mathcal{L}_{MLM}$ are listed in Appendix~\ref{subsec:app_classification} and~\ref{subsec:app_mlm}, respectively.}
The overall objective function of the proposed InfoDCL framework can be given by:
\begin{multline}
\mathcal{L}_{InfoDCL} = \lambda_1 \mathcal{L}_{MLM} + \lambda_2 \mathcal{L}_{SLP} \\ 
    + (1-\lambda_1-\lambda_2) \mathcal{L}_{DCL-LiT}, \label{eq:overall}
\end{multline}
where $\lambda_1$ and $\lambda_2$ are the loss scaling factors. We also employ a mechanism for randomly re-pairing an anchor with a new positive sample at the beginning of each epoch. We describe this epoch-wise repairing in Appendix~\ref{subsec:weakly_pairing}.




\begin{table*}[!ht]
\centering
\setlength\tabcolsep{4pt}
\scriptsize
\begin{tabular}{lllcccccccccrc|cc}
\toprule
\multicolumn{1}{c}{\textbf{}}           & \multicolumn{1}{c}{\textbf{Task}}            & \multicolumn{1}{c}{\textbf{RB}} & \textbf{MLM} & \textbf{E-MLM}                  & \textbf{SLP} & \textbf{Mir-B} & \textbf{Sim-S} & \textbf{Sim-D}                  & \textbf{SCL} & \textbf{LCL} & \textbf{WCL} & \multicolumn{1}{l}{\textbf{DCL}} & \textbf{InfoD-R}                                      & \textbf{BTw}                        & \textbf{InfoD-B}                 \\ \midrule
                                        & Crisis\textsubscript{Oltea} & 95.87                           & 95.81        & 95.91                           & 95.89        & 95.79          & 95.71          & 95.94                           & 95.88        & 95.87        & 95.83        & 95.92                                                    & \multicolumn{1}{c|}{\textbf{96.01}} & 95.76                               & \textbf{95.84} \\
                                        & Emo\textsubscript{Moham}    & 78.76                           & 79.68        & 80.79                           & 81.25        & 78.27          & 77.00          & 81.05                           & 78.79        & 77.66        & 77.65        & 80.54                                                    & \multicolumn{1}{c|}{\textbf{81.34}} & 80.23                               & \textbf{81.96} \\
                                        & Hate\textsubscript{Was}     & 57.01                           & 56.87        & 56.65                           & 57.05        & 57.09          & 56.70          & 57.13                           & 56.94        & 56.96        & 57.19        & 57.14                                                    & \multicolumn{1}{c|}{\textbf{57.30}} & {57.32}                           & \textbf{57.65} \\
                                        & Hate\textsubscript{Dav}     & 76.04                           & 77.55        & \textbf{77.79} & 75.70        & 75.88          & 74.40          & 77.15                           & 77.20        & 75.90        & 76.87        & 76.79                                                    & \multicolumn{1}{c|}{77.29}                           & 76.93                               & \textbf{77.94} \\
                                        & Hate\textsubscript{Bas}     & 47.85                           & 52.56        & 52.33                           & 52.58        & 45.49          & 46.81          & 52.32                           & 48.24        & 48.93        & 50.68        & 52.17                                                    & \multicolumn{1}{c|}{\textbf{52.84}} & {53.62}                           & \textbf{53.95} \\
\multirow{7}{*}{\rotatebox[origin=c]{90}{\textbf{In-Domain}}}     & Humor\textsubscript{Mea}    & 93.28                           & 93.62        & 93.73                           & 93.31        & 93.37          & 91.55          & 93.42                           & 92.82        & 93.00        & 92.45        & \textbf{94.13}                                           & \multicolumn{1}{c|}{93.75}                           & \textbf{94.43}     & 94.04                           \\
                                        & Irony\textsubscript{Hee-A}  & 72.87                           & 74.15        & 75.94                           & 76.89        & 70.62          & 66.40          & 75.36                           & 73.58        & 73.86        & 71.24        & \textbf{77.15}                                           & \multicolumn{1}{c|}{76.31}                           & {77.03}                           & \textbf{78.72} \\
                                        & Irony\textsubscript{Hee-B}  & 53.20                           & 52.87        & 55.85                           & 56.38        & 49.60          & 46.26          & 54.06                           & 50.68        & 53.63        & 52.80        & \textbf{57.48}                                           & \multicolumn{1}{c|}{57.22}                           & 56.73                               & \textbf{59.15} \\
                                        & Offense\textsubscript{Zamp} & 79.93                           & 80.75        & 80.72                           & 80.07        & 78.79          & 77.28          & 80.80                           & 79.96        & 80.75        & 79.48        & 79.94                                                    & \multicolumn{1}{c|}{\textbf{81.21}} & {79.35}                           & \textbf{79.83} \\
                                        & Sarc\textsubscript{Riloff}  & 73.71                           & 74.87        & 77.34                           & 77.97        & 66.60          & 64.41          & \textbf{80.27} & 73.92        & 74.82        & 73.68        & 79.26                                                    & \multicolumn{1}{c|}{78.31}                           & 78.76                               & \textbf{80.52} \\
                                        & Sarc\textsubscript{Ptacek}  & 95.99                           & 95.87        & 96.02                           & 95.89        & 95.62          & 95.27          & 96.07                           & 95.89        & 95.62        & 95.72        & \textbf{96.13}                                           & \multicolumn{1}{c|}{96.10}                           & {96.40}                           & \textbf{96.67} \\
                                        & Sarc\textsubscript{Rajad}   & 85.21                           & 86.19        & 86.38                           & 86.89        & 84.31          & 84.06          & 87.20                           & 85.18        & 84.74        & 85.89        & \textbf{87.45}                                           & \multicolumn{1}{c|}{87.00}                           & 87.13                               & \textbf{87.20} \\
                                        & Sarc\textsubscript{Bam}     & 79.79                           & 80.48        & 80.66                           & 81.08        & 79.02          & 77.58          & 81.40                           & 79.32        & 79.62        & 79.53        & 81.31                                                    & \multicolumn{1}{c|}{\textbf{81.49}} & {81.76}                           & \textbf{83.20} \\
                                        & Senti\textsubscript{Rosen}  & 89.55                           & 89.69        & 90.41                           & 91.03        & 85.87          & 84.54          & 90.64                           & 89.82        & 89.79        & 89.69        & 90.65                                                    & \multicolumn{1}{c|}{\textbf{91.59}} & 89.53                               & \textbf{90.41} \\
                                        & Senti\textsubscript{Thel}   & 71.41                           & 71.31        & 71.50                           & 71.79        & 71.23          & 70.11          & 71.68                           & 70.57        & 70.10        & 71.30        & 71.73                                                    & \multicolumn{1}{c|}{\textbf{71.87}} & {71.64}                           & \textbf{71.98} \\
                                        & Stance\textsubscript{Moham} & 69.44                           & 69.47        & 70.50                           & 69.54        & 66.23          & 64.96          & 70.48                           & 69.14        & 69.55        & 70.33        & 69.74                                                    & \multicolumn{1}{c|}{\textbf{71.13}} & {\textbf{68.33}} & 68.22                           \\ \cdashline{2-16}
                                        & Average                                      & 76.24                           & 76.98        & 77.66                           & 77.71        & 74.61          & 73.32          & 77.81                           & 76.12        & 76.30        & 76.27        & 77.97                            & \multicolumn{1}{c|}{\textbf{78.17}} & 77.81                               & \textbf{78.58} \\ \midrule
                                        & Emotion\textsubscript{Wall} & 66.51                           & 66.02        & 67.89                           & 67.28        & 62.33          & 59.59          & 67.68                           & 66.56        & 67.55        & 63.99        & 68.36                                                    & \multicolumn{1}{c|}{\textbf{68.41}} & {64.48}                           & \textbf{65.61} \\
\multirow{7}{*}{\rotatebox[origin=c]{90}{\textbf{Out-of-Domain}}} & Emotion\textsubscript{Dem}  & 56.59                           & 56.77        & 56.80                           & 56.67        & 57.13          & 56.69          & 55.27                           & 54.14        & 56.82        & 55.61        & \textbf{57.43}                                           & \multicolumn{1}{c|}{57.28}                           & 53.33                               & \textbf{54.99} \\
                                        & Sarc\textsubscript{Walk}    & 67.50                           & 66.16        & 67.42                           & 68.78        & 63.95          & 59.39          & 65.04                           & 66.98        & 66.93        & 65.46        & 67.39                                                    & \multicolumn{1}{c|}{\textbf{68.45}} & {67.27}                           & \textbf{67.30} \\
                                        & Sarc\textsubscript{Ora}     & 76.92                           & 76.34        & 77.10                           & 77.25        & 75.57          & 74.68          & 77.12                           & 76.94        & 75.99        & 76.95        & \textbf{77.76}                                           & \multicolumn{1}{c|}{77.41}                           & \textbf{77.33}     & 76.88                           \\
                                        & Senti-MR                                     & 89.00                           & 89.67        & \textbf{89.97} & 89.58        & 88.66          & 87.81          & 89.09                           & 89.14        & 89.33        & 89.47        & 89.15                                                    & \multicolumn{1}{c|}{89.43}                           & {87.94}                           & \textbf{88.21} \\
                                        & Senti-YT                                     & 90.22                           & 91.33        & 91.22                           & 91.98        & 88.63          & 85.27          & 92.23                           & 90.29        & 89.82        & 91.07        & \textbf{92.26}                                           & \multicolumn{1}{c|}{91.98}                           & 92.25                               & \textbf{92.41} \\
                                        & SST-5                                        & 54.96                           & 55.83        & 56.15                           & 55.94        & 54.18          & 52.84          & 55.09                           & 55.33        & 54.28        & 55.30        & 56.00                                                    & \multicolumn{1}{c|}{\textbf{56.37}} & {55.74}                           & \textbf{55.93} \\
                                        & SST-2                                        & 94.57                           & 94.33        & 94.39                           & 94.51        & 93.97          & 91.49          & 94.29                           & 94.50        & 94.24        & 94.61        & 94.64                                                    & \multicolumn{1}{c|}{\textbf{94.98}} & 93.32                               & \textbf{93.73} \\ \cdashline{2-16}
                                        & Average                                      & 74.53                           & 74.55        & 75.12                           & 75.25        & 73.05          & 70.97          & 74.48                           & 74.24        & 74.37        & 74.06        & 75.37                            & \textbf{75.54}                      & {73.96}                           & \textbf{74.38} \\ \bottomrule
\end{tabular}
\caption{Fine-tuning results on our $24$ SM datasets (average macro-$F_1$ over five runs). \textbf{RB:} Fine-tuning on original pre-trained RoBERTa~\cite{liu2019roberta}; \textbf{MLM:} Further pre-training RoBERTa with MLM objective; \textbf{E-MLM:} Emoji-based MLM~\cite{corazza-2020-hybrid}; \textbf{SLP:} Surrogate label prediction; \textbf{Mir-B:} Mirror-BERT~\cite{liu2021fast}; \textbf{Sim-S:} SimCSE-Self~\cite{gao2021simcse}; \textbf{Sim-D:} (Ours) SimCSE-Distant trained with distantly supervised positive pairs and SSCL loss; \textbf{SCL:} Supervised contrastive loss~\cite{khosla2020supervised}; \textbf{LCL:} label-aware contrastive loss~\cite{suresh2021not}; \textbf{BTw:} BERTweet~\cite{nguyen-etal-2020-bertweet}; \textbf{WCL:} Weakly-supervised contrastive learning~\cite{zheng2021weakly}; \textbf{DCL:} (Ours) Trained with $\mathcal{L}_{DCL}$ only (without MLM and SLP objectives); \textbf{InfoD-R} and \textbf{InfoD-B:} (Ours) Continue training RoBERTa and BERTweet, respectively, with proposed InfoDCL framework.}\label{tab:main_result}
\end{table*}

\subsection{Data for Representation Learning}
We exploit emojis as surrogate labels using an English language dataset with $31$M tweets and a total of $1,067$ unique emojis (\texttt{TweetEmoji-EN}). In addition, we acquire representation learning data for \textbf{(1)} our experiments on three additional languages (i.e., Arabic, Italian, and Spanish) and to \textbf{(2)} investigate of the utility of hashtags as surrogate labels. More about how we develop \texttt{TweetEmoji-EN} and all our other representation learning data is in Appendix~\ref{subsec:app_pretrain_data}.
\subsection{Evaluation Data and Splits}

\paragraph{\textbf{In-Domain Data.}} We collect $16$ \textit{\textbf{English language}} Twitter datasets representing eight different SM tasks. These are (1) crisis awareness, (2) emotion recognition, (3) hateful and offensive language detection, (4) humor identification, (5) irony and sarcasm detection, (6) irony type identification, (7) sentiment analysis, and (8) stance detection. We also evaluate our framework on nine Twitter datasets, three from each of \textit{\textbf{Arabic, Italian, and Spanish}}. More information about our English and multilingual datasets is in Appendix~\ref{subsec:app_eval_data}. 

\paragraph{\textbf{Out-of-Domain Data.}} We also identify eight datasets of SM involving emotion, sarcasm, and sentiment derived from outside the Twitter domain (e.g., data created by psychologists, debate fora, YouTube comments, movie reviews). We provide more information about these datasets in Appendix~\ref{subsec:app_eval_data}.  

\paragraph{\textbf{Data Splits.}} 
For datasets without Dev split, we use $10\%$ of the respective training samples as Dev. For datasets originally used in cross-validation, we randomly split into $80\%$ Train, $10\%$ Dev, and $10\%$ Test. Table~\ref{tab:eval_data} in Appendix~\ref{sec:app_data} describes statistics of our evaluation datasets
\subsection{Implementation and Baselines}\label{subsec:implement}
For experiments on English, we initialize our model with the pre-trained English RoBERTa\textsubscript{Base}.\footnote{For short, we refer to the official released English RoBERTa\textsubscript{Base} as RoBERTa in the rest of the paper.} For multi-lingual experiments (reported in Appendix~\ref{subsec:app_multi-ling}), we use the pre-trained XLM-RoBERTa\textsubscript{Base} model~\cite{xlmr-2020-alexis} as our initial checkpoint. More details about these two models are in Appendix~\ref{subsec:app_implement}. We tune hyper-parameters of our InfoDCL framework based on performance on development sets of downstream tasks, finding our model to be resilient to changes in these as detailed in Appendix~\ref{append:subsec:hyper}. To evaluate on downstream tasks, we fine-tune trained models on each task for \textbf{\textit{five times}} with different random seeds and report the averaged model performance. Our main metric is macro-averaged $F_1$ score. To evaluate the overall ability of a model, we also report an aggregated metric that averages over the $16$ in-domain datasets, eight out-of-domain tasks, and the nine multi-lingual Twitter datasets, respectively. 

\paragraph{\textbf{NPMI Weighting Matrix.}} We randomly sample $150$M tweets from our original $350$M Twitter dataset, each with at least two emojis. We extract all emojis in each tweet and count the frequencies of emojis as well as co-occurrences between emojis. To avoid noisy relatedness from low frequency pairs, we filter out emoji pairs $(y_i, y_a)$ whose co-occurrences are less than $20$ times.  We employ Eq.~\ref{eq:npmi} (Appendix~\ref{subsec:append_npmi}) to calculate NPMI for each emoji pair. 

\paragraph{\textbf{Baselines.}} \textit{We compare our methods to $11$ baselines, as described in Appendix~\ref{sec:baseline}. }



\section{Main Results}\label{sec:main_results}

Table~\ref{tab:main_result} shows our main results.
We refer to our models trained with $\mathcal{L}_{DCL}$ (Eq.~\ref{eq:dcl}) and $\mathcal{L}_{InfoDCL}$ (Eq.~\ref{eq:overall}) in Table~\ref{tab:main_result} as DCL and InfoDCL, respectively. We compare our models to $11$ baselines on the $16$ Twitter (in-domain) datasets and eight out-of-domain datasets. 

\paragraph{\textbf{In-Domain Results.}}
InfoDCL outperforms Baseline (1), i.e., fine-tuning original RoBERTa, on each of the $16$ in-domain datasets, with $1.93$ average $F_1$ improvement. 
InfoDCL also outperforms both the MLM and surrogate label prediction (SLP) methods with $1.19$ and $0.46$ average $F_1$ scores, respectively. Our proposed framework is thus able to learn more effective representations for SM. 
We observe that both Mirror-BERT and SimCSE-Self negatively impact downstream task performance, suggesting that while the excessive uniformity they result in is useful for semantic similarity tasks~\cite{gao2021simcse,liu2021fast}, it hurts downstream SM tasks.\footnote{The analyses in Sections~\ref{sec:ablation_study_main} and~\ref{subsec:append_senteval} illustrate this behavior.}  
We observe that our proposed variant of SimCSE, SimCSE-Distant, achieves sizable improvements over both Mirror-BERT and SimCSE-Self ($3.20$ and $4.49$ average $F_1$, respectively). This further demonstrates effectiveness of our distantly supervised objectives. SimCSE-Distant, however, cannot surpass our proposed InfoDCL framework on average $F_1$ over all the tasks. 
We also note that InfoDCL outperforms SCL, LCL, and WCL with $2.05$, $1.87$, and $1.90$ average $F_1$, respectively. Although our simplified model, i.e., DCL, underperforms InfoDCL with $0.20$ average $F_1$, it outperforms all the baselines. Overall, our proposed models (DCL and InfoDCL) obtain best performance in $14$ out of $16$ tasks, and InfoDCL acquires the best average $F_1$. We further investigate the relation between model performance and emoji presence, finding that our proposed approach not only improves tasks involving high amounts of emoji content (e.g., the test set of Emo\textsubscript{Moham} has $23.43\%$ tweets containing emojis) but also those \textit{without any} emoji content (e.g., Hate\textsubscript{Dav}).~\footnote{Statistics of emoji presence of each downstream task is shown in Table~\ref{tab:eval_data} in Appendix.} 
Compared to the original BERTweet, our InfoDCL-RoBERTa is still better ($0.36$ higher $F_1$). This demonstrates not only effectiveness of our approach as compared to domain-specific models pre-trained simply with MLM, but also its data efficiency: BERTweet is pre-trained with $\sim27\times$ more data ($850$M tweets vs. only $31$M for our model). Moreover, the BERTweet we continue training with our framework obtains an average improvement of $0.77$ $F_1$ (outperforms it on $14$ individual tasks). The results demonstrate that our framework can enhance the domain-specific PLM as well. 
%
\paragraph{
\textbf{Out-of-Domain Results.}} 
InfoDCL achieves an average improvement of $1.01$ $F_1$ ($F_1$ = $75.54$) over the eight out-of-domain datasets compared to Baseline (1) as Table~\ref{tab:main_result} shows. Our DCL and InfoDCL models also surpass all baselines on average, achieving highest on seven out of eight datasets. We notice the degradation of BERTweet when we evaluate on the out-of-domain data. Again, this shows generalizability of our proposed framework for leaning SM. 
\paragraph{\textbf{Significance Tests.}}  We conduct two types of significance test on our results, i.e., the classical paired student’s t-test~\cite{fisher1936design} and Almost Stochastic Order (ASO)~\cite{dror-2019-deep}. The t-test shows that our InfoDCL-RoBERTa significantly ($p<.05$) outperforms $9$ out of $11$ baselines (exceptions are SimCSE-Distant and BERTweet) on the average scores over $16$ in-domain datasets and $10$ baselines (exception is SLP) on the average scores over eight out-of-domain datasets. ASO concludes that InfoDCL-RoBERTa significantly ($p<.01$) outperforms all $11$ baselines on both average scores of in-domain and out-of-domain datasets. InfoDCL-BERTweet also significantly ($p<.05$ by t-test, $p<.01$ by ASO) outperforms BERTweet on the average scores. We report standard deviations of our results and significance tests in Appendix~\ref{sec:app_std_sig}.

\paragraph{\textbf{Additional Results.}}
\textit{\textbf{Comparisons to Individual SoTAs.}} We compare our models on each dataset to the task-specific SoTA model on that dataset, acquiring strong performance on the majority of these as we show in Table~\ref{tab:compare}, Sec.~\ref{subsec:append_sota} in Appendix.
\textit{\textbf{Beyond English.}} We also demonstrate effectiveness and generalizability of our proposed framework on nine SM tasks in three additional languages in Sec.~\ref{subsec:app_multi-ling}. 
\textit{\textbf{Beyond Emojis.}} To show the generalizability of our framework to surrogate labels other than emojis, we train DCL and InfoDCL with \textit{hashtags} and observe comparable gains (Sec.~\ref{sec:hashtag}). 
\textit{\textbf{Beyond Sociopragmatics.}} Although the main objective of our proposed framework is to improve model representation for SM, we also evaluate our models on two topic classification datasets and a sentence evaluation benchmark, SentEval~\cite{conneau-2018-senteval}. This allows us to show both strengths of our framework (i.e., improvements beyond SM) and its limitations (i.e., on textual semantic similarity). More about SentEval is in Appendix~\ref{subsec:app_eval_data}, and results are in Sections~\ref{subsec:append_topic} and~\ref{subsec:append_senteval}. 


\paragraph{\textbf{Few-Shot Learning Results.} }
 Since DCL and InfoDCL exploit an extensive set of cues, allowing them to capture a broad range of nuanced concepts of SM, we hypothesize they will be particularly effective in few-shot learning. We hence fine-tune our DCL, InfoDCL, strongest two baselines, and the original RoBERTa with varying amounts of downstream data.\footnote{Data splits for few-shot experiments are in Appendix~\ref{subsec:app_eval_data}.} As Table~\ref{tab:fewshot_number} shows, for in-domain tasks, with only $20$ and $100$ training samples per task, our InfoDCL-RoBERTa strikingly improves $11.66$ and $17.52$ points over the RoBERTa baseline, respectively. Similarly, InfoDCL-RoBERTa is $13.88$ and $17.39$ over RoBERTa with $20$ and $100$ training samples for out-of-domain tasks. These gains also persist when we compare our framework to all other strong baselines, including as we increase data sample size. Clearly, our proposed framework remarkably alleviates the challenge of labelled data scarcity even under severely few-shot settings.
 \footnote{We offer additional few-shot results in Appendix~\ref{subsec:append_fewshot}.}

\begin{table}[ht]
\centering
\tiny
\begin{tabular}{llrrrr@{}}
\toprule
\multicolumn{1}{r}{} & \multicolumn{1}{r}{\textbf{N}} & \multicolumn{1}{c}{\textbf{20}} & \multicolumn{1}{c}{\textbf{100}} & \multicolumn{1}{c}{\textbf{500}} & \multicolumn{1}{c}{\textbf{1000}} \\ \midrule
\multirow{6}{*}{\rotatebox[origin=c]{90}{\textbf{In-Domain}}}& RoBERTa                        & 35.22                           & 41.92                            & 70.06                            & 72.20                             \\
& BERTweet                            & 39.14                           & 38.23                            & 68.35                            & 73.50                             \\ \cdashline{2-6}
& Ours (SimCSE-Distant)                         & 44.99                           & 54.06                            & 71.56                            & 73.39                             \\
& Ours (DCL)                         & 46.60                           & 58.31                            & 72.00                            & 73.86                             \\
& Ours (InfoDCL-RoBERTa)                       & \textbf{46.88}                           & \textbf{59.44}                            & \textbf{72.72}                            & \textbf{74.47}                             \\
& Ours (InfoDCL-BERTweet)                      & 45.29                           & 52.64                            & 71.31                            & 74.03                             \\ \midrule
\multirow{6}{*}{\vspace{10pt}\rotatebox[origin=c]{90}{\textbf{Out-of-Domain}}} & RoBERTa                       & 27.07                           & 41.12                            & 69.26                            & 71.42                             \\
& BERTweet                            & 30.89                           & 39.40                            & 62.52                            & 68.22                             \\ \cdashline{2-6}
& Ours (SimCSE-Distant)                         & 39.02                           & 53.95                            & 66.85                            & 70.50                             \\ 
& Ours (DCL)                         & \textbf{42.19}                           & 56.62                            & 68.22                            & 71.21                             \\ 
& Ours (InfoDCL-RoBERTa)                       & 40.96                           & \textbf{58.51}                            & \textbf{69.36}                            & \textbf{71.92}                             \\
& Ours (InfoDCL-BERTweet)                      & 38.72                           & 48.87                            & 65.64                            & 69.25                             \\ \bottomrule
\end{tabular}
\caption{Few-shot results in average $F_1$ on downstream tasks with $N=20,100,500,1000$ labelled samples. }\label{tab:fewshot_number}
\end{table}


\section{Ablation Studies and Analyses}
\label{sec:ablation_study_main}
\begin{table}[t]
\centering
\scriptsize
\begin{tabular}{@{}llc@{}}
\toprule
\multicolumn{1}{c}{\textbf{Model}} & \multicolumn{1}{c}{\textbf{Avg $F_1$}} & \multicolumn{1}{c}{\textbf{Diff}} \\ \midrule
InfoDCL                              & 78.17~~~~~~($\pm$0.19)                                      & \multicolumn{1}{c}{-}             \\ \cdashline{1-3} 
wo CCL                             & 77.75\textsuperscript{$\dagger$}\textsuperscript{$\star$}~($\pm$0.18)                                     & -0.42                             \\
wo LCL                             & 78.09\textsuperscript{$\dagger$}~~~~($\pm$0.28)                                      & -0.08                             \\
wo CCL \& LCL                   & 77.98\textsuperscript{$\dagger$}~~~~($\pm$0.19)                                      & -0.19                             \\
wo SLP                             & 76.37\textsuperscript{$\dagger$}\textsuperscript{$\star$}~($\pm$0.35)                                      & -0.80                             \\
wo MLM                             & 77.12~~~~~~($\pm$0.31)                                      & -0.05                             \\
wo SLP \& MLM (Our DCL)                     & 77.97\textsuperscript{$\dagger$}~~~~($\pm$0.24)                                      & -0.20                             \\
wo EpW-RP                         & 78.00\textsuperscript{$\dagger$}~~~~($\pm$0.41)                                      & -0.17                             \\
w additional weighting model                     & 78.16~~~~~~($\pm$0.21)                                      & -0.02                             \\
\cdashline{1-3} 
InfoDCL+Self-Aug     &   77.79\textsuperscript{$\dagger$}\textsuperscript{$\star$}~($\pm$0.27)  &   -0.38        \\ \bottomrule 
\end{tabular}
\caption{Result of ablation studies (average macro-$F_1$ across $16$ in-domain datasets). \textbf{\textsuperscript{$\dagger$}} indicates significant ($p<.01$) deterioration based on ASO test. \textbf{\textsuperscript{$\star$}} indicates significant ($p<.05$) deterioration based on t-test.}\label{tab:ablation_avg} 
\end{table}

\textbf{Ablation Studies.} We investigate effectiveness of each of the ingredients in our proposed framework through ablation studies exploiting \texttt{TweetEmoji-EN} for pre-training.
We evaluate on the $16$ in-domain SM datasets with the same hyper-parameters identified in Sec.~\ref{append:subsec:hyper} and report results over five runs. 
As Table~\ref{tab:ablation_avg} shows, InfoDCL outperforms all other settings, demonstrating the utility of the various components in our model. Results show the SLP objective is the most important ingredient in InfoDCL (with a drop of $0.80$ average $F_1$ when removed). However, when we drop both SLP and MLM objectives, DCL (our second best proposed model) only loses $0.20$ $F_1$ as compared to InfoDCL. Results also show that our proposed CCL is more effective than LCL: CCL is second most important component and results in $0.42$ $F_1$ drop vs. only $0.08$ $F_1$ drop when ablating LCL. Interestingly, when we remove \textit{both} CCL and LCL, the model is relatively less affected (i.e., $0.19$ $F_1$ drop) than when we remove CCL alone. We hypothesize this is the case since CCL and LCL are two somewhat opposing objectives: LCL tries to make individual samples distinguishable across confusable classes, while CCL tries to keep the semantic relations between confusable classes. Overall, our results show the utility of distantly supervised contrastive loss. Although distant labels are intrinsically noisy, our InfoDCL is able to mitigate this noise by using CCL and LCL losses. 
Our epoch-wise re-pairing (EpW-RP) strategy is also valuable, as removing it results in a drop of $0.18$ average $F_1$. We believe EpW-RP helps regularize our model as we dynamically re-pair an anchor with a new positive pair for each training epoch. We also train an additional network to produce the weight vector, $w_i$, in LCL loss as~\citet{suresh2021not} proposed instead of using our own main model to assign this weight vector end-to-end. We observe a slight drop of $0.02$ average $F_1$ with the additional model, showing the superiority of our end-to-end approach (which is less computational costly). We also adapt a simple self-augmentation method introduced by~\citet{liu2021fast} to our distant supervision setting: given an anchor $x_i$, we acquire a positive set $\{x_i, x_{m+i}, x_{2m+i}, x_{3m+i}\}$ where $x_{m+i}$ is a sample with the same emoji as the anchor,  $x_{2m+i}$ is an augmented version (applying dropout and masking) of $x_i$, and $x_{3m+i}$ is an augmented version of $x_{m+i}$. As Table~\ref{tab:ablation_avg} shows, InfoDCL+Self-Aug underperforms InfoDCL ($0.38$ $F_1$ drop). We investigate further issues as to how to handle inter-class relations in our models and answer the following questions: 


\noindent\textbf{Should we cluster or push apart the large number of fine-grained (correlated) classes?}
In previous works, contrastive learning is used to push apart samples from different classes.~\citet{suresh2021not} propose the LCL to penalize samples that is more confusable. In this paper, we hypothesize that we should also incorporate inter-class relations into learning objectives (our CCL). Hence, we introduce the PMI score into SCL to \textbf{\textit{scale down}} the loss of a pair belonging to semantically related classes (emojis) as defined in Section~\ref{subsec:ccl} (which should help cluster our fine-grained classes). Here, we investigate an alternative strategy where we explore using the PMI scores as weights to \textbf{\textit{scale up}} the loss of a pair with related labels (which should keep the fine-grained emoji classes separate). Hence, we set $w_{y_i,y_a} = 1+Sim(y_i,y_a)$ where $Sim(y_i,y_a) = max(0, npmi(y_i, y_a))$. We train RoBERTa on $5$M random samples from the training set of \texttt{TweetEmoji-EN} with the overall loss function in Eq.~\ref{eq:overall}, one time using this new weighting method and another time using the weighting method used in all our reported models so far: $w_{y_i,y_a} = 1-Sim(y_i,y_a)$. Given these two ways to acquire $w_{y_i,y_a}$ in Eq.~\ref{eq:iwscl}, we fine-tune the trained model on the $16$ Twitter tasks. Our results in Table~\ref{tab:weighting} show the penalizing strategy to perform lower than our original clustering strategy reported in all experiments in this paper. We also present their performance on each dataset in Table~\ref{tab:penalty_weighting}.

\begin{table}[h]
\centering
\tiny
\begin{tabular}{@{}rlc@{}}
\toprule
              \multicolumn{1}{c}{\textbf{$w_{y_i,y_a}$}}                       & \multicolumn{1}{c}{\textbf{Method}} & \multicolumn{1}{c}{\textbf{Average}} \\ \midrule
\multirow{2}{*}{\textbf{$1-Sim(y_i,y_a)$}} & PMI                                 & 77.70                        \\
                                     & EC-Emb                             & 77.53                          \\ \midrule
\multirow{2}{*}{\textbf{$1+Sim(y_i,y_a)$}} & PMI                                 & 77.39                           \\
                                     & EC-Emb                             & 77.36                          \\ \bottomrule
\end{tabular}
\caption{Comparing different weighting strategies and methods of measuring inter-class similarity. } \label{tab:weighting}
\end{table}

\begin{table}[h]
\centering
\tiny
\begin{tabular}{@{}lcccc|c@{}}
\toprule
 \multicolumn{1}{c}{$w_{y_i, y_a}$} & \multicolumn{2}{c}{$1-Sim(y_i,y_a)$}                                 & \multicolumn{2}{c}{\textbf{$1+Sim(y_i,y_a)$}}      & \multirow{2}{*}{\textbf{RB}}                           \\ \cmidrule(l){2-3} \cmidrule(l){4-5} 
\multicolumn{1}{c}{\textbf{Method}}                               & \multicolumn{1}{c}{\textbf{PMI}} & \multicolumn{1}{c}{\textbf{CLS-emb}} & \multicolumn{1}{c}{\textbf{PMI}} & \multicolumn{1}{c|}{\textbf{CLS-emb}} \\ \midrule
Crisis\textsubscript{Oltea}       & 95.93                            & 95.93                                & 95.88                            & 95.95  &       95.87                        \\
Emo\textsubscript{Moham}          & 81.03                            & 81.30                                & 81.00                            & 80.43  &     78.76                              \\
Hate\textsubscript{Was}           & 57.26                            & 57.16                                & 57.35                            & 57.26          &         57.01             \\
Hate\textsubscript{Dav}           & 76.07                            & 77.42                                & 76.95                            & 76.59        &       76.04                  \\
Hate\textsubscript{Bas}           & 51.86                            & 50.47                                & 52.04                            & 51.68      &       47.85                   \\
Humor\textsubscript{Mea}       & 93.77                            & 93.66                                & 93.65                            & 93.53     &         93.28                   \\
Irony\textsubscript{Hee-A}        & 75.39                            & 73.95                                & 74.09                            & 74.32     & 72.87                           \\
Irony\textsubscript{Hee-B}        & 57.02                            & 55.50                                & 56.99                            & 55.10     & 53.20                           \\
Offense\textsubscript{Zamp}       & 80.29                            & 80.89                                & 81.08                            & 80.81    & 79.93                            \\
Sarc\textsubscript{Riloff}        & 76.73                            & 75.90                                & 72.45                            & 74.64    & 73.71                            \\
Sarc\textsubscript{Ptacek}        & 96.01                            & 95.98                                & 95.99                            & 95.73      & 95.99                          \\
Sarc\textsubscript{Rajad}         & 86.81                            & 86.28                                & 86.22                            & 86.13      & 85.21                          \\
Sarc\textsubscript{Bam}           & 81.40                            & 81.02                                & 81.18                            & 80.48      & 79.79                          \\
Senti\textsubscript{Rosen}        & 91.30                            & 91.64                                & 91.45                            & 91.95    &   89.55                          \\
Senti\textsubscript{Thel}         & 71.72                            & 71.71                                & 72.02                            & 71.65      &  71.44                          \\
Stance\textsubscript{Moham}       & 70.69                            & 71.60                                & 69.91                            & 71.57       &  69.44                         \\ \cdashline{1-6}
Average                                            & 77.70                            & 77.53                                & 77.39                            & 77.36       & 76.24                         \\ \bottomrule
\end{tabular}
\caption{Comparing different weighting strategies and methods of measuring inter-class similarity. \textbf{RB:} Fine-tuning the original RoBERTa, Baseline (1).} \label{tab:penalty_weighting}
\end{table}

\begin{figure}[h]
\centering
\begin{subfigure}[]{.24\textwidth}
  \centering
  \includegraphics[width=\linewidth]{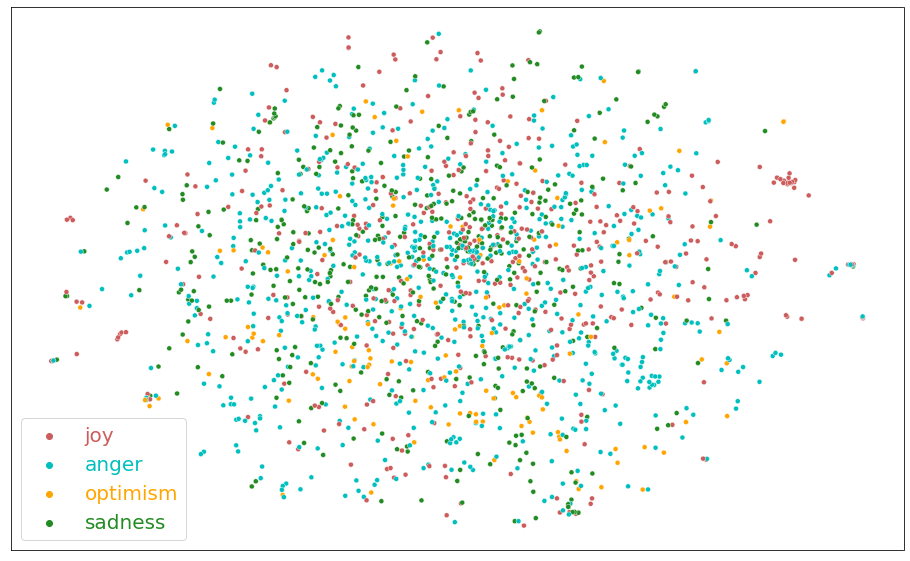}
  \caption{BERTweet on Emo\textsubscript{Moham}}
  \label{fig:sub1}
\end{subfigure}%
\begin{subfigure}[]{.24\textwidth}
  \centering
  \includegraphics[width=\linewidth]{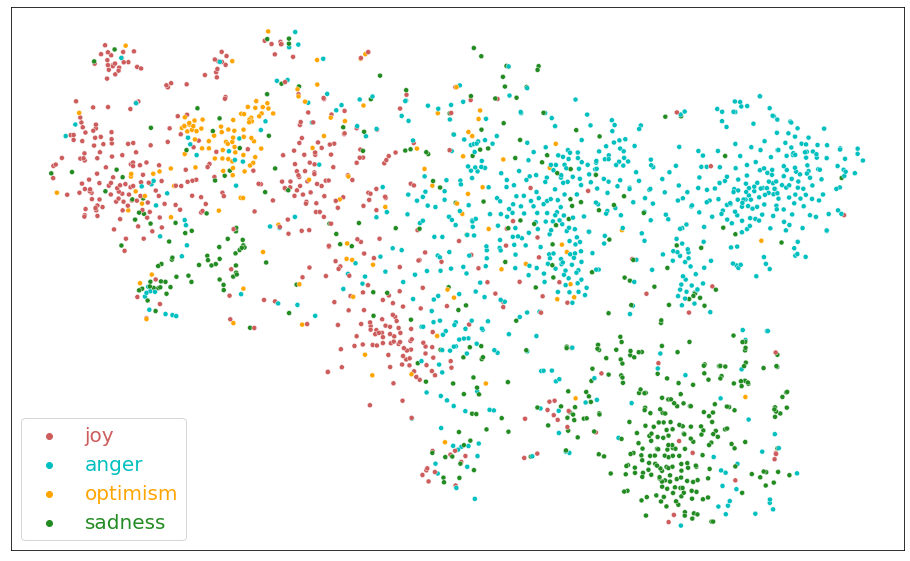}
  \caption{InfoDCL-B on Emo\textsubscript{Moham}}
\end{subfigure}
\begin{subfigure}[]{.24\textwidth}
  \centering
  \includegraphics[width=\linewidth]{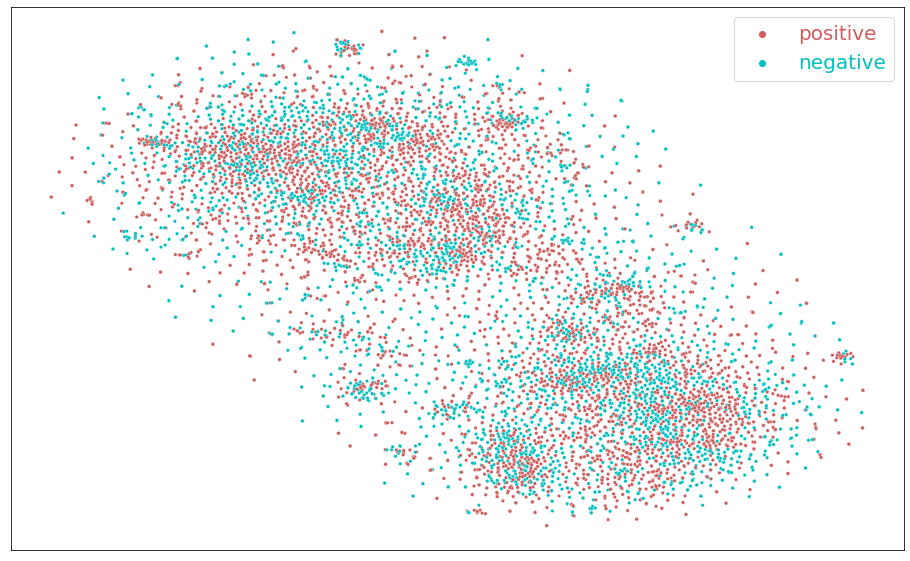}
  \caption{RoBERTa on SST-2}
\end{subfigure}%
\begin{subfigure}[]{.24\textwidth}
  \centering
  \includegraphics[width=\linewidth]{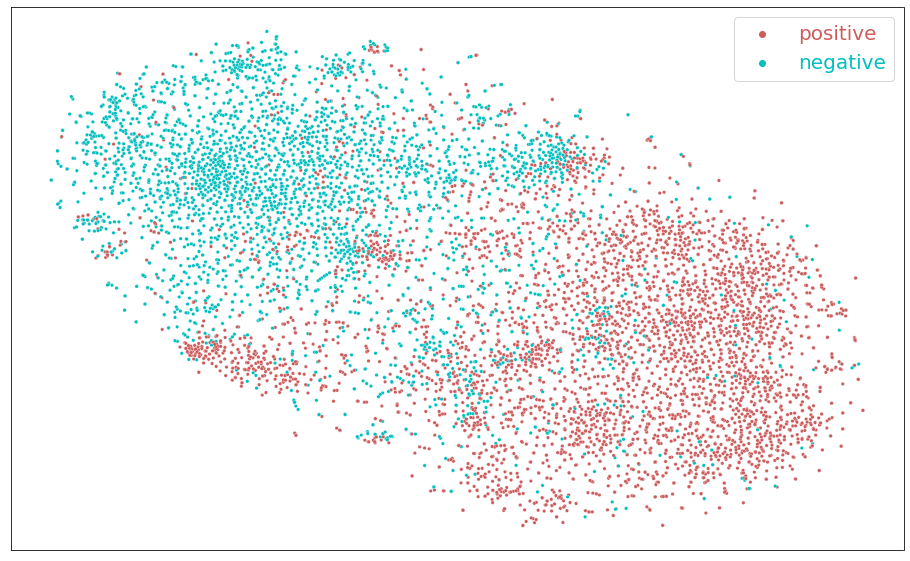}
  \caption{InfoDCL-R on SST-2}
\end{subfigure} 
\caption{$t$-SNE plots of the learned embeddings on Dev and Test sets of two downstream datasets. \textbf{InfoDCL-B:} InfoDCL-BERTweet, \textbf{InfoDCL-R:} InfoDCL-RoBERTa.}
\label{fig:tsne_task}
\end{figure}

\noindent\textbf{Can we use the emoji class embedding (EC-Emb) for corpus-level weighting?}
We experiment with using the embedding of the emoji class (EC-Emb) as an alternative weighting method in place of PMI. Namely, we train RoBERTa on SLP (using the training set of \texttt{TweetEmoji-EN}) for three epochs with a standard cross-entropy loss. We then extract weights of the last classification layer and use these weights as class embeddings, $E=\{e_1, e_2, \dots, e_{C}\}$, where $e_i={R}^{d}$, $d$ is hidden dimension (i.e., $768$), and $|C|$ is the size of classes (i.e., $1,067$). The correlation of each pair of emojis is computed using cosine similarity, i.e., $Sim(y_i, y_a) = \frac{e_i^\top e_a}{\|e_i\|\cdot\|e_a\|}$.\footnote{Self-similarity is set to $0$.} As Table~\ref{tab:weighting} and \ref{tab:penalty_weighting} shows, using PMI scores performs slightly better than using class embeddings in both the clustering and penalizing strategies mentioned previously in the current section. 
For more intuition, we hand-pick three query emojis and manually compare the quality of similarity measures produced by both PMI and class embeddings for these. As Table~\ref{tab:emoji_sim} in Appendix shows, both PMI and EC-Emb are capable of capturing sensible correlations between emojis (although the embedding approach includes a few semantically distant emojis, such as the emoji `\loudlycryingface' being highly related to `\grinningfacewithsmilingeyes'). 
\paragraph{\textbf{Qualitative Analysis.}}
To further illustrate the effectiveness of the representation learned by InfoDCL, we compare a $t$-SNE~\cite{van2008visualizing} visualization of it to that of two strong baselines on two SM datasets.\footnote{Note that we use our model representations \textit{without} downstream fine-tuning.} Fig.~\ref{fig:tsne_task} shows that our model has clearly learned to cluster the samples with similar semantics and separate semantically different clusters before fine-tuning on the gold downstream samples, for both in-domain and out-of-domain tasks. We provide more details about how we obtain the $t$-SNE vitalization and provide another visualization study in Appendix~\ref{subsec:app_qualitative}. 

\begin{figure}[!ht]
\centering
\includegraphics[width=0.93\linewidth]{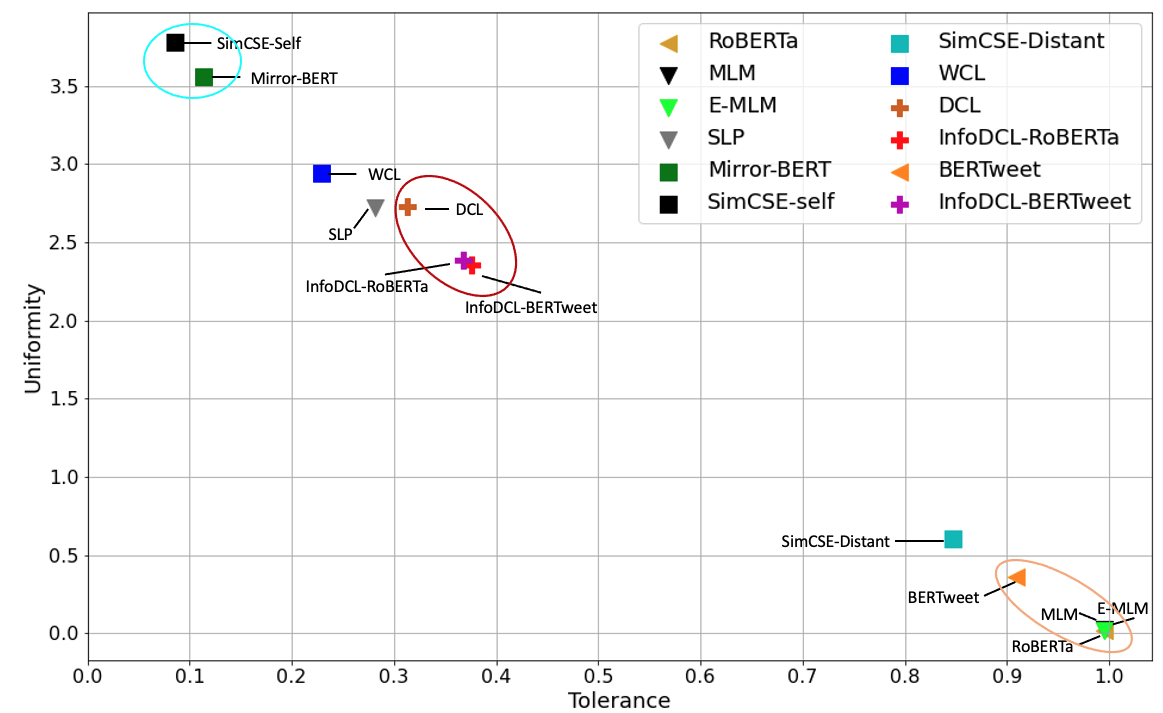} 
\caption{Uniformity and tolerance (higher is better).}\label{fig:data_uni-tol}
\end{figure} 
\noindent\textbf{Uniformity-Tolerance Dilemma.} Following \citet{wang-2021-understanding}, we investigate uniformity and tolerance of our models using Dev data of downstream tasks.\footnote{For details see Sec.~\ref{sec:uni-tole} in Appendix.} As Fig.~\ref{fig:data_uni-tol} shows, unlike other models, our proposed DCL and InfoDCL models make a balance between uniformity and tolerance (which works best for SM).





\section{Conclusion}
We proposed InfoDCL, a novel framework for adapting PLMs to SM exploiting surrogate labels in contrastive learning. We demonstrated effectiveness of our framework on $16$ in-domain and eight out-of-domain datasets as well as nine non-English datasets. Our model outperforms $11$ strong baselines and exhibits strikingly powerful performance in few-shot learning. 
\section{Limitations}
We identify the potential limitations of our work as follow: (1) Distant labels may not be available in every application domain (e.g., patient notes in clinical application), although domain adaptation can be applied in these scenarios. We also believe that distantly supervised contrastive learning can be exploited in tasks involving image and video where surrogate labels are abundant. (2) We also acknowledge that the offline NPMI matrix of our proposed CCL method depends on a dataset (distantly) labeled with multiple classes in each sample. To alleviate this limitation, we explore an alternative method that uses learned class embeddings to calculate the inter-class relations in Section~\ref{sec:ablation_study_main}. This weighting approach achieves sizable improvement over RoBERTa on $16$ in-domain datasets, though it underperforms our NPMI-based approach. (3) Our framework does not always work on tasks outside SM. For example, our model underperforms self-supervised CL models, i.e., SimCSE-Self and Mirror-BERT, on semantic textual similarity task in Appendix~\ref{subsec:append_senteval}. As we showed, however, our framework exhibits promising performance on some other tasks. For example, our hashtag-based model acquires best performance on the topic classification task, as shown in Appendix~\ref{subsec:append_topic}. 


\section*{Ethical Considerations}\label{sec:ethic}
All our evaluation datasets are collected from publicly available sources. Following privacy protection policy, all the data we used for model pre-training and fine-tuning are anonymized. Some annotations in the downstream data (e.g., for hate speech tasks) can carry annotator bias. We will accompany our data and model release with model cards. We will also provide more detailed ethical considerations on a dedicated GitHub repository. All our models will be distributed for research with a clear purpose justification.  

\section*{Acknowledgements}\label{sec:acknow}
We gratefully acknowledge support from the Natural Sciences and Engineering Research Council of Canada (NSERC; RGPIN-2018-04267), the Social Sciences and Humanities Research Council of Canada (SSHRC; 435-2018-0576; 895-2020-1004; 895-2021-1008), Canadian Foundation for Innovation (CFI; 37771), Digital Research Alliance of Canada (the Alliance),\footnote{\href{https://ccdb.alliancecan.ca}{https://www.computecanada.ca}} and UBC ARC-Sockeye.\footnote{\href{https://arc.ubc.ca/ubc-arc-sockeye}{https://arc.ubc.ca/ubc-arc-sockeye}} Any opinions, conclusions or recommendations expressed in this material are those of the author(s) and do not necessarily reflect the views of NSERC, SSHRC, CFI, the Alliance, or UBC ARC-Sockeye.

\bibliography{custom}
\bibliographystyle{acl_natbib}

\newpage 
\appendix
\noindent\textbf{\huge Appendices}
\section{Survey of Contrastive Learning Frameworks.}\label{sec:app_related} 
There has been a flurry of recent contrastive learning frameworks introducing self-supervised, semi-supervised, weakly-supervised, and strongly supervised learning objectives. These frameworks differ across a number of key dimensions: \textbf{(i)} \textit{type of the object} (e.g., image, sentence, document), \textbf{(ii)} \textit{positive example} creation method (e.g., same class as anchor, anchor with few words replaced with synonyms), \textbf{(iii)} \textit{negative example} creation method (e.g., random sample, anchor with few words replaced with antonyms), \textbf{(iv)} \textit{supervision} level (e.g., self, semi, weakly, hybrid, strong), and \textbf{(v)} \textit{weighing of negative samples} (e.g., equal, confidence-based). Table~\ref{fig:related_work} provides a summary of previous frameworks, comparing them with our proposed framework. 
\begin{table*}[!ht]
\centering
\tiny
\begin{tabular}{@{}p{0.75in}p{0.38in}p{1.9in}p{0.9in}p{0.45in}p{0.7in}@{}}
\toprule
\multicolumn{1}{c}{\textbf{Reference}} & \multicolumn{1}{c}{\textbf{Object Type}} & \multicolumn{1}{c}{\textbf{Positive Sample}} & \multicolumn{1}{c}{\textbf{Neg. Sample}} & \multicolumn{1}{c}{\textbf{Supervision}}  & \multicolumn{1}{c}{\textbf{Neg. Weighting}} \\ \midrule
\newcite{khosla2020supervised} & Image & Same class as anchor & Random sample & Strong & Equal \\ 
\newcite{giorgi2020declutr} & Textual span & Span that overlaps with, adjacent to, or subsumed by anchor span & Random span & Self & Equal \\
\newcite{gunel2020supervised} & Document & Same class as anchor & Random sample & Strong & Equal \\
\newcite{zhang2021few} & Utterance & Few tokens masked from anchor / Same class as anchor & Random sample & Self / Strong & Equal \\
\newcite{gao2021simcse} & Sentence & Anchor with different hidden dropout / Sentence entails with anchor & Random sample / Sentence contradicts with anchor & Self / Strong & Equal \\
\newcite{wang2021cline} & Sentence & Anchor with few words replaced with synonyms, hypernyms and morphological changes & Anchor with few words replaced with antonyms and random words & Self & Equal  \\
\newcite{yu2021fine} & Sentence & Same class as anchor & Different class as anchor & Semi- & Equal \\
\newcite{zheng2021weakly} & Image & Same class as anchor & Different class as anchor & Weak & Equal \\
\newcite{zhang-2021-pairwise} & Sentence & Sentence entails with anchor & Sentence contradicts with anchor \& Random sample  & Strong & Similarity \\
\newcite{suresh2021not} & Sentence & Anchor with few words replaced with synonyms / Same class as anchor & Random sample & Self / Strong & Confidence \\ 
\newcite{cocolm_neurips21} & Textual span & Randomly cropped contiguous span & Random sample & Self & Equal \\
\newcite{zhou-etal-2022-debiased} & Sentence & Anchor with different hidden dropout & Random samples and Gaussian noise based samples & Self / Strong & Semantic similarity   \\
\newcite{cao-etal-2022-exploring} & Sentence & Anchor with different hidden dropout  and fast gradient sign method & Random sample & Self & Equal \\ \cdashline{1-6}
\textbf{Ours} & Sentence & Same class as anchor & Random sample & Distant & Confidence \& PMI \\ 
 \bottomrule
\end{tabular}\vspace{-5pt}
\caption{Summary of key differences in existing and proposed contrastive learning frameworks.}
\label{fig:related_work}
\end{table*}
\section{Method}\label{sec:append_method}

\subsection{Normalized Point-Wise Mutual Information}\label{subsec:append_npmi}
The normalized point-wise mutual information (NPMI)~\cite{bouma2009normalized} between $y_a$ and $y_i$.  $npmi(y_i, y_a) \in [-1, 1]$ is formulated as:
\begin{equation}
\small
npmi(y_i, y_a) = \left(\log\frac{p(y_i, y_a)}{p(y_i)p(y_a)} \right) / - \log p(y_i, y_a). \label{eq:npmi} 
\end{equation}

When $npmi(y_i, y_a)=1$, $y_a$ and $y_i$ only occur together and are expected to express highly similar semantic meanings. When $npmi(y_i, y_a)=0$, $y_a$ and $y_i$ never occur together and are expected to express highly dissimilar (i.e., different) semantic meanings. 
We only utilize NPMI scores of related class pairs, i.e., $npmi(y_i, y_a) > 0$. As the NPMI score of $y_a$ and $y_i$ is higher, the weight $w_{y_i, y_a}$ is lower. As a result of incorporating NPMI scores into the negative comparison in the SCL, we anticipate that the representation model would learn better inter-class correlations and cluster the related fine-grained classes. 

\subsection{Surrogate Label Predication}\label{subsec:app_classification}
Our proposed framework also exploits a surrogate label prediction (SLP) objective, where the encoder $\Phi$ is optimized for the surrogate label prediction task using cross entropy.  
Specifically, we pass the hidden representation $h_i$ through two feed-forward layers with $Tanh$ non-linearity in between and obtain the prediction $\hat{y}_{i}$. Then, the surrogate classification loss based on cross entropy can be formalized as: 
\begin{equation}
    \mathcal{L}_{SLP} = -\frac{1}{2m} \sum_{i=1}^{2m} \sum_{c=1}^{C}
    y_{i,c} \cdot \log \hat{y}_{i,c}, \label{eq:ce}
\end{equation}
where $\hat{y}_{i,c}$ is the predicted probability of sample $x_i$ w.r.t class $c$. 

\subsection{Masked Language Modeling Objective}\label{subsec:app_mlm}
Our proposed framework also exploits a MLM objective to mitigate the effect of catastrophic forgetting of the token-level knowledge.
Following~\citet{liu2019roberta}, we randomly corrupt an input sentence by replacing $15\%$ of its tokens with `[MASK]' tokens. Given the corrupted input sequence, we then train our model to predict original tokens at masked positions. Formally, given an input sequence, $x_i=\{t_1,\dots,t_n\}$, the loss function of MLM is formulated as: 
        \begin{equation}
        \mathcal{L}_{MLM} = -\frac{1}{2m} \sum_{i=1}^{2m} \sum_{t_j\in mk(x_i)} \log (p(t_j|t_{cor(x_i)})), \label{eq:mlm_loss}
        \end{equation}
where $mk(x_i)$ indicates the set of masked tokens of the input sequence $x_i$ and $cor(x_i)$ denotes the corrupted input sequence $x_i$.

\subsection{Epoch-Wise Re-Pairing}\label{subsec:weakly_pairing}
Rather than augmenting a batch $D$ with using some data augmentation technique, in our framework, the positive sample $x_{m+i}$ of the anchor $x_i$ is a sample that uses the same emoji. To alleviate any potential noise in our distant labels, 
we introduce an epoch-wise re-pairing (EpW-RP) mechanism where the pairing of a positive sample with a given anchor is not fixed for epochs: at the beginning of each epoch, we flexibly re-pair the anchor with a new positive pair $x_{m+i}$ randomly re-sampled from the whole training dataset using the same emoji as $x_i$. This ensures that each anchor in a given batch will have at least one positive sample.\footnote{Note that each sample in the training dataset is used only once at each epoch, either as the anchor or as a positive sample of the anchor.} 

\section{Data}\label{sec:app_data}
\subsection{Representation Learning Data and Pre-Processing.}\label{subsec:app_pretrain_data}

\paragraph{\textbf{Emoji Pre-Training Dataset.}} We normalize tweets by converting user mentions and hyperlinks to `USER' and `URL', respectively. We keep all the tweets, retweets, and replies but remove the `RT USER:' string in front of retweets. We filter out short tweets ($<5$ actual English word without counting the special tokens such as hashtag, emoji, USER, URL, and RT) to ensure each tweet contains sufficient context. Following previous works~\cite{felbo2017using,barbieri2018semeval,bamman2015contextualized}, we only keep the tweet that contains only a unique type of emoji (regardless of the number of emojis) and that uses a emoji at the end of the tweet. We then extract the emoji as a label of the tweet and remove the emoji from the tweet. We exclude emojis occurring less than $200$ times, which gives us a set of $1,067$ emojis in $32$M tweets. Moreover, we remove few tweets overlapped with Dev and Test sets of our evaluation tasks by Twitter ID and string matching. We refer to this dataset as \texttt{TweetEmoji-EN} and split it into a training ($31$M) and validation ($1$M) set.

\paragraph{\textbf{Hashtag Pre-Training Dataset.}} We also explore using hashtags as surrogate labels for InfoDCL training. Following our data pre-processing procedure on \texttt{TweetEmoji-EN}, we randomly extract $300$M English tweets each with at least one hashtags from a larger in-house dataset collected between $2014$ and $2020$. We only keep tweets that contain a single hashtag used at the end. We then extract the hashtag as a distant label and remove it from the tweet. We exclude hashtags occurring less than $200$ times, which gives us a set of $12,602$ hashtags in $13$M tweets. We refer to this dataset as \texttt{TweetHashtag-EN} and split the tweets into a training set ($12$M) and a validation ($1$M) set.

\paragraph{\textbf{Multilingual Emoji Pre-Training Dataset.}} We collect a multilingual dataset to train multilingual models with our proposed framework. We apply the same data pre-processing and filtering conditions used on English data, and only include tweets that use the $1,067$ emojis in TweetEmoji-EN. We obtain $1$M tweets from our in-house dataset for three languages, i.e., Arabic, Italian, and Spanish.\footnote{However, we were only able to obtain $500$K Italian tweets satisfying our conditions.} We refer to these datasets as \texttt{TweetEmoji-AR}, \texttt{TweetEmoji-IT}, and \texttt{TweetEmoji-ES}. We also randomly extract $1$M English tweets from our \texttt{TweetEmoji-EN} and refer to is as\texttt{ TweetEmoji-EN-1M}. We then combine these four datasets and call the combined dataset \texttt{TweetEmoji-Multi}.
\begin{table*}[!ht]
\centering
\scriptsize
\begin{tabular}{@{}llcccrr@{}}
\toprule
\multicolumn{1}{c}{\textbf{Task}}                                                  & \multicolumn{1}{c}{\textbf{Study}}                            & \textbf{Cls} & \textbf{Domain} & \textbf{Lang} & \multicolumn{1}{c}{\textbf{\begin{tabular}[c]{@{}c@{}}Data Split \\ (Train/Dev/Test)\end{tabular}}} & \multicolumn{1}{c}{\textbf{\begin{tabular}[c]{@{}c@{}}\% of Emoji Samples \\ (Train/Dev/Test)\end{tabular}}} \\ \midrule
Crisis\textsubscript{Oltea}                              & \citet{olteanu2014crisislex}        & 2                             & Twitter                          & EN                             & 48,065/6,008/6,009                             & 0.01/0.02/0.00                                 \\
Emo\textsubscript{Moham}                                 & \citet{mohammad-2018-semeval}       & 4                             & Twitter                          & EN                             & 3,257/374/1,422                                & 11.39/27.81/23.43                              \\
Hate\textsubscript{Was}                                  & \citet{waseem-2016-hateful}         & 3                             & Twitter                          & EN                             & 8,683/1,086/1,086                              & 2.23/2.03/2.76                                 \\
Hate\textsubscript{Dav}                                  & \citet{davidson-2017-hateoffensive} & 3                             & Twitter                          & EN                             & 19,826/2,478/2,479                             & 0.00/0.00/0.00                                 \\
Hate\textsubscript{Bas}                                  & \citet{basile-2019-semeval}         & 2                             & Twitter                          & EN                             & 9,000/1,000/3,000                                 & 6.50/1.50/11.57                                \\
Humor\textsubscript{Mea}                                 & \citet{meaney2021hahackathon}       & 2                             & Twitter                          & EN                             & 8,000/1,000/1,000                              & 0.55/0.00/1.00                                 \\
Irony\textsubscript{Hee-A}                               & \citet{van-hee2018semeval}          & 2                             & Twitter                          & EN                             & 3,450/384/784                                  & 10.58/10.94/11.22                              \\
Irony\textsubscript{Hee-B}                               & \citet{van-hee2018semeval}          & 4                             & Twitter                          & EN                             & 3,450/384/784                                  & 10.58/10.94/11.22                              \\
Offense\textsubscript{Zamp}                              & \citet{zampieri-2019-predicting}    & 2                             & Twitter                          & EN                             & 11,916/1,324/860                               & 11.43/10.88/13.37                              \\
Sarc\textsubscript{Riloff}                               & \citet{riloff2013sarcasm}           & 2                             & Twitter                          & EN                             & 1,413/177/177                                  & 5.38/3.39/4.52                                 \\
Sarc\textsubscript{Ptacek}                               & \citet{ptavcek2014sarcasm}          & 2                             & Twitter                          & EN                             & 71,433/8,929/8,930                             & 4.34/4.36/4.92                                 \\
Sarc\textsubscript{Rajad}                                & \citet{rajadesingan2015sarcasm}     & 2                             & Twitter                          & EN                             & 41,261/5,158/5,158                             & 16.94/18.01/17.10                              \\
Sarc\textsubscript{Bam}                                  & \citet{bamman2015contextualized}    & 2                             & Twitter                          & EN                             & 11,864/1,483/1,484                             & 8.47/8.29/9.64                                 \\
Senti\textsubscript{Rosen}                               & \citet{rosenthal-2017-semeval}      & 3                             & Twitter                          & EN                             & 42,756/4,752/12,284                            & 0.00/0.00/6.59                                 \\
Senti\textsubscript{Thel}                                & \citet{thelwall2012sentiment}       & 2                             & Twitter                          & EN                             & 900/100/1,113                                  & 0.00/0.00/0.00                                 \\
Stance\textsubscript{Moham}                              & \citet{mohammad-2016-semeval}       & 3                             & Twitter                          & EN                             & 2,622/292/1,249                                & 0.00/0.00/0.00                                 \\ \midrule
Emo\textsubscript{Wall} & \citet{wallbott1986universal}       & 7                             & Questionnaire                    & EN                             & 900/100/6,481                                  & 0.00/0.00/0.00                                 \\
Emo\textsubscript{Dem}                                   & \citet{demszky2020goemotion}        & 27                            & Reddit                           & EN                             & 23,486/2,957/2,985                             & 0.00/0.00/0.00                                 \\
Sarc\textsubscript{Walk}                                 & \citet{walker2012corpus}            & 2                             & Debate Forums                    & EN                             & 900/100/995                                    & 0.00/0.00/0.00                                 \\
Sarc\textsubscript{Ora}                                  & \citet{oraby2016creating}           & 2                             & Debate Forums                    & EN                             & 900/100/2,260                                  & 0.00/0.00/0.10                                 \\
Senti-MR                                                                  & \citet{pang2005seeing}              & 2                             & Moview reviews                   & EN                             & 8,529/1,066/1,067                              & 2.01/1.76/1.84                                 \\
Senti-YT                                                                  & \citet{thelwall2012sentiment}       & 2                             & Video comments                   & EN                             & 900/100/1,142                                  & 0.00/0.00/0.00                                 \\
SST-5                                                                     & \citet{socher-2013-recursive}       & 5                             & Moview reviews                   & EN                             & 8,544/1,100/2,209                              & 0.00/0.00/0.00                                 \\
SST-2                                                                     & \citet{socher-2013-recursive}       & 2                             & Moview reviews                   & EN                             & 6,919/871/1,820                                & 0.00/0.00/0.00                                 \\
\midrule
Emo\textsubscript{Mag}   & \citet{mageed-2020-aranet}          & 8                             & Twitter                          & AR                             & 189,902/910/941                                & 16.58/25.27/25.40                              \\
Emo\textsubscript{Bian}                                  & \citet{bianchi2021feel}             & 4                             & Twitter                          & IT                             & 1,629/204/204                                  & 27.62/28.43/32.84                              \\
Emo-es\textsubscript{Moham}                              & \citet{mohammad-2018-semeval}       & 4                             & Twitter                          & ES                             & 4,541/793/2,616                                & 23.67/21.94/22.71                              \\
Hate\textsubscript{Bos}                                  & \citet{bosco2018overview}           & 2                             & Twitter                          & IT                             & 2,700/300/1,000                                & 1.93/1.67/1.50                                 \\
Hate-es\textsubscript{Bas}                               & \citet{basile-2019-semeval}         & 2                             & Twitter                          & ES                             & 4,500/500/1,600                                & 11.07/10.00/7.63                               \\
Irony\textsubscript{Ghan}                                & \citet{idat2019ghanem}              & 2                             & Twitter                          & AR                             & 3,621/403/805                                  & 8.62/9.68/7.95                                 \\
Irony\textsubscript{Cig}                                 & \citet{cignarella2018overview}      & 2                             & Twitter                          & IT                             & 3,579/398/872                                  & 1.68/2.01/5.50                                 \\
Irony\textsubscript{Ort}                                 & \citet{ortega2019overview}          & 2                             & Twitter                          & ES                             & 2,160/240/600                                  & 11.94/15.00/10.00                              \\
Offense\textsubscript{Mub}                               & \citet{mubarak2020overview}         & 2                             & Twitter                          & AR                             & 6,839/1,000/2,000                              & 38.79/36.50/38.75                              \\ \midrule
AGNews                               & \citet{ranking-2005-corso}         & 4                             & News                          & EN                             & 108,000/12,000/7,600                              & 0.00/0.00/0.00                              \\
Topic\textsubscript{Dao}                               & \citet{daouadi2021optimizing}         & 2                             & Twitter                          & EN                             & 11,943/1,328/5,734                           & 0.00/0.00/0.00                              \\ \bottomrule
\end{tabular}
\caption{Description of benchmark datasets. We include 16 English in-domain datasets, eight English out-of-domain datasets, nine Twitter datasets in three different languages, and two topic classification datasets. To facilitate reference, we give each dataset a name as Task column shows. \textbf{Cls} column indicates the number of classes. \textbf{Lang:} Language, \textbf{\% of Emoji Samples:} Percentage of samples of downstream datasets containing emojis.}\label{tab:eval_data}
\end{table*} 

\subsection{Evaluation Data}\label{subsec:app_eval_data}
\paragraph{\textbf{In-Domain Datasets.}} We collect $16$ English witter datasets representing eight different SM tasks to evaluate our models, including (1) crisis awareness task~\cite{olteanu2014crisislex}, (2) emotion recognition~\cite{mohammad-2018-semeval}, (3) hateful and offensive language detection~\cite{waseem-2016-hateful,davidson-2017-hateoffensive,basile-2019-semeval,zampieri-2019-predicting}, (4) humor identification~\cite{meaney2021hahackathon}, (5) irony and sarcasm detection~\cite{van-hee2018semeval,riloff2013sarcasm,ptavcek2014sarcasm,rajadesingan2015sarcasm,bamman2015contextualized}, (6) irony type identification~\cite{van-hee2018semeval} (7) sentiment analysis~\cite{thelwall2012sentiment, rosenthal-2017-semeval}, and (8) stance detection~\cite{mohammad-2016-semeval}. We present the distribution, the number of labels, and the short name of each dataset in Table~\ref{tab:eval_data}.

\paragraph{\textbf{Out-of-Domain Datasets.}} We evaluate our model on downstream SM tasks from diverse social media platforms and domains. For emotion recognition task, we utilize (1) PsychExp~\cite{wallbott1986universal}, a seven-way classification dataset of self-described emotional experiences created by psychologists, and (2) GoEmotion~\cite{demszky2020goemotion}, a dataset of Reddit posts annotated with 27 emotions (we exclude neutral samples). For sarcasm detection task, we use two datasets from the Internet Argument Corpora~\cite{walker2012corpus,oraby2016creating} that posts from debate forums. For sentiment analysis, we utilize (1) five-class and binary classification versions of the Stanford Sentiment Treebank~\cite{socher-2013-recursive} (SST-5 and SST-2) that include annotated movie reviews with sentiment tags, (2) movie review (MR) for binary sentiment classification~\cite{pang2005seeing}, and (3) SentiStrength for YouTube comments (SS-YouTube)~\cite{thelwall2012sentiment}.

\paragraph{\textbf{Multilingual Datasets.}} As explained, to evaluate the effectiveness of our framework on different languages, we collect nine Twitter tasks in three languages: Arabic, Italian, and Spanish. For each language, we include three emotion-related tasks, (1) emotion recognition\cite{mageed-2020-aranet, bianchi2021feel, mohammad-2018-semeval}, (2) irony identification~\cite{idat2019ghanem, cignarella2018overview, ortega2019overview}, and (3) offensive language/hate speech detection~\cite{mubarak2020overview, bosco2018overview, basile-2019-semeval}.

\paragraph{\textbf{Few-Shot Data.}} We conduct our few-shot experiments only on our English language downstream data. We use different sizes from the set \{$20$, $100$, $500$, $1,000$\} sampled randomly from the respective Train splits of our data. For each of these sizes, we randomly sample five times with replacement (as we report the average of five runs in our experiments). We also run few-shot experiments with varying percentages of the Train set of each task (i.e., $1\%$, $5\%$, $10\%$, $20\%$ \dots $90\%$). We randomly sample \textbf{five} different training sets for each percentage, evaluate each model on the original Dev and Test sets, and average the performance over five runs.

\paragraph{\textbf{Topic Classification Datasets.}} To investigate the generalizability of our models, we evaluate our models on two topic classifcation datasets: AGNews~\cite{ranking-2005-corso} and Topic\textsubscript{Dao}~\cite{daouadi2021optimizing}. Given a news title and a short description, AGNews classifies the input text into four categories, including world, sports, business, and Sci/Tech. Topic\textsubscript{Dao} identifies if a given tweet is related to politics or not. The data distribution is presented in Table~\ref{tab:eval_data}.

\paragraph{\textbf{SentEval.}} We utilize SentEval benchmark~\cite{conneau-2018-senteval}\footnote{\url{https://github.com/facebookresearch/SentEval}}, a toolkit for evaluating the quality of sentence representations, to evaluate on seven semantic textual similarity (STS) datasets and eight transfer learning datasets. Seven STS datasets include STS 2012-2016~\cite{agirre-2012-sts,agirre-2013-sts,agirre-2014-sts,agirre-2015-sts,agirre-2016-sts}, SICK-Relatedness~\cite{marelli-2014-sick}, and STS Benchmark~\cite{cer-2017-semeval}. Eight transferring classification datasets consist of four sentiment analysis (i.e., movie review (MR)~\cite{pang2005seeing}, product review (CR)~\cite{hu-2004-mining}, SST2, and SST5~\cite{socher-2013-recursive}), subjectivity detection (SUBJ)~\cite{pang-2004-sentiment}, opinion polarity (MPQA)~\cite{weibe-2005-annotating}, question-type classification (TREC)~\cite{voorhees-2000-building}, and paraphrase detection (MRPC)~\cite{dolan-2005-automatically}. The data distribution and evaluation metrics are presented in Table~\ref{tab:data_senteval}. The STS datasets only have test set since they do not train any model. Tasks of MR, CR, SUBJ and MPQA are evaluated by nested 10-fold cross-validation, TREC and MRPC use cross-validation, and two SST datasets have standard development and test sets. 

\begin{table}[ht]
\small
\centering
\begin{tabular}{@{}lcccc@{}}
\toprule
\multicolumn{1}{c}{\textbf{Task}} & \textbf{Train} & \textbf{Dev} & \textbf{Test} & \textbf{Metric} \\ \midrule
STS12                             & -              & -            & 3.1K          & spearman        \\
STS13                             & -              & -            & 1.5K          & spearman        \\
STS14                             & -              & -            & 3.7K          & spearman        \\
STS15                             & -              & -            & 8.5K          & spearman        \\
STS16                             & -              & -            & 9.2K          & spearman        \\
SICK-R                            & -              & -            & 1.4K          & spearman        \\
STS-B                             & -              & -            & 4.9K          & spearman        \\ \midrule
MR                                & 10.6K          & -            & 10.6K         & accuracy        \\
CR                                & 3.7K           & -            & 3.7K          & accuracy        \\
SUBJ                              & 10.0K          & -            & 10.0K         & accuracy        \\
MPQA                              & 10.6K          & -            & 10.6K         & accuracy        \\
SST2                              & 67.3K          & 872          & 1.8K          & accuracy        \\
SST5                              & 8.5K           & 1.1K         & 2.2K          & accuracy        \\
TREC                              & 5.5K           & -            & 500           & accuracy        \\
MRPC                              & 4.1K           & -            & 1.7K          & accuracy        \\ \bottomrule
\end{tabular}
\caption{Description of SentEval benchmark~\cite{conneau-2018-senteval}. For STS datasets, we report overall Spearman’s correlation across all topics.}\label{tab:data_senteval}
\end{table}

\section{Experiment}

\subsection{Implementation}\label{subsec:app_implement}
For experiments on English language datasets, we initialize our model with a pre-trained English RoBERTa\textsubscript{Base}~\cite{liu2019roberta} model from Huggingface's Transformers~\cite{wolf2020transformers} library. RoBERTa\textsubscript{Base} consists of $12$ Transformer Encoder layers, $768$ hidden units each, $12$ attention heads, and contains $110$M parameters in entire model. RoBERTa uses a byte-pair-encoding vocabulary with a size of $50,265$ tokens. RoBERTa was pre-trained on large English corpora (e.g., Bookcorpus) with the MLM objective. In accordance with convention~\cite{liu2019roberta,gao2021simcse}, we pass the hidden state corresponding to the `[CLS]' token from the last layer through a feed-forward layer with hidden size of $768$ and a hyperbolic tangent function and, then, use the output as the sentence-level embedding, $h_i$. For the classification objective, we feed $h_i$ into a feed-forward layer with hidden size of $1,067$~\footnote{The number of Emoji classes is $1,067$.}, a softmax function and a dropout of $0.1$. For multi-lingual experiments, we utilize the pre-trained XLM-RoBERTa\textsubscript{Base} model\footnote{For short, we refer to the official released XLM-RoBERTa\textsubscript{Base} as XLM-R in the rest of the paper.}~\cite{xlmr-2020-alexis} as our initial checkpoint. XLM-R\textsubscript{Base} has the same architecture as RoBERTa. XLM-R includes a vocabulary of $250,002$ BPE tokens for 100 languages and is pre-trained on $2.5$TB of filtered CommonCrawl.

We fine-tune pre-trained models on each downstream task for five times with different random seeds and report the averaged model performance. Our main metric is macro-averaged $F_1$ score. To evaluate the overall ability of a model, we also report an aggregated metric that averages over the $16$ Twitter datasets, eight out-of-domain tasks, and the nine multi-lingual Twitter datasets, respectively. 

\paragraph{\textbf{NPMI weighting matrix.}} We randomly sample $150$M tweets from the $350$M tweets with at least one emoji each. We extract all emojis in each tweet and count the frequencies of emojis as well as co-occurrences between emojis. To avoid noisy relatedness from low frequency pairs, we filter out emoji pairs, $(y_i, y_a)$, whose co-occurrences are less than $20$ times or $0.02\times$ frequency of $y_i$. We employ Eq.~\ref{eq:npmi} to calculate NPMI for each emoji pair. Similarly, we calculate the NPMI weighting matrix using $150M$ with at least one hashtag each and filtering out low frequency pairs.  

\subsection{Baselines}\label{sec:baseline}

We compare our proposed framework against $11$ strong baselines, which we describe here. \textbf{(1) RoBERTa:} The original pre-trained RoBERTa, fine-tuned on downstream tasks with standard cross-entropy loss. \textbf{(2) MLM:} We continue pre-training RoBERTa on our pre-training dataset (\texttt{TweetEmoji-EN} for emoji-based experiment and \texttt{TweetHashtag-EN} for hashtag-based experiment) with solely the MLM objective in Eq.~\ref{eq:mlm_loss} (Appendix~\ref{subsec:app_mlm}), then fine-tune on downstream tasks. \textbf{(3) Emoji-Based MLM (E-MLM):} Following~\citet{corazza-2020-hybrid}, we mask emojis in tweets and task the model to predict them, then fine-tune on downstream tasks.\footnote{For hashtag-based experiment, we adapt this method to masking hashtags in tweets and refer to it as Hashtag-based MLM (H-MLM).} \textbf{(4) SLP.} A RoBERTa model fine-tuned on the \textit{surrogate label prediction} task (e.g., emoji prediction)~\cite{zhang-abdul-mageed-2022-improving} with cross-entropy loss, then fine-tuned on downstream tasks. 
\textbf{Supervised Contrastive Learning:} We also compare to state-of-the-art supervised contrastive fine-tuning frameworks. We take the original pre-trained RoBERTa and fine-tune it on each task with \textbf{(5) SCL}~\cite{gunel2020supervised} and \textbf{(6) LCL}~\cite{suresh2021not}, respectively. Both works combine supervised contrastive loss with standard cross-entropy as well as augmentation of the training data to construct positive pairs. We follow the augmentation technique used in~\citet{suresh2021not}, which replaces $30\%$ of words in the input sample with their synonyms in WordNet dictionary~\cite{miller1995wordnet}. \textbf{Self-Supervised Contrastive Learning.} We further train RoBERTa on different recently proposed self-supervised contrastive learning frameworks. \textbf{(7) SimCSE-Self.}~\citet{gao2021simcse} introduce SimCSE where they produce a positive pair by applying different dropout masks on input text twice. We similarly acquire a positive pair using the same droput method. \textbf{(8) SimCSE-Distant.}~\citet{gao2021simcse} also propose a supervised SimCSE that utilizes gold NLI data to create positive pairs where an anchor is a premise and a positive sample is an entailment. 
Hence, we adapt the supervised SimCSE framework to our distantly supervised data and construct positive pairs applying our epoch-wise re-pairing strategy. Specifically, each anchor has one positive sample that employs the same emoji as the anchor in a batch. 
\textbf{(9) Mirror-BERT.}~\cite{liu2021fast} construct positive samples in Mirror-BERT by random span masking as well as different dropout masks. After contrastive learning, sentence-encoder models are fine-tuned on downstream tasks with the cross-entropy loss.
\textbf{(10) Weakly-supervised Contrastive Learning.} We simplify and adapt the WCL framework of~\citet{zheng2021weakly} to language: We first encode unlabelled tweets to sequence-level representation vectors using the hidden state of the `[CLS]' token from the last layer of RoBERTa. All unlabelled tweets are clustered by applying $k$-means to their representation vectors. We then use the cluster IDs as weak labels to perform an SCL to pull the tweets assigned to the same cluster closer. Following~\citet{zheng2021weakly}, we also include an SSCL loss by augmenting the positive sample of an anchor using random span as well as dropout masking. We jointly optimize the SCL and SSCL losses in our implementation. 
\textbf{(11) Domain-Specific PLM (BTw):} We compare to the SoTA domain-specific PLM, BERTweet~\cite{nguyen-etal-2020-bertweet}. BERTweet was pre-trained on $850$M tweets with RoBERTa\textsubscript{Base} architecture. We download the pre-trained BERTweet checkpoint from Huggingface's Transformers~\cite{wolf2020transformers} library and fine-tune it on each downstream task with cross-entropy loss. More details about hyper-parameters of these baselines are in Appendix~\ref{append:subsec:hyper}.


\subsection{Hyper-Parameters} \label{append:subsec:hyper}
\begin{figure*}[!ht]
\centering
\begin{subfigure}[]{.32\textwidth}
  \centering
  \includegraphics[width=\linewidth]{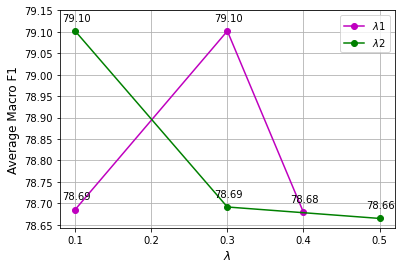}
  \caption{$\lambda$}
\end{subfigure}%
\begin{subfigure}[]{.32\textwidth}
  \centering
  \includegraphics[width=\linewidth]{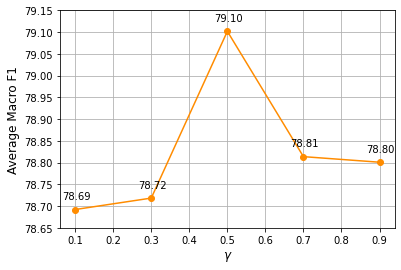}
  \caption{$\gamma$}
\end{subfigure}
\begin{subfigure}[]{.32\textwidth}
  \centering
  \includegraphics[width=\linewidth]{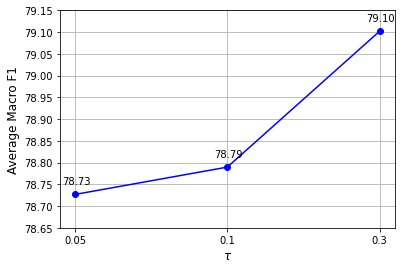}
  \caption{$\tau$}
\end{subfigure}%
\caption{Hyper-parameter Optimization. We report the average validation $F_1$ across $15$ English in-domain datasets.}\label{fig:hyper_sensitive}
\end{figure*}
\paragraph{\textbf{InfoDCL Training Hyper-Parameters.}} For hyper-parameter tuning of our proposed InfoDCL framework, we randomly sample $5$M tweets from the training set of our \texttt{TweetEmoji-EN}. We continue training the pre-trained RoBERTa for three epochs with Adam optimizer with a weight decay of $0.01$ and a peak learning rate of $2e-5$. The batch size is $128$, and the total number of input samples is $256$ after constructing positive pairs. As~\citet{gao2021simcse} find contrastive learning is not sensitive to the learning rate nor batch size when further training a PLM, we do not fine-tune these (i.e., the learning rate and batch size) in this paper. Following~\cite{liu2019roberta}, we mask $15\%$ of tokens for our MLM objective. We fine-tune the loss scaling weights $\lambda_1$ in a set of $\{0.1, 0.3, 0.4\}$, $\lambda_2$ in a set of $\{0.1, 0.3, 0.5\}$, and $\gamma$ in a set of $\{0.1, 0.3, 0.5, 0.7, 0.9\}$. To reduce search space, we use the same temperature value for the $\tau$ in Eq.~\ref{eq:lcl} and Eq.~\ref{eq:iwscl} and fine-tune in a set of $\{0.1, 0.3, 0.5, 0.7, 0.9\}$. We use grid search to find the best hyper-parameter set and evaluate performance on the Dev set of the $15$ English language Twitter datasets (excluding Senti\textsubscript{Thel}).~\footnote{We fine-tune the learned model on each downstream task with an arbitrary learning rate of $5e-6$, a batch size of $16$, and a training epoch of $20$. The performance is macro-$F_1$ over three runs with random seeds.} We select the best hyper-parameter set that achieves the best macro-$F_1$ averaged over the $15$ downstream tasks. Our best hyper-parameter set is $\lambda_1=0.3$, $\lambda_2=0.1$, $\gamma=0.5$, and $\tau = 0.3$. As Figure~\ref{fig:hyper_sensitive} shows, our model is not sensitive to changes of these hyper-parameters, and we observe that all the differences are less than $0.45$ comparing to the best hyper-parameter set. Finally, we continue training RoBERTa/BERTweet on the full training set of \texttt{TweetEmoji-EN} with InfoDCL framework and best hyper-parameters. We train InfoDCL model for three epochs and utilize $4$ Nvidia A100 GPU ($40$GB each) and $24$ CPU cores. Each epoch takes around $7$ hours.

\paragraph{\textbf{Downstream Task Fine-Tuning Hyper-Parameters.}} Furthermore, we take the model trained with the best hyper-parameters and search the best hyper-parameter set of downstream task fine-tuning. We search the batch size in a set of $\{8, 16, 32, 64\}$ and the peak learning rate in a set of $\{2e-5, 1e-5, 5e-6\}$. We identify the best fine-tuning hyper-parameters based on the macro-$F_1$~\footnote{We run three times and use the mean of them.} on Dev sets averaged over the $16$ English language Twitter datasets. Our best hyper-parameters for fine-tuning is a learning rate of $1e-5$ and a batch size of $32$. For all the downstream task fine-tuning experiments in this paper, we train a model on the task for $20$ epochs with early stop ($patience = 5$ epochs). We use the same hyper-parameters identified in this full data setting for our few-shot learning. For each dataset, we fine-tune for five times with a different random seed every time, and report the mean macro-$F_1$ of the five runs. Each downstream fine-tuning experiment use a single Nvidia A100 GPU ($40$GB) and $4$ CPU cores.

\paragraph{\textbf{Baseline Hyper-Parameters.}} Our \textbf{Baseline (1)} is directly fine-tuning RoBERTa on downstream tasks. We fine-tune Baseline (1) hyper-parameters as follows: The batch size is chosen from a set of $\{8, 16, 32, 64\}$ and the peak learning rate in a set of $\{2e-5, 1e-5, 5e-6\}$. The best hyper-parameters for RoBERTa fine-tuning is a learning rate of $2e-5$ and a batch size of $64$. 

For \textbf{Baseline (2-3)}, we further pre-train the RoBERTa model for three epochs (same as our InfoDCL) with the MLM objective with an arbitrary learning rate of $5e-5$ and a batch size of $4,096$. We mask $15\%$ of tokens in each input tweet. For Baseline (3), we give priority to masking emojis in a tweet: if the emoji tokens are less than $15\%$, we then randomly select regular tokens to complete the percentage of masking to the $15\%$. \textbf{Baseline (4)} is about surrogate label prediction (with emojis). We also train Baseline (4) for three epochs with  a learning rate of $2e-5$ and a batch size of $4,096$. After training, models are fine-tuned on downstream tasks using the same hyper-parameters as our proposed model. 

\textbf{Baselines (5-7).} \textbf{SimCSE}~\cite{gao2021simcse} was trained in two setups, i.e., self-supervised and supervised by label data. We also train RoBERTa on both settings. For \textit{self-supervised SimCSE}, we train RoBERTa on our pre-training dataset for three epochs with a learning rate of $2e-5$, a batch size of $256$, and $\tau$ of $0.05$. For the \textit{distantly-supervised SimCSE}, we construct positive pairs as described in Section~\ref{subsec:weakly_pairing}. Similar to self-supervised SimCSE, we train RoBERTa for three epochs with a learning rate of $2e-5$ but with a batch size of $128$.~\footnote{After pairing, each batch include $256$ unique tweets.} The pre-training of \textbf{Mirror-BERT} is similar to the pre-training of self-supervised SimCSE. We set the span masking rate of $k=3$,  a temperature of $0.04$, a learning rate of $2e-5$, and a batch size of $256$. Trained models, then, are fine-tuned on downstream tasks. For downstream task fine-tuning with baselines 2-7, we use the same hyper-parameters identified with InfoDCL downstream task fine-tuning.
%

\textbf{Baselines (8-9).} \textbf{SCL}~\cite{gunel2020supervised} and LCL~\cite{suresh2021not} directly fine-tune on downstream tasks with cross-entropy loss. We reproduce these two methods on our evaluation tasks. For SCL, we follow~\citet{gunel2020supervised} and fine-tune each task with a temperature of $\tau = 0.3$, a SCL scaling weighting of $0.9$, and a learning rate of $2e-5$. For LCL, we fine-tune each task with a temperature $\tau$ of $0.3$, a LCL scaling weighting of $0.5$, and a learning rate of $2e-5$.

\textbf{Baselines (10).} We implement \textbf{WCL}~\cite{zheng2021weakly} to continue train RoBERTa with our emoji dataset. We remove all emojis in the $31$M tweets and encode tweets using the hidden state of `[CLS]' token from the last layer of RoBERTa. The tweets are then clustered by \textit{k}-means clustering algorithm.\footnote{We use mini-batch \textit{k}-means clustering from scikit-learn~\cite{scikitlearn-2011-fabian}. } For hyper-parmeter tuning of WCL, we randomly sample $5$M tweets from the training set of \texttt{TweetEmoji-EN} and train a model for three epochs with different hyper-parmeter sets. We search the number of clusters in a set of $\{200, 500, 1067, 2000\}$ and temperature $\tau$ in a set of $\{0.1, 0.3\}$. To reduce the search space, we use the same temperature value for SSCL and SCL losses. We evaluate performance on the Dev set of the $16$ English language Twitter datasets~\footnote{We fine-tune the trained WCL model with a learning rate of $1e-5$ and a batch size of $32$.} and find the best hyper-parameter set is $k=1067$ and $\tau=0.1$. We then train WCL on the \texttt{TweetEmoji-EN} dataset for three epochs with our best hyper-parameters and fine tune the model on $24$ downstream tasks with the same hyper-parameters identified for InfoDCL downstream fine-tuning.\footnote{For hashtag-based experiment, we use the same hyper-parameters.}

\textbf{Baseline (11).} We fine-tune BERTweet with hyperparameters utilized in~\cite{nguyen-etal-2020-bertweet} that are a fixed learning of $1e-5$ and a batch size of $32$.
\begin{table}[h]
\centering
\tiny
\begin{tabular}{@{}lccccrr@{}}
\toprule
                     & \multicolumn{1}{c}{$\lambda_1$} & \multicolumn{1}{c}{$\lambda_2$} & \multicolumn{1}{c}{$\gamma$} & \multicolumn{1}{c}{$\tau$} & \multicolumn{1}{c}{\textbf{lr}} & \multicolumn{1}{c}{\textbf{batch}} \\ \midrule
InfoDCL PT (emoji) & 0.3                             & 0.1                             & 0.5                          & 0.3                        & $2e-5$                 & 128                            \\
InfoDCL PT (hashtag) & 0.4                             & 0.1                             & 0.1                          & 0.1                        & $2e-5$                 & 128                            \\
DCL PT (emoji) & -                             & -                            & 0.5                          & 0.3                        & $2e-5$                 & 128                            \\
DCL PT (hashtag) & -                             & -                             & 0.1                          & 0.1                        & $2e-5$                 & 128                            \\
Downstream FT  & -                               & -                               & -                            & -                          & $1e-5$                 & 32                             \\
RoBERTa FT             & -                               & -                               & -                            & -                          & $2e-5$                 & 64                             \\
MLM                  & -                               & -                               & -                            & -                          & $5e-5$                 & 4,096                          \\
E-MLM                & -                               & -                               & -                            & -                          & $5e-5$                 & 4,096                          \\
SLP                  & -                               & -                               & -                            & -                          & $2e-5$                 & 4,096                          \\
SimCSE-Self          & -                               & -                               & -                            & 0.05                       & $2e-5$                 & 256                            \\
SimCSE-Distant       & -                               & -                               & -                            & 0.05                       & $2e-5$                 & 128                            \\
Mirror-BERT          & -                               & -                               & -                            & 0.04                       & $2e-5$                 & 256                            \\
SCL                  & -                               & -                               & -                            & 0.30                        & $2e-5$                 & 32                             \\
LCL                  & -                               & -                               & -                            & 0.30                        & $2e-5$                 & 32                             \\
WCL                  & -                               & -                               & -                            & 0.10                        & $2e-5$                 & 256                            \\
BERTweet FT            & -                               & -                               & -                            & -                          & $1e-5$                 & 32                             \\ \bottomrule
\end{tabular}
\caption{Hyper-parameter values using in this paper. \textbf{PT:} Pre-training, \textbf{FT:} Downstream fine-tuning. }
\end{table}

\paragraph{\textbf{Multi-Lingual Experiment Hyper-Parameters.}} For multi-lingual experiments, we utilize the pre-trained XLM-RoBERTa\textsubscript{Base} model~\cite{xlmr-2020-alexis} as our initial checkpoint. We continue training XLM-R on multi-lingual tweets with our framework and the best hyperparameters identified for English. For the downstream fine-tuning, we use as same as the best hyperparameters identified for English tasks.

\paragraph{\textbf{Hahstag Experiment Hyper-Parameters.}} For the hashtag-based experiments presented in Section~\ref{sec:hashtag}, we use the same hyper-parameter optimization set up to find the best hyper-parameter set for hashtag-based models. The best hyper-parameter set for hashtag-based models is $\lambda_1=0.4$, $\lambda_2=0.1$, $\gamma=0.1$, and $\tau = 0.1$. We then use the same downstream fine-tuning hyper-parameters identified with emoji-based InfoDCL for downstream task.
 
\section{Results}
\subsection{Standard Deviation and Significance Tests}\label{sec:app_std_sig}
\begin{table*}[ht]
\tiny
\centering
\begin{tabular}{@{}llcccccccccccc|cc@{}}
\toprule
 & \multicolumn{1}{c}{\textbf{Task}}            & \textbf{RB} & \textbf{MLM} & \textbf{E-MLM} & \textbf{SLP} & \textbf{Mir-B} & \textbf{Sim-Self} & \textbf{Sim-D} & \textbf{SCL} & \textbf{LCL} & \textbf{WCL} & \textbf{DCL} & \textbf{InfoDCL-R} & \textbf{BTw} & \textbf{InfoDCL-B} \\ \midrule
 & Crisis\textsubscript{Oltea} & 0.15        & 0.15         & 0.23           & 0.17         & 0.24           & 0.30              & 0.25           & 0.23         & 0.13         & 0.29         & 0.25         & 0.15            & 0.26         & 0.07            \\
 & Emo\textsubscript{Moham}    & 1.60        & 0.85         & 0.72           & 1.05         & 0.50           & 0.85              & 0.70           & 0.56         & 0.37         & 0.53         & 0.93         & 0.79            & 0.66         & 0.70            \\
 & Hate\textsubscript{Was}     & 0.21        & 0.63         & 0.79           & 0.55         & 0.21           & 0.19              & 0.40           & 0.21         & 0.25         & 0.24         & 0.67         & 0.41            & 0.63         & 0.57            \\
 & Hate\textsubscript{Dav}     & 1.31        & 0.85         & 0.58           & 0.36         & 1.71           & 1.39              & 1.04           & 0.43         & 1.24         & 0.93         & 0.81         & 0.61            & 0.78         & 0.76            \\
 & Hate\textsubscript{Bas}     & 1.96        & 2.20         & 1.86           & 1.64         & 0.82           & 1.62              & 2.65           & 3.52         & 1.20         & 2.21         & 0.47         & 1.00            & 3.50         & 1.88            \\
 \multirow{7}{*}{\rotatebox[origin=c]{90}{\textbf{In-Domain}}} & Humor\textsubscript{Mea}    & 0.47        & 0.38         & 0.65           & 0.38         & 0.38           & 0.87              & 0.59           & 0.65         & 0.66         & 0.73         & 0.19         & 0.62            & 0.15         & 0.48            \\
 & Irony\textsubscript{Hee-A}  & 1.30        & 1.06         & 0.85           & 1.02         & 1.11           & 0.87              & 1.35           & 1.13         & 0.95         & 1.46         & 1.38         & 1.51            & 1.38         & 0.85            \\
 & Irony\textsubscript{Hee-B}  & 1.60        & 0.63         & 2.43           & 2.38         & 0.56           & 0.84              & 2.70           & 2.03         & 1.44         & 0.89         & 1.05         & 0.53            & 2.06         & 3.19            \\
 & Offense\textsubscript{Zamp} & 1.41        & 0.37         & 0.78           & 0.50         & 1.32           & 1.67              & 0.60           & 0.83         & 0.15         & 0.42         & 0.85         & 1.51            & 1.96         & 0.92            \\
 & Sarc\textsubscript{Riloff}  & 1.47        & 1.34         & 2.58           & 1.26         & 4.32           & 2.06              & 1.86           & 2.79         & 2.03         & 1.15         & 0.85         & 1.09            & 1.69         & 1.60            \\
 & Sarc\textsubscript{Ptacek}  & 0.30        & 0.10         & 0.10           & 0.22         & 0.18           & 0.28              & 0.21           & 0.23         & 0.14         & 0.17         & 0.12         & 0.07            & 0.23         & 0.10            \\
 & Sarc\textsubscript{Rajad}   & 0.51        & 0.30         & 0.30           & 0.71         & 0.57           & 0.27              & 0.22           & 0.55         & 0.55         & 0.58         & 0.47         & 0.49            & 0.73         & 0.64            \\
 & Sarc\textsubscript{Bam}     & 0.54        & 0.61         & 0.87           & 0.38         & 0.69           & 1.18              & 0.60           & 0.83         & 0.78         & 0.36         & 0.48         & 0.39            & 0.31         & 0.71            \\
 & Senti\textsubscript{Rosen}  & 0.93        & 1.64         & 0.35           & 0.91         & 1.06           & 0.57              & 0.67           & 1.14         & 0.40         & 0.73         & 0.76         & 0.52            & 0.40         & 0.43            \\
 & Senti\textsubscript{Thel}   & 0.61        & 1.01         & 0.69           & 0.33         & 0.65           & 0.50              & 0.56           & 1.29         & 0.85         & 0.54         & 0.78         & 0.62            & 0.63         & 0.66            \\
 & Stance\textsubscript{Moham} & 0.87        & 1.55         & 0.80           & 1.07         & 1.40           & 1.94              & 1.67           & 1.01         & 1.66         & 1.11         & 1.25         & 1.33            & 1.35         & 1.37            \\ \cdashline{2-16}
 & Average                                      & 0.24        & 0.24         & 0.20           & 0.26         & 0.23           & 0.17              & 0.31           & 0.35         & 0.42         & 0.23         & 0.24         & 0.19            & 0.33         & 0.20            \\ \midrule
 & Emotion\textsubscript{Wall} & 0.41        & 0.78         & 0.69           & 1.01         & 1.14           & 0.40              & 0.33           & 0.73         & 0.36         & 0.73         & 1.13         & 0.26            & 1.50         & 0.85            \\
\multirow{7}{*}{\rotatebox[origin=c]{90}{\textbf{Out-of-Domain}}}  & Emotion\textsubscript{Dem}  & 0.58        & 0.60         & 0.42           & 0.80         & 0.71           & 0.88              & 0.74           & 0.52         & 1.05         & 0.86         & 1.28         & 0.61            & 1.20         & 1.73            \\
 & Sarc\textsubscript{Walk}    & 1.29        & 1.14         & 0.99           & 0.98         & 1.25           & 4.09              & 1.01           & 0.88         & 1.19         & 0.59         & 1.66         & 1.11            & 0.69         & 0.72            \\
 & Sarc\textsubscript{Ora}     & 1.20        & 1.41         & 0.99           & 0.24         & 1.56           & 1.85              & 0.32           & 1.33         & 1.70         & 1.21         & 0.68         & 0.77            & 1.05         & 1.00            \\
 & Senti-MR                                     & 0.56        & 0.29         & 0.70           & 0.50         & 0.32           & 0.27              & 0.27           & 0.46         & 0.41         & 0.61         & 0.30         & 0.39            & 0.57         & 0.43            \\
 & Senti-YT                                     & 0.52        & 0.59         & 0.43           & 0.36         & 1.00           & 0.95              & 0.37           & 0.37         & 0.62         & 0.29         & 0.53         & 0.26            & 0.25         & 0.52            \\
 & SST-5                                        & 0.35        & 0.56         & 0.64           & 1.18         & 0.72           & 0.55              & 0.57           & 1.06         & 0.78         & 0.79         & 0.97         & 0.64            & 0.90         & 0.53            \\
 & SST-2                                        & 0.39        & 0.41         & 0.40           & 0.22         & 0.38           & 0.35              & 0.50           & 0.34         & 0.30         & 0.35         & 0.32         & 0.24            & 0.32         & 0.22            \\ \cdashline{2-16}
 & Average                                      & 0.31        & 0.15         & 0.27           & 0.41         & 0.19           & 0.42              & 0.21           & 0.26         & 0.17         & 0.14         & 0.54         & 0.27            & 0.28         & 0.12            \\ \bottomrule
\end{tabular}
\caption{Fine-tuning results on our $24$ SM datasets (standard deviation of macro-$F_1$ over five runs).}\label{app_tab:all_std}
\end{table*}
\begin{table}[h!]
\tiny
\centering
\begin{tabular}{@{}lcccc@{}}
\toprule
            & \multicolumn{2}{c}{\textbf{$p$-value (t-test)}} & \multicolumn{2}{c}{\textbf{Minimal Distance $\epsilon$ (ASO)}} \\ \cmidrule(l){2-3} \cmidrule(l){4-5}
            & \textbf{In-Domain}  & \textbf{Out-of-Domain}  & \textbf{In-Domain}         & \textbf{Out-of-Domain}         \\ \midrule
\multicolumn{5}{c}{\textbf{InfoDCL-RoBERTa vs.}}                                                                          \\ \midrule
RoBERTa     & 0.0000              & 0.0075                  & 0.0000                     & 0.0000                         \\
MLM         & 0.0002              & 0.0020                  & 0.0000                     & 0.0000                         \\
E-MLM       & 0.0100              & 0.0410                  & 0.0000                     & 0.0000                         \\
SLP         & 0.0213              & 0.0843                  & 0.0000                     & 0.0011                         \\
Mirror-B    & 0.0000              & 0.0001                  & 0.0000                     & 0.0000                         \\
SimSCE-self & 0.0000              & 0.0000                  & 0.0000                     & 0.0000                         \\
SimCSE-D    & 0.0818              & 0.0005                  & 0.0000                     & 0.0000                         \\
SCL         & 0.0003              & 0.0014                  & 0.0000                     & 0.0000                         \\
LCL         & 0.0003              & 0.0001                  & 0.0000                     & 0.0000                         \\
WCL         & 0.0001              & 0.0001                  & 0.0000                     & 0.0000                         \\
BERTweet    & 0.0960              & 0.0000                  & 0.0000                     & 0.0000                         \\ \midrule
\multicolumn{5}{c}{\textbf{InfoDCL-BERTweet vs.}}                                                                         \\ \midrule
BERTweet    & 0.0076              & 0.0377                  & 0.0321                     & 0.0000                         \\ \bottomrule
\end{tabular}
\caption{Significance tests on average macro-$F_1$ scores over $16$ in-domain datasets and eight out-of-domain datasets. For t-test, we compare our proposed models to all the baselines and report $p$-values. For ASO test, we report the minimal distance $\epsilon$ at significance level of $0.01$.}\label{app_tab:significance}
\end{table}
Table~\ref{app_tab:all_std} shows the standard deviations of our emoji-based InfoDCl models and all baselines over five runs. We conduct two significance tests on our results, i.e., the classical paired student’s t-test~\cite{fisher1936design} and Almost Stochastic Order (ASO)~\cite{dror-2019-deep} (better adapts to results of neural networks). As we pointed out earlier, we run each experiment five times with different random seeds. Hence, we conduct these two significance tests by inputting the obtained five evaluation scores on the Test set. Table~\ref{app_tab:significance} presents $p$-values for t-test and minimal distance $\epsilon$ at significance level of $0.01$ for ASO test. We also conduct significance tests on the results of individual tasks, finding that our InfoDCL-RoBERTa significantly ($p<0.05$) improves the original RoBERTa on $13$ (out of $24$) and $24$ (out of $24$) datasets based on t-test and ASO, respectively. InfoDCL-RoBERTa also significantly ($p<0.05$) outperforms BERTweet (the strongest baseline) on $10$ (out of $24$) and $15$ (out of $24$) tasks based on t-test and ASO, respectively. 

\subsection{Comparisons to Individual SoTAs.}\label{subsec:append_sota}
Although the focus of our work is on producing effective representations suited to the whole class of SM tasks, rather than to one or another of these tasks, we also compare our models on each dataset to other reported task-specific SoTA models on that particular dataset in Table~\ref{tab:compare}. 
We compare our methods on each dataset to other reported task-specific SoTA models on that particular dataset as shown. Due to diverse metrics used in previous studies, we compare models of each task reporting the corresponding metric of the SoTA method. Some SoTA models are trained on different data splits or use different evaluation approaches (e.g., \citet{olteanu2014crisislex} is evaluated by cross-validation). To provide meaningful comparisons, we thus fine-tune BERTweet on our splits and report against our models. Our InfoDCL-RoBERTa outperform SoTA on $11$ out of $16$ in-domain datasets and four out of eight out-of-domain datasets. We achieve the best average score over $16$ in-domain datasets applying our model on BERTweet. Further training RoBERTa with our framework obtains the best average score across the eight out-of-domain datasets. 
We note that some SoTA models adopt task-specific approaches and/or require task-specific resources. For example,~\citet{ke-2020-sentilare} utilize SentiWordNet to identify the sentiment polarity of each word. In this work, our focus on producing effective representations suited for the whole class of SM tasks, rather than one or another of these tasks. Otherwise, we hypothesize that task-specific approaches can be combined with our InfoDCL framework to yield even better performance on individual tasks.
\begin{table}[h!]
\centering
\tiny
\begin{tabular}{@{}llllccc@{}}
\toprule
&           \multicolumn{1}{c}{\textbf{Task}}            & \multicolumn{1}{c}{\textbf{Metric}} & \multicolumn{1}{c}{\textbf{SoTA}} & \multicolumn{1}{c}{\textbf{BTw}} & \multicolumn{1}{c}{\textbf{\begin{tabular}[c]{@{}c@{}}InfoDCL\\ RB\end{tabular}}} & \multicolumn{1}{c}{\textbf{\begin{tabular}[c]{@{}c@{}}InfoDCL\\ BTw\end{tabular}}} \\ \midrule
&           Crisis\textsubscript{Oltea} & M-$F_1$                    & \underline{95.60}\textsuperscript{$\star$}                    & 95.76                   & \textbf{96.01}                        & 95.84                         \\
&           Emo\textsubscript{Moham}    & M-$F_1$                    & 78.50\textsuperscript{$\spadesuit$}                    & 80.23                   & 81.34                        & \textbf{81.96}                         \\
&           Hate\textsubscript{Was}     & W-$F_1$                    & \underline{73.62}\textsuperscript{$\star\star$}                    & 88.95                   & 88.73                        & \textbf{89.12}                         \\
&           Hate\textsubscript{Dav}     & W-$F_1$                    & \underline{90.00}\textsuperscript{$\dagger$}                    & 91.26                   & 91.12                        & \textbf{91.27}                         \\
&           Hate\textsubscript{Bas}     & M-$F_1$                    & \textbf{65.10}\textsuperscript{$\heartsuit$}                    & 53.62                   & 52.84                        & 53.95                         \\
\multirow{7}{*}{\rotatebox[origin=c]{90}{\textbf{In-Domain}}}   &           Humor\textsubscript{Mea} & M-$F_1$                    & \textbf{98.54}\textsuperscript{$=$}                    & 94.43                   & 93.75                        & 94.04                         \\
&           Irony\textsubscript{Hee-A}  & $F^{(i)}_1$ & 70.50\textsuperscript{$\dagger\dagger$}                    & 73.99                   & 72.10                        & \textbf{74.81}                         \\
&           Irony\textsubscript{Hee-B}  & M-$F_1$                    & 50.70\textsuperscript{$\dagger\dagger$}                    & 56.73                   & 57.22                        & \textbf{59.15}                         \\
&           Offense\textsubscript{Zamp} & M-$F_1$                    & \textbf{82.90}\textsuperscript{$\ddagger$}                    & 79.35                   & 81.21                        & 79.83                         \\
&           Sarc\textsubscript{Riloff}  & $F^{(s)}_1$ & \underline{51.00}\textsuperscript{$\ddagger\ddagger$}                    & 66.59                   & 65.90                        & \textbf{69.28}                         \\
&           Sarc\textsubscript{Ptacek}  & M-$F_1$                    & \underline{92.37}\textsuperscript{$\mathsection$}                    & 96.40                   & 96.10                        & \textbf{96.67}                         \\
&           Sarc\textsubscript{Rajad}   & Acc                        & \underline{92.94}\textsuperscript{$\mathsection\mathsection$}                    & 95.30                   & 95.20                        & \textbf{95.32}                         \\
&           Sarc\textsubscript{Bam}     & Acc                        & \underline{\textbf{85.10}}\textsuperscript{$\|$}                    & 81.79                   & 81.51                        & 83.22                         \\
&           Senti\textsubscript{Rosen}  & M-Rec                      & 72.60\textsuperscript{$\spadesuit$}                    & \textbf{72.91}                   & 72.77                        & 72.46                         \\
&           Senti\textsubscript{Thel}   & Acc                        & 88.00\textsuperscript{$\diamondsuit$}                    & 89.81                   & \textbf{91.81}                        & 90.67                         \\
&           Stance\textsubscript{Moham} & Avg(a,f)                 & 71.00\textsuperscript{$\clubsuit$}                    & 71.26                   & \textbf{73.31}                        & 72.09                         \\ \cdashline{2-7}
&           Average                                      & \multicolumn{1}{c}{-}      & 78.65                    & 80.52                   & 80.68                        & \textbf{81.23}                         \\ \midrule
&           Emotion\textsubscript{Wall} & M-$F_1$                    & 57.00\textsuperscript{$\diamondsuit$}                    & 64.48                   & \textbf{68.41}                        & 65.61                         \\
\multirow{7}{*}{\rotatebox[origin=c]{90}{\textbf{Out-of-Domain}}} &           Emotion\textsubscript{Dem}  & W-$F_1$                    & 64.80\textsuperscript{$\bot$}                    & 64.53                   & \textbf{65.16}                        & 64.80                         \\
&           Sarc\textsubscript{Walk}    & M-$F_1$                    & \textbf{69.00}\textsuperscript{$\diamondsuit$}                    & 67.27                   & 68.45                        & 67.30                         \\
&           Sarc\textsubscript{Ora}     & M-$F_1$                    & 75.00\textsuperscript{$\diamondsuit$}                    & 77.33                   & \textbf{77.41}                        & 76.88                         \\
&           Senti-MR                                    & Acc                        & \underline{\textbf{90.82}}\textsuperscript{$\flat$}                    & 87.95                   & 89.43                        & 88.21                         \\
&           Senti-YT                                     & Acc                        & 93.00\textsuperscript{$\diamondsuit$}                    & 93.24                   & 93.12                        & \textbf{93.47}                         \\
&           SST-5                                        & Acc                        & \textbf{58.59}\textsuperscript{$\flat$}                    & 56.32                   & 57.74                        & 57.23                         \\
&           SST-2                                        & Acc                        & \textbf{96.70}\textsuperscript{$\natural$}                    & 93.32                   & 94.98                        & 93.73                         \\ \cdashline{2-7}
&           Average                                      & \multicolumn{1}{c}{-}      & 75.61                    & 75.55                   & \textbf{76.84}                        & 75.90                         \\ \bottomrule
\end{tabular}
\caption{Model comparisons. \textbf{SoTA:} Previous state-of-the-art performance on each respective dataset. \textbf{Underscore} indicates that our models are trained on different data splits to the SoTA model, where the result is not directly comparable. \textbf{BTw:} BERTweet~\cite{nguyen-etal-2020-bertweet}, a SOTA Transformer-based pre-trained language model for English tweets. We compare using the same metrics employed on each dataset. \textbf{Metrics:} \textbf{M-$F_1$:} macro $F_1$,  \textbf{W-$F_1$:} weighted $F_1$, $F_1^{(i)}$: $F_1$ irony class, \textbf{$F_1^{(i)}$:} $F_1$ irony class, $F_1^{(s)}$: $F_1$ sarcasm class, \textbf{M-Rec:} macro recall,  \textbf{Avg(a,f)}: Average $F_1$ of the \textit{against} and \textit{in-favor} classes (three-way dataset). \textsuperscript{$\star$}~\citet{liu2020crisisbert}, \textsuperscript{$\spadesuit$}~\citet{barbieri-2020-tweeteval},\textsuperscript{$\star\star$}~\citet{waseem-2016-hateful}, \textsuperscript{$\dagger$}~\citet{davidson-2017-hateoffensive}, \textsuperscript{$\heartsuit$}~\citet{basile-2019-semeval},
\textsuperscript{$^=$}~\citet{meaney2021hahackathon},
\textsuperscript{$\dagger\dagger$}~\citet{van-hee2018semeval}, \textsuperscript{$\ddagger$}~\citet{zampieri-2019-semeval}, \textsuperscript{$\ddagger\ddagger$}~\citet{riloff2013sarcasm}, \textsuperscript{$\mathsection$}~\citet{ptavcek2014sarcasm},  \textsuperscript{$\mathsection\mathsection$}~\citet{rajadesingan2015sarcasm}, \textsuperscript{$\|$}~\citet{bamman2015contextualized}, \textsuperscript{$\diamondsuit$}~\citet{felbo2017using}, \textsuperscript{$\clubsuit$}~\citet{mohammad-2016-semeval}, \textsuperscript{$\bot$}~\citet{suresh2021not}, \textsuperscript{$\flat$}~\citet{ke-2020-sentilare}, \textsuperscript{$\natural$}~\citet{tian-2020-skep}. 
}\label{tab:compare}
\end{table}

\begin{table}[t]
\centering
\scriptsize
\begin{tabular}{@{}llcccc|l@{}}
\toprule
\multirow{2}{*}{\textbf{L}}    & \multirow{2}{*}{\textbf{Task}}               & \multicolumn{1}{l}{\multirow{2}{*}{\textbf{XLM}}} & \multicolumn{3}{c}{\textbf{InfoDCL-XLMR}}  & \multirow{2}{*}{\textbf{SoTA}}                                                                                     \\ \cmidrule(l){4-6} 
                                  &                                              & \multicolumn{1}{c}{}                                & \multicolumn{1}{c}{\textbf{EN}} & \multicolumn{1}{c}{\textbf{Mono}} & \multicolumn{1}{c|}{\textbf{Mult}} & \\ \midrule
\multirow{3}{*}{\textbf{AR}}  & Emo\textsubscript{Mag}      & 72.23                                               & 72.08                                & 72.59                                     & \textbf{72.56}  &     60.32\textsuperscript{$\star$}                               \\
                                  & Irony\textsubscript{Ghan}   & 81.15                                               & 78.75                                & 81.85                                     & \textbf{82.23}         &   84.40\textsuperscript{$\dagger$}                          \\
                                  & Offense\textsubscript{Mub}  & 84.87                                               & 85.08                                & 85.61                                     & \textbf{87.10}             &         \underline{90.50}\textsuperscript{$\ddagger$}                \\ \hdashline
\multirow{3}{*}{\textbf{IT}} & Emo\textsubscript{Bian}     & 70.37                                               & 73.51                                & 73.58                                     & \textbf{74.36}                &         71.00\textsuperscript{$\mathsection$}             \\
                                  & Irony\textsubscript{Cig}    & 73.22                                               & 73.52                                & \textbf{74.07}                                     & 73.42                &         73.10\textsuperscript{$\spadesuit$}             \\
                                  & Hate\textsubscript{Bos}     & 78.63                                               & 78.06                                & 79.44                                     & \textbf{79.77}                    &       79.93\textsuperscript{$\diamondsuit$}           \\ \hdashline
\multirow{3}{*}{\textbf{ES}} & Emo-es\textsubscript{Moham} & 76.61                                               & 76.59                                & 77.29                                     & \textbf{77.66}                       &   ---               \\
                                  & Irony\textsubscript{Ort}    & 72.88                                               & 73.11                                & 72.98                                     & \textbf{74.91}               &      71.67\textsuperscript{$\clubsuit$}                \\
                                  & Hate-es\textsubscript{Bas}  & 76.07                                               & 75.33                                & 76.33                                     & \textbf{77.03}               &      73.00\textsuperscript{$\heartsuit$}                     \\ \hdashline
\textbf{}                         & \textbf{Average}                             & 76.23                                               & 76.23                                & 77.08                                     & \textbf{77.67}    &  ---                                  \\ \bottomrule
\end{tabular}
\caption{Results of multi-lingual tasks on macro-$F_1$. \textbf{SoTA:} Previous SoTA performance on each respective dataset. \textbf{Underscore} indicates that our models are trained on different data splits to the SoTA model. \textbf{L:} Language, \textbf{XLM:} XLM-R. Downstream task: \textbf{AR:} Arabic, \textbf{IT:} Italian, \textbf{ES:} Spanish. Pre-raining data: \textbf{EN:} English monolingual tweets, \textbf{Mono:} mono-lingual tweets in corresponding language, \textbf{Mul:} combined data that includes four languages and a total number of 4.5M tweets. \textsuperscript{$\star$}~\cite{mageed-2020-aranet}, \textsuperscript{$\dagger$}~\cite{idat2019ghanem}, \textsuperscript{$\ddagger$}~\cite{mubarak2020overview}, \textsuperscript{$\mathsection$}~\cite{bianchi2021feel},  \textsuperscript{$\spadesuit$}~\cite{cignarella2018overview}, \textsuperscript{$\diamondsuit$}~\cite{bosco2018overview}, \textsuperscript{$\clubsuit$}~\cite{ortega2019overview}, \textsuperscript{$\heartsuit$}~\cite{basile-2019-semeval}. }\label{tab:multi-lingual}
\end{table}
\begin{table*}[!ht]
\centering
\scriptsize
\begin{tabular}{@{}llcccccccccc|cc@{}}
\toprule
\multicolumn{1}{c}{\textbf{}}   & \multicolumn{1}{c}{\textbf{Task}}            & \textbf{RB} & \textbf{MLM}   & \textbf{H-MLM} & \textbf{SLP}   & \textbf{Mir-B} & \textbf{Sim-S} & \textbf{Sim-D} & \textbf{WCL} & \textbf{DCL}   & \textbf{InfoD-R} & \textbf{BTw}   & \textbf{InfoD-B} \\ \midrule
                                & Crisis\textsubscript{Oltea} & 95.87       & 95.75          & 95.74          & 95.96          & 96.12          & 95.88          & \textbf{95.94} & 95.84        & 95.92          & \textbf{95.94}  & 95.76          & \textbf{95.84}  \\
                                & Emo\textsubscript{Moham}    & 78.76       & 79.17          & 79.70          & 78.85          & 78.67          & 77.58          & 80.55          & 77.33        & 80.36          & \textbf{80.58}  & \textbf{80.23} & 80.22           \\
                                & Hate\textsubscript{Was}     & 57.01       & 57.70          & 57.22          & 57.55          & 56.78          & 56.40          & 56.40          & 57.59        & 57.17          & 56.64           & \textbf{57.32} & 57.11           \\
                                & Hate\textsubscript{Dav}     & 76.04       & 76.81          & \textbf{77.59} & 77.40          & 76.71          & 75.81          & 76.75          & 76.82        & 77.44          & 77.17           & 76.93          & \textbf{78.31}  \\
                                & Hate\textsubscript{Bas}     & 47.85       & 50.28          & \textbf{50.96} & 49.11          & 46.26          & 45.90          & 50.22          & 48.04        & 48.93          & 49.99           & 53.62          & \textbf{53.75}  \\
\multirow{7}{*}{\rotatebox[origin=c]{90}{\textbf{In-Domain}}}   & Humor\textsubscript{Mea}    & 93.28       & 93.30          & 93.46          & 93.55          & 92.21          & 91.81          & 94.07          & 92.51        & \textbf{94.64} & 93.88           & \textbf{94.43} & 94.25           \\
                                & Irony\textsubscript{Hee-A}  & 72.87       & 73.05          & 73.68          & 73.87          & 71.64          & 69.76          & \textbf{77.41} & 72.88        & 76.41          & 75.94           & 77.03          & \textbf{79.51}  \\
                                & Irony\textsubscript{Hee-B}  & 53.20       & 51.12          & 54.75          & 54.76          & 50.70          & 48.68          & 55.38          & 51.84        & \textbf{57.36} & 55.74           & 56.73          & \textbf{58.78}  \\
                                & Offense\textsubscript{Zamp} & 79.93       & 79.81          & 79.20          & \textbf{80.74} & 79.73          & 79.74          & 80.56          & 79.53        & 80.55          & 80.65           & 79.35          & \textbf{79.36}  \\
                                & Sarc\textsubscript{Riloff}  & 73.71       & 70.04          & 72.44          & 74.12          & 68.73          & 67.92          & 75.22          & 70.51        & \textbf{75.90} & 74.51           & 78.76          & \textbf{78.83}  \\
                                & Sarc\textsubscript{Ptacek}  & 95.99       & 95.99          & 96.15          & 95.99          & 95.57          & 95.20          & 96.07          & 95.68        & \textbf{96.19} & 95.98           & 96.40          & \textbf{96.66}  \\
                                & Sarc\textsubscript{Rajad}   & 85.21       & 85.97          & 85.79          & 85.72          & 84.60          & 83.93          & 86.71          & 85.61        & 86.76          & \textbf{86.77}  & 87.13          & \textbf{87.43}  \\
                                & Sarc\textsubscript{Bam}     & 79.79       & 80.32          & 80.84          & 80.09          & 78.95          & 78.31          & \textbf{81.45} & 79.79        & 81.24          & 80.33           & 81.76          & \textbf{83.87}  \\
                                & Senti\textsubscript{Rosen}  & 89.55       & 89.59          & 90.20          & 89.05          & 87.33          & 85.58          & 90.35          & 88.34        & 90.76          & \textbf{90.93}  & 89.53          & \textbf{89.59}  \\
                                & Senti\textsubscript{Thel}   & 71.41       & \textbf{72.19} & 71.72          & 71.81          & 71.12          & 70.66          & \textbf{72.19} & 71.63        & 71.71          & 71.93           & 71.64          & \textbf{71.82}  \\
                                & Stance\textsubscript{Moham} & 69.44       & 69.95          & 70.34          & 69.77          & 65.47          & 64.76          & 70.16          & 68.80        & \textbf{70.87} & 70.73           & \textbf{68.33} & 67.30           \\ \cdashline{2-14}
                                & Average              & 76.24       & 76.31          & 76.86          & 76.77          & 75.04          & 74.25          & 77.46          & 75.80        & \textbf{77.64} & 77.36           & 77.81          & \textbf{78.29}  \\ \midrule
                                & Emotion\textsubscript{Wall} & 66.51       & 66.41          & 67.34          & 65.27          & 63.92          & 62.19          & \textbf{68.37} & 63.45        & 67.78          & 67.74           & 64.48          & \textbf{64.64}  \\
\multirow{7}{*}{\rotatebox[origin=c]{90}{\textbf{Out-of-Domain}}} & Emotion\textsubscript{Dem}  & 56.59       & 56.19          & 56.50          & 56.00          & 56.15          & 56.20          & \textbf{56.68} & 55.78        & 56.24          & 55.76           & 53.33          & \textbf{55.61}  \\
                                & Sarc\textsubscript{Walk}    & 67.50       & 67.90          & \textbf{68.66} & 65.06          & 63.65          & 66.15          & 67.48          & 66.87        & 66.53          & 68.44           & 67.27          & \textbf{67.86}  \\
                                & Sarc\textsubscript{Ora}     & 76.92       & 77.41          & 76.06          & 76.85          & 75.37          & 76.34          & 76.82          & 76.44        & 77.38          & \textbf{77.77}  & \textbf{77.33} & 77.04           \\
                                & Senti-MR                                     & 89.00       & 89.90          & 89.48          & 88.96          & 88.86          & 88.73          & \textbf{90.29} & 88.94        & 90.14          & 90.12           & 87.94          & \textbf{88.06}  \\
                                & Senti-YT                                     & 90.22       & 90.65          & 90.40          & 90.19          & 89.59          & 87.74          & 91.81          & 90.44        & 91.68          & \textbf{92.16}  & 92.25          & \textbf{92.65}  \\
                                & SST-5                                        & 54.96       & 55.92          & 55.52          & 55.69          & 55.00          & 54.35          & 56.26          & 54.18        & 55.40          & \textbf{56.33}  & 55.74          & \textbf{55.97}  \\
                                & SST-2                                        & 94.57       & 94.69          & 94.34          & 94.39          & 93.76          & 93.07          & 94.14          & 94.12        & 94.42          & \textbf{95.15}  & 93.32          & \textbf{93.72}  \\ \cdashline{2-14}
                               & Average              & 74.53       & 74.88          & 74.79          & 74.05          & 73.29          & 73.10          & 75.23          & 73.78        & 74.94          & \textbf{75.43}  & 73.96          & \textbf{74.44}  \\ \bottomrule
\end{tabular}
\caption{Results of using hashtags as distant labels. Models are evaluated on $24$ SM benchmarks. We report average macro-$F_1$ over five runs. \textbf{RB:} Fine-tuning on original pre-trained RoBERTa~\cite{liu2019roberta}; \textbf{MLM:} Further pre-training RoBERTa with MLM objective; \textbf{H-MLM:} Hashtag-based MLM; \textbf{SLP:} Surrogate label prediction; \textbf{Mir-B:} Mirror-BERT~\cite{liu2021fast}; \textbf{Sim-S:} SimCSE-Self~\cite{gao2021simcse}; \textbf{Sim-D:} (Ours) SimCSE-Distant trained with distantly supervised positive pairs and SSCL loss; \textbf{BTw:} BERTweet~\cite{nguyen-etal-2020-bertweet}; \textbf{WCL:} Weakly-supervised contrastive learning~\cite{zheng2021weakly}; \textbf{DCL:} (Ours) Trained with $\mathcal{L}_{DCL}$ only (without MLM and SLP objectives); \textbf{InfoD-R} and \textbf{InfoD-B:} (Ours) continue training RoBERTa and BERTweet, respectively, with proposed InfoDCL framework.}\label{tab:hashtag_result}
\end{table*}

\subsection{Multilingual Tasks}\label{subsec:app_multi-ling}
We also investigate the effectiveness of our proposed model on multilingual tasks. Table~\ref{tab:multi-lingual} shows the performance on nine downstream tasks in three different languages. Here, we continue training XLM-R with our proposed objectives. We experiment with three settings: \textbf{(1)} English only: training on the \texttt{TweetEmoji-1M} and evaluating on the nine multilingual datasets, \textbf{(2)} Target mono-lingual: training on each $1$M mono-lingual tweets in the target language independently (i.e., \texttt{TweetEmoji-AR} for Arabic, \texttt{TweetEmoji-IT} for Italian, and \texttt{TweetEmoji-ES} for Spanish) and evaluating on the respective dataset corresponding to the same language as training data, and \textbf{(3)} Multilingual: training on the \texttt{TweetEmoji-Multi} dataset and evaluating on the nine multilingual datasets. We still use the NPMI weighting matrix generated from English tweets in these experiments.~\footnote{We plan to explore generating the NPMI weighting matrix from mutlilingual data in future work.} Table~\ref{tab:multi-lingual} shows that our models outperform the original XLM-R on all the datasets and obtains improvements of $1.44$ and $0.85$ average $F_1$ across the nine datasets under the multilingual and target mono-lingual settings, respectively. Training on English mono-lingual data helps four datasets, but cannot benefit all the nine non-English datasets on average. Compared to previous SoTA models, our proposed methods outperform these on six out of nine datasets.~\footnote{For Emo-es\textsubscript{Moham}, we use fine-tuning XLM-R as SoTA model because we convert the intensity regression task to a emotion classification and there is no SoTA model.} These results demonstrate that our methods are not only task-agnostic within the realm of SM tasks, but also language-independent.

\subsection{Using Hashtag as Distant Supervision}\label{sec:hashtag}
As Table~\ref{tab:hashtag_result} presents, our proposed framework also can enhance the representation quality using hashtags as distantly supervised labels. InfoDCL-RoBERTa, the model further training RoBERTa on the training set of \texttt{TweetHashtag-EN} with our framework, obtains average $F_1$ of $77.36$ and $75.43$ across the $16$ in-domain and eight out-of-domain datasets, respectively. Comapred to baselines, our DCL obtains the best performance average $F_1$ score across $16$ in-domain datasets ( $F_1 = 77.64$). InfoDCL-BERTweet, the further pre-trained BERTweet on the training set of \texttt{TweetHashtag-EN} with our framework, obtains average $F_1$ of $78.29$ and $74.44$ across the $16$ in-domain and eight out-of-domain datasets, respectively.   

\subsection{Topic Classification}\label{subsec:append_topic}
We fine-tune baselines and our models on two topic classification datasets and report macro $F_1$ scores in Table~\ref{tab:topic_res}. We find that our hashtag-based InfoDCL model acquires best performance on both datasets, for AGNews $F_1=97.42$, and for Topic\textsubscript{Dao} $F_1=94.80$. These results indicate that our framework can also effectively improve topic classification when we use hashtags as distant labels. 
\begin{table}[!ht]
\centering
\tiny
\begin{tabular}{@{}lcc:c|lcc:c@{}}
\toprule
\multicolumn{4}{c}{\textbf{Emoji-based}}                                                                                & \multicolumn{4}{c}{\textbf{Hashtag-based}}                                                                              \\ \cmidrule(l){1-4} \cmidrule(l){5-8} 
\multicolumn{1}{c}{\textbf{Model}} & \textbf{AGN} & \textbf{Topic} & \textbf{Ave}   & \multicolumn{1}{c}{\textbf{Model}} & \textbf{AGN} & \textbf{Topic} & \textbf{Ave}   \\ \midrule
RB                            & 96.97          & \textbf{94.75}                                   & 95.86          & -                            & -          & -                                            & -          \\
MLM                                & 97.00          & 94.58                                            & 95.79          & MLM                                & 97.01          & 94.78                                            & 95.89          \\
E-MLM                              & 96.97          & 94.73                                            & 95.85          & E-MLM                              & 97.13          & 94.66                                            & 95.90          \\
SLP                                & 97.12          & 94.54                                            & 95.83          & SLP                                & 97.04          & 94.63                                            & 95.84          \\
Mir-B                        & 96.86          & 94.72                                            & 95.79          & Mir-B                        & 97.13          & 94.66                                            & 95.90          \\
Sim-S                        & 96.88          & 94.73                                            & 95.81          & Sim-S                        & 96.90          & 94.65                                            & 95.78          \\
Sim-D                              & 97.08          & 94.70                                            & \textbf{95.89} & Sim-D                              & 97.30          & 94.79                                            & 96.04          \\
WCL                                & \textbf{97.13} & 94.65                                            & \textbf{95.89} & WCL                                & 97.09          & 94.56                                            & 95.83          \\
DCL                                & 97.08          & 94.59                                            & 95.84          & DCL                                & 97.23          & 94.64                                            & 95.93          \\
InfoD-R                            & 97.01          & 94.48                                            & 95.74          & InfoD-R                            & \textbf{97.42} & \textbf{94.80}                                   & \textbf{96.11} \\ \hline
BTw                           & 97.00          & 94.43                                            & 95.72          & -                           & -          & -                                            & -         \\
InfoD-B                           & \textbf{97.05}          & \textbf{94.47}                                            & \textbf{95.76}          & InfoD-B                           & \textbf{97.26 }         & \textbf{94.49}                                            & \textbf{95.87}          \\ \bottomrule
\end{tabular}
\caption{Results on topic classification. We report macro average $F_1$ over five runs. \textbf{Dataset:} \textbf{AGN:} AGNews, \textbf{Topic:} Topic\textsubscript{Dao}. }\label{tab:topic_res}
\end{table}

\begin{table*}[!ht]
\scriptsize
\centering
\begin{tabular}{@{}lcccccccccc|cc@{}}
\toprule
\multicolumn{1}{c}{\textbf{Task}} & \textbf{RB} & \textbf{MLM} & \textbf{E-MLM} & \textbf{SLP}   & \textbf{Mir-B} & \textbf{Sim-S} & \textbf{Sim-D} & \textbf{WCL} & \textbf{DCL}   & \textbf{InfoD-R} & \textbf{BTw}   & \textbf{InfoD-B} \\ \midrule
STS12                             & 15.88       & 37.71        & 34.55          & 50.07          & \textbf{59.07} & 54.18          & 46.13          & 34.81        & 46.46          & 48.13           & 29.20          & \textbf{42.54}  \\
STS13                             & 38.11       & 55.72        & 53.90          & 53.87          & \textbf{69.89} & 65.06          & 45.99          & 37.56        & 47.24          & 51.44           & 36.26          & \textbf{44.40}  \\
STS14                             & 28.58       & 40.16        & 40.86          & 44.88          & \textbf{63.82} & 59.18          & 43.20          & 24.51        & 42.76          & 46.79           & 33.76          & \textbf{38.95}  \\
STS15                             & 40.22       & 59.49        & 56.35          & 61.83          & \textbf{73.78} & 70.30          & 52.76          & 50.36        & 49.11          & 58.04           & 49.19          & \textbf{54.67}  \\
STS16                             & 50.12       & 62.13        & 65.12          & 58.41          & \textbf{74.20} & 70.45          & 51.17          & 36.33        & 45.39          & 57.09           & 46.99          & \textbf{49.42}  \\
SICK-R                            & 62.54       & 64.42        & 63.48          & 64.21          & \textbf{64.29} & 63.53          & 57.14          & 47.22        & 56.93          & 62.81           & 48.76          & \textbf{59.15}  \\
STS-B                             & 46.63       & 56.00        & 58.50          & 59.93          & \textbf{68.75} & 64.49          & 53.00          & 42.24        & 50.64          & 56.65           & 38.24          & \textbf{52.46}  \\  \cdashline{1-13}
\textbf{Average}                  & 40.30       & 53.66        & 53.25          & 56.17          & \textbf{67.69} & 63.88          & 49.91          & 39.00        & 48.36          & 54.42           & 40.34          & \textbf{48.80}  \\ \midrule
MR                                & 75.92       & 76.85        & 80.62          & 86.79          & 76.72          & 73.77          & 86.04          & 78.96        & \textbf{86.83} & 86.66           & 79.58          & \textbf{86.12}  \\
CR                                & 69.59       & 77.35        & 84.79          & 89.69          & 81.48          & 80.19          & 89.48          & 83.74        & \textbf{90.36} & 89.75           & 80.82          & \textbf{89.62}  \\
SUBJ                              & 91.50       & 90.63        & 91.01          & 92.24          & 91.57          & 90.29          & 91.24          & 92.91        & 92.61          & \textbf{93.71}  & 93.03          & \textbf{93.53}  \\
MPQA                              & 73.75       & 80.40        & 78.54          & \textbf{87.93} & 85.39          & 83.92          & 87.18          & 85.30        & 87.51          & 87.12           & 71.78          & \textbf{86.21}  \\
SST2                              & 82.81       & 85.50        & 88.14          & 92.53          & 81.05          & 78.69          & 91.87          & 85.28        & 91.43          & \textbf{92.59}  & 86.66          & \textbf{91.10}  \\
SST5                              & 38.46       & 41.81        & 46.65          & 52.31          & 44.48          & 41.45          & 48.60          & 43.48        & 50.77          & \textbf{53.08}  & 43.71          & \textbf{52.13}  \\
TREC                              & 61.40       & 73.20        & 72.20          & 78.60          & \textbf{87.00} & 86.00          & 74.60          & 84.20        & 75.80          & 83.00           & 80.80          & \textbf{83.40}  \\
MRPC                              & 71.42       & 73.04        & 74.09          & 74.61          & \textbf{74.67} & 74.49          & 71.59          & 71.88        & 71.54          & 73.22           & \textbf{72.35} & 72.00           \\ \cdashline{1-13}
\textbf{Average}                  & 70.61       & 74.85        & 77.01          & 81.84          & 77.80          & 76.10          & 80.08          & 78.22        & 80.86          & \textbf{82.39}  & 76.09          & \textbf{81.76}  \\ \bottomrule
\end{tabular}
\caption{Evaluate on SentEval benchmark. All the models are pre-trained on \texttt{TweetEmoji-EN}. For STS task, we report the Spearman’s correlation, ``all" setting. For transferring tasks, we report accuracy. }\label{tab:senteval_emoji}
\end{table*}
\subsection{SentEval}\label{subsec:append_senteval}
Each STS dataset includes pairs of sentences each with a gold semantic similarity score ranging from $0$ to $5$. We encode each sentence by the hidden state of `[CLS]' token from the last Transformer encoder layer. We then calculate the Spearman's correlation between cosine similarity of sentence embeddings and the gold similarity score of each pair. Same as Mirror-BERT~\cite{liu2021fast} and SimCSE~\cite{gao2021simcse}, we report the overall Spearman's correlation. For transfer learning tasks, we follow the evaluation protocal of SentEval, where a trainable logistic regression classifier is added on top of a frozen encoder that is an PLM. We report classification accuracy of eight transfer learning datasets in Tables~\ref{tab:senteval_emoji}. Although our InfoDCL underperforms Mirror-BERT on all STS datasets, but it still outperforms than Baseline 1, 2, and 3. Our InfoDCL is not designed to improve STS task but it does not hurt performance compared to Baseline 2. Moreover, our InfoDCL achieves the best average performance on eight transferring datasets. We note that four datasets are SM tasks. Only regarding the other four non-SM tasks, our InfoDCL model still outperforms most baselines and achieves the second best performance on average, which is only $0.40$ $F_1$ points lower than Mirror-BERT.

\begin{figure}[ht]
\centering
\tiny
\begin{subfigure}[]{.24\textwidth}
  \centering
  \includegraphics[width=\linewidth]{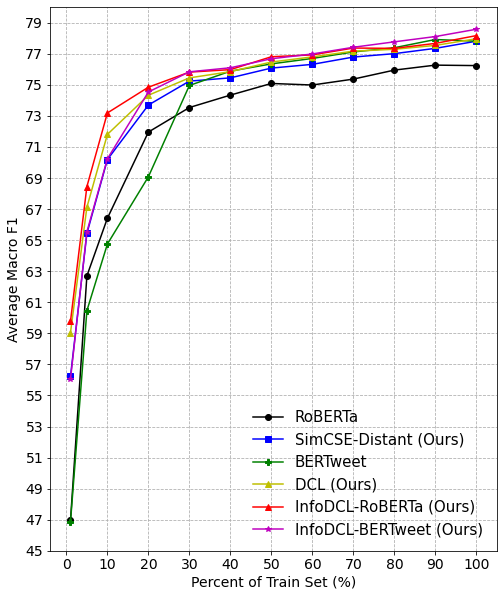}
  \caption{In-domain}
  \label{fig:few-indomain}
\end{subfigure}%
\begin{subfigure}[]{.24\textwidth}
  \centering
  \includegraphics[width=\linewidth]{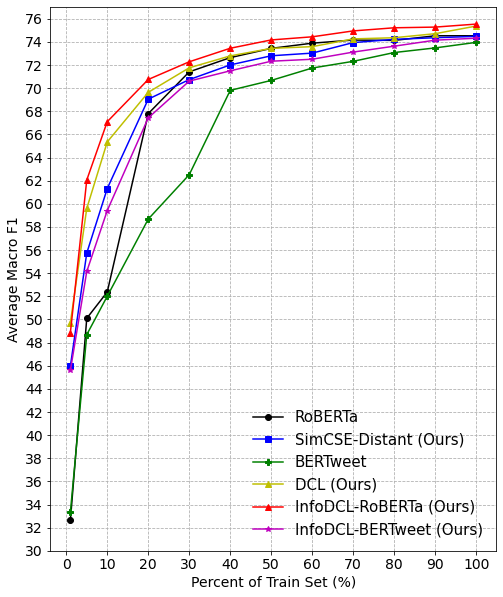}
  \caption{Out-of-domain}
  \label{fig:few-outofdomain}
\end{subfigure}
\caption{Few-shot learning on downstream tasks where we use varying percentages of Train sets. The $y$-axis indicates the average Test macro $F_1$ across $16$ Twitter and eight out-of-domain benchmarks. The $x$-axis indicates the percentage of Train set used to fine-tune the model. }\label{fig:data_percent}
\end{figure} 

\subsection{Few Shot Learning}\label{subsec:append_fewshot}

Since InfoDCL exploits an extensive set of cues in the data that capture a broad range of fine-grained SM concepts, we hypothesize it will be also effective in few-shot learning. Hence, we test this hypothesis for both in-domain and out-of-domain tasks. Figure~\ref{fig:data_percent} and Table~\ref{tab:few_shot-percent} compare our models to three strong baselines when they are trained with different percentages of training samples. Results show that our proposed InfoDCL model always outperforms all baselines on average $F_1$ scores across both in-domain and out-of-domain tasks. For $16$ in-domain tasks, our InfoDCL-RoBERTa remarkably surpasses the RoBERTa baseline with a sizable $12.82$ average $F_1$ scores when we only provide $1\%$ training data from downstream tasks. Compared to other strong baselines, fine-tuning BERTweet and SimCSE-Distant (also our method), InfoDCL-RoBERTa outperforms these with $12.91$ and $3.55$ average $F_1$ scores, respectively, when we use $1\%$ training data for downstream fine-tuning. With only $5\%$ of gold data, InfoDCL-RoBERTa improves $5.76$ points over the RoBERTa baseline. 
For eight out-of-domain tasks, InfoDCL-RoBERTa outperforms the RoBERTa, BERTweet, and SimCSE-Distant baselines with $16.23$, $15.52$, and $2.89$ average $F_1$ scores, respectively, when the models are only fine-tuned on $1\%$ training data of downstream tasks. As Figure~\ref{fig:few-outofdomain} and Table~\ref{tab:few_shot-percent} show, InfoDCL-RoBERTa consistently outperforms all the baselines given \textit{any} percentage of training data. Tables~\ref{tab:few_shot-roberta}, \ref{tab:few_shot-bertweet}, \ref{tab:few_shot-simcse-distant}, \ref{tab:few_shot-dcl}, \ref{tab:few_shot-infodcl-roberta},  and~\ref{tab:few_shot-infodcl-bertweet}, respectively, present the performance of RoBERTa, BERTweet, SimCSE-Distant, DCL, InfoDCL-RoBERTa, InfoDCL-BERTweet on all our $24$ English downstream datasets and various few-shot settings.

\section{Analyses}

\subsection{Model Analysis}
Table~\ref{tab:emoji_sim} shows that both PMI and EC-Emb are capable of capturing sensible correlations between emojis (although the embedding approach includes a few semantically distant emojis, such as the emoji `\loudlycryingface' being highly related to `\grinningfacewithsmilingeyes').

\begin{table*}[ht]
\centering
\setlength\tabcolsep{10pt}
\begin{tabular}{@{}clcccccccccc@{}}
\toprule
\textbf{Q}  & \textbf{Method}    & \textbf{1} & \textbf{2} & \textbf{3} & \textbf{4} & \textbf{5} & \textbf{6} & \textbf{7} & \textbf{8} & \textbf{9} & \textbf{10} \\ \midrule
\multirow{2}{*}{\grinningfacewithsmilingeyes } & PMI     & \begin{tabular}[c]{@{}c@{}}\smilingfacewithsmilingeyes\\ .11\end{tabular}         & \begin{tabular}[c]{@{}c@{}}\grinningfacewithbigeyes \\ .11\end{tabular}         & \begin{tabular}[c]{@{}c@{}}\grinningface \\ .10\end{tabular}         & \begin{tabular}[c]{@{}c@{}}\thumbsup \\ .10\end{tabular}         & \begin{tabular}[c]{@{}c@{}}\confettiball \\ .10\end{tabular}         & \begin{tabular}[c]{@{}c@{}}\partypopper\\ .10\end{tabular}         & \begin{tabular}[c]{@{}c@{}}\squintingfacewithtongue \\ .10\end{tabular}         & \begin{tabular}[c]{@{}c@{}}\winkingfacewithtongue \\ .09\end{tabular}         & \begin{tabular}[c]{@{}c@{}}\wrappedgift \\ .09\end{tabular}         & \begin{tabular}[c]{@{}c@{}}\grinningsquintingface \\ .07\end{tabular}          \\
                    & E-em & \begin{tabular}[c]{@{}c@{}}\grimacingface \\ .34\end{tabular}         & \begin{tabular}[c]{@{}c@{}}\smilingfacewithsmilingeyes \\ .32\end{tabular}          & \begin{tabular}[c]{@{}c@{}}\grinningsquintingface \\ .31\end{tabular}         & \begin{tabular}[c]{@{}c@{}}\grinningfacewithbigeyes \\ .28\end{tabular}         & \begin{tabular}[c]{@{}c@{}}\facewithtearsofjoy \\ .28\end{tabular}         & \begin{tabular}[c]{@{}c@{}}\grinningfacewithsweat \\ .28\end{tabular}         & \begin{tabular}[c]{@{}c@{}}\loudlycryingface \\ .28\end{tabular}         & \begin{tabular}[c]{@{}c@{}}\relievedface \\ .27\end{tabular}         & \begin{tabular}[c]{@{}c@{}}\grinningface \\ .27\end{tabular}         & \begin{tabular}[c]{@{}c@{}}\squintingfacewithtongue \\ .26\end{tabular} \\ \midrule         
\multirow{2}{*}{\flaggermany} & PMI     & \begin{tabular}[c]{@{}c@{}}\flagchina \\ .67\end{tabular}         & \begin{tabular}[c]{@{}c@{}}\flagrussia \\ .67\end{tabular}         & \begin{tabular}[c]{@{}c@{}}\flagsouthkorea \\ .66\end{tabular}         & \begin{tabular}[c]{@{}c@{}}\flagspain \\ .66\end{tabular}         & \begin{tabular}[c]{@{}c@{}}\flagfrance \\ .62\end{tabular}         & \begin{tabular}[c]{@{}c@{}}\flagjapan\\ .62\end{tabular}         & \begin{tabular}[c]{@{}c@{}}\flagitaly \\ .61\end{tabular}         & \begin{tabular}[c]{@{}c@{}}\flagunitedkingdom \\ .55\end{tabular}         & \begin{tabular}[c]{@{}c@{}}\flagportugal \\ .54\end{tabular}         & \begin{tabular}[c]{@{}c@{}}\flagaustralia \\ .46\end{tabular}          \\
                  & E-em & \begin{tabular}[c]{@{}c@{}}\flagrussia \\ .36\end{tabular}         & \begin{tabular}[c]{@{}c@{}}\flagireland \\ .36\end{tabular}         & \begin{tabular}[c]{@{}c@{}}\flagunitedkingdom \\ .36\end{tabular}         & \begin{tabular}[c]{@{}c@{}}\flagaustralia \\ .36\end{tabular}         & \begin{tabular}[c]{@{}c@{}}\flagitaly \\ .36\end{tabular}         & \begin{tabular}[c]{@{}c@{}}\shamrock\\ .35\end{tabular}         & \begin{tabular}[c]{@{}c@{}}\flaginhole \\ .35\end{tabular}         & \begin{tabular}[c]{@{}c@{}}\icehockey \\ .34\end{tabular}         & \begin{tabular}[c]{@{}c@{}}\flagpakistan \\ .34\end{tabular}         & \begin{tabular}[c]{@{}c@{}}\flagphilippines \\ .33\end{tabular} \\ \midrule
\multirow{2}{*}{\cloudwithsnow} & PMI     & \begin{tabular}[c]{@{}c@{}}\snowmanwithoutsnow \\ .65\end{tabular}         & \begin{tabular}[c]{@{}c@{}}\windface\\ .53\end{tabular}         & \begin{tabular}[c]{@{}c@{}}\snowflake\\ .53\end{tabular}         & \begin{tabular}[c]{@{}c@{}}\sunbehindraincloud \\ .52\end{tabular}         & \begin{tabular}[c]{@{}c@{}}\cloudwithlightning\\ .52\end{tabular}         & \begin{tabular}[c]{@{}c@{}}\fog \\ .50\end{tabular}         & \begin{tabular}[c]{@{}c@{}}\cloudwithrain \\ .49\end{tabular}         & \begin{tabular}[c]{@{}c@{}}\sunbehindlargecloud \\ .45\end{tabular}         & \begin{tabular}[c]{@{}c@{}}\cloudwithlightningandrain \\ .45\end{tabular}         & \begin{tabular}[c]{@{}c@{}}\tornado \\ .43\end{tabular}          \\
                    & E-em & \begin{tabular}[c]{@{}c@{}}\cloudwithrain \\ .36\end{tabular}          & \begin{tabular}[c]{@{}c@{}}\foggy \\ .34\end{tabular}         & \begin{tabular}[c]{@{}c@{}}\snowflake \\ .34\end{tabular}          & \begin{tabular}[c]{@{}c@{}}\helicopter \\ .34\end{tabular}          & \begin{tabular}[c]{@{}c@{}}\snowmanwithoutsnow \\ .34\end{tabular}          & \begin{tabular}[c]{@{}c@{}}\cloudwithlightningandrain \\ .32\end{tabular}          & \begin{tabular}[c]{@{}c@{}}\closedumbrella \\ .32\end{tabular}          & \begin{tabular}[c]{@{}c@{}}\mountain \\ .32\end{tabular}         & \begin{tabular}[c]{@{}c@{}}\sunbehindlargecloud \\ .32\end{tabular}         & \begin{tabular}[c]{@{}c@{}}\bus \\ .32\end{tabular} 
                \\ \bottomrule
\end{tabular}
\caption{Ranking of emoji similarity by different methods. \textbf{PMI} is normalized point-wise mutual information. \textbf{E-em:} EC-Emb is the cosine similarity between class embeddings. Emojis are ranked by the similarity scores (under emojis) between them and the query. \textbf{Q:} Query emoji.}\label{tab:emoji_sim}
\end{table*}

\subsection{Qualitative Analysis}\label{subsec:app_qualitative}
\begin{figure*}[ht]
\centering
\includegraphics[width=\linewidth]{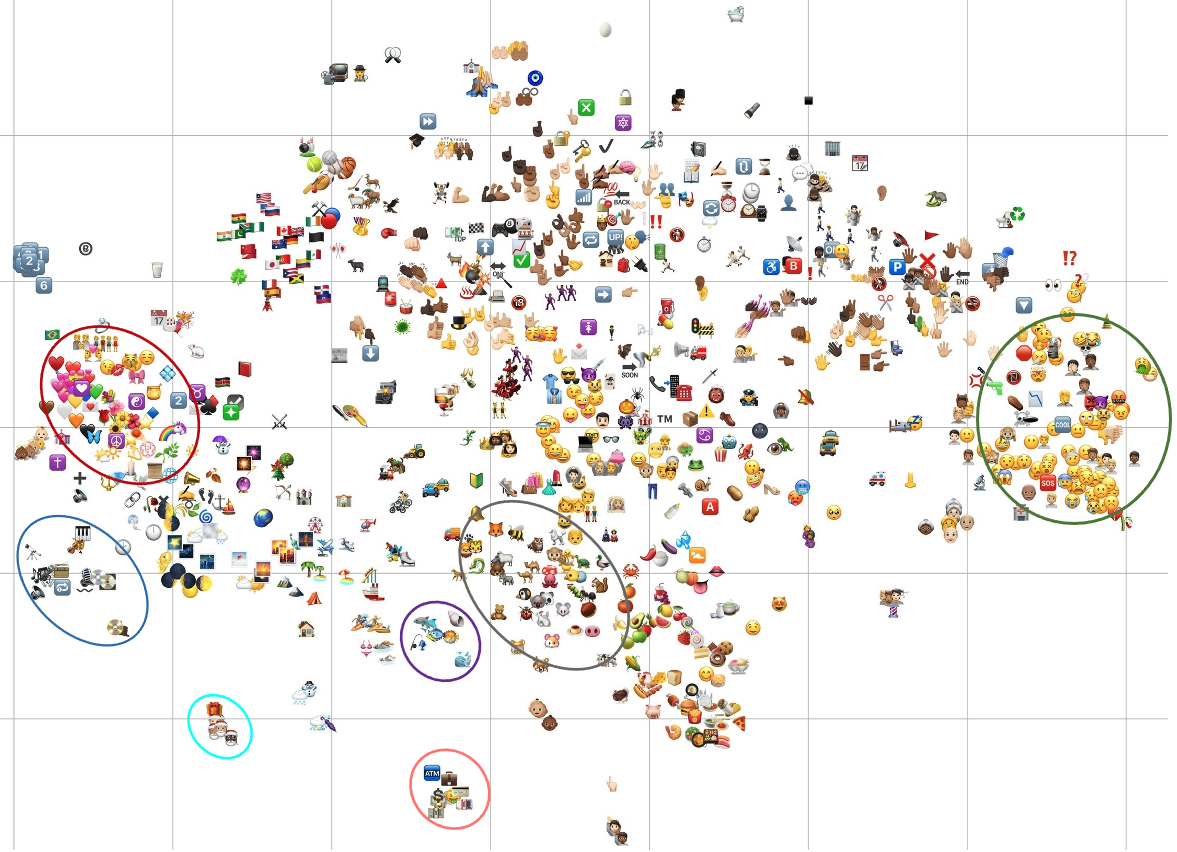}

\caption{Visualizing emojis in two-dimensional space using $t$-SNE. We can clearly observe some clusters of similar emojis, such as love and marriage (in red circle), music (in blue circle), money (in orange circle), unhappiness (in green circle), Christmas (in cyan circle).}\label{fig:emoji-t-sne}
\end{figure*}

We provide a qualitative visualization analysis of our model representation. For this purpose, we use our InfoDCL-RoBERTa to obtain representations of samples in the \texttt{TweetEmoji-EN}'s validation set (`[CLS]' token from the last encoder layer) then average the representations of all tweets with the same surrogate label (emoji). We then project these emoji embeddings into a two-dimensional space using $t$-SNE. As Fig.~\ref{fig:emoji-t-sne} shows, we can observe a number of distinguishable clusters. For instance, a cluster of love and marriage is grouped in the left region, unhappy and angry faces are in the right side, and food at the bottom. We can also observe sensible relations between clusters. For instance, the cluster of love and marriage is close to the cluster of smiling faces but is far away from the cluster of unhappy faces. In addition, the cluster of aquatic animals (middle bottom) is close to terrestrial animals while each of these is still visually distinguishable. We also note that emojis which contain the same emoji character but differ in skin tone are clustered together. An example of these is emojis of Santa Claus (left bottom). This indicates that our InfoDCL model has meticulously captured the relations between the emoji surrogate labels. 

\section{Uniformity and Tolerance}\label{sec:uni-tole}
\citet{wang-2021-understanding} investigate representation quality measuring the uniformity of an embedding distribution and the tolerance to semantically similar samples. Given a dataset $D$ and an encoder $\Phi$, the uniformity is based on a gaussian potential kernel introduced by~\newcite{wang-2020-understanding} and is formulated as:
\begin{equation}
    \mathcal{L}_{uniformity} = log \mathop{\mathbb{E}}_{x_i, x_j\in D}[ e^{ -t ||\Phi(x_i) - \Phi(x_j)||^2_2 } ], 
\end{equation}
where $t=2$. \citet{wang-2021-understanding} use $-\mathcal{L}_{uniformity}$ as the uniformity metric, thus a higher uniformity score indicates that the embedding distribution is closer to a uniform distribution.

The tolerance metric measures the mean of similarities of samples belonging to the same class, which defined as:
\begin{equation}
\small
    Tolerance = \mathop{\mathbb{E}}_{x_i, x_j\in D}[(\Phi(x_i)^T\Phi(x_j)) \cdot I_{l(x_i)=l(x_j)}],
\end{equation}

where $l(x_i)$ is the supervised label of sample $x_i$. $I_{l(x_i)=l(x_j)}$ is an indicator function, giving the value of $1$ for $l(x_i)=l(x_j)$ and the value of $0$ for $l(x_i)\neq l(x_j)$. In our experiments, we use gold development samples from our downstream SM datasets.

\begin{table*}[h]
\centering
\small
\begin{tabular}{@{}lcccccccccc@{}}
\toprule
\textbf{Task}                                & \textbf{InfoDCL} & \textbf{A} & \textbf{B} & \textbf{C} & \textbf{D} & \textbf{E} & \textbf{F} & \textbf{G} & \textbf{H} & \textbf{I} \\ \midrule
Crisis\textsubscript{Oltea} & 96.01            & 95.91      & 95.88      & 95.91      & 95.83      & 95.96      & 95.92      & 95.75      & 95.96      & 95.79      \\
Emo\textsubscript{Moham}    & 81.34            & 82.31      & 82.03      & 80.98      & 80.06      & 81.28      & 80.54      & 81.27      & 82.11      & 81.49      \\
Hate\textsubscript{Was}     & 57.30            & 57.13      & 57.09      & 57.03      & 57.30      & 57.24      & 57.14      & 56.89      & 57.08      & 57.12      \\
Hate\textsubscript{Dav}     & 77.29            & 76.82      & 77.88      & 77.59      & 76.74      & 76.11      & 76.79      & 77.69      & 77.40      & 77.15      \\
Hate\textsubscript{Bas}     & 52.84            & 51.77      & 52.39      & 51.90      & 52.79      & 51.26      & 52.17      & 51.67      & 53.63      & 50.97      \\
Humor\textsubscript{Mea}    & 93.75            & 93.08      & 93.62      & 93.17      & 94.23      & 93.64      & 94.13      & 93.26      & 93.87      & 93.78      \\
Irony\textsubscript{Hee-A}  & 76.31            & 76.41      & 77.14      & 77.11      & 74.99      & 78.19      & 77.15      & 76.95      & 76.55      & 76.18      \\
\bottomrule
\end{tabular}
\caption{Full results of ablation studies. , \textbf{A:} without CCL, \textbf{B:} without LCL, \textbf{C:} without LCL \& CCL, \textbf{D:} without SLP, \textbf{E:} without MLM, \textbf{F:} without SLP \& MLM (i.e., DCL), \textbf{G:} without epoch-wise re-pairing, \textbf{H:} with additional weighting model, \textbf{I:} InfoDCL+Self data augmentation. } \label{tab:ablation_full}
\end{table*}

\begin{table*}[h]
\centering
\small
\begin{tabular}{@{}lllllllllllll@{}}
\toprule
\multicolumn{1}{c}{\textbf{Percentage}} & \multicolumn{1}{c}{\textbf{1}} & \multicolumn{1}{c}{\textbf{5}} & \multicolumn{1}{c}{\textbf{10}} & \multicolumn{1}{c}{\textbf{20}} & \multicolumn{1}{c}{\textbf{30}} & \multicolumn{1}{c}{\textbf{40}} & \multicolumn{1}{c}{\textbf{50}} & \multicolumn{1}{c}{\textbf{60}} & \multicolumn{1}{c}{\textbf{70}} & \multicolumn{1}{c}{\textbf{80}} & \multicolumn{1}{c}{\textbf{90}} & \multicolumn{1}{c}{\textbf{100}} \\ \midrule
\multicolumn{13}{c}{\textbf{In-Domain}}                                                                                                                                                                                                                                                                                                                                                                                                                        \\ \midrule
RoBERTa                                 & 46.96                          & 62.70                          & 66.41                           & 71.96                           & 73.54                           & 74.34                           & 75.09                           & 74.99                           & 75.37                           & 75.95                           & 76.27                           & 76.24                            \\
BERTweet                                & 46.87                          & 60.46                          & 64.75                           & 69.08                           & 74.96                           & 75.88                           & 76.35                           & 76.70                           & 77.12                           & 77.39                           & 77.92                           & 77.81                            \\
\bottomrule
\end{tabular}
\caption{Few-shot learning on downstream tasks where we use varying percentages of Train sets. We report the averaged Test macro-$F_1$ score across $16$ in-domain tasks and eight out-of-domain tasks, respectively. \textbf{Sim-D:} SimCSE-Distant, \textbf{RB:} RoBERTa, \textbf{BTw:} BERTweet.}\label{tab:few_shot-percent}
\end{table*}

\begin{table*}[h]
\centering
\tiny     
\begin{tabular}{@{}lrrrrrrrrrrrr|rrrr@{}}
\toprule
\multicolumn{1}{c}{}                         & \multicolumn{12}{c|}{\textbf{Percentage}}                                                                                                                                                                                                                                                                                                                                                                             & \multicolumn{4}{c}{\textbf{\# of Training Samples}}                                                                                       \\ \cmidrule(r){2-13} \cmidrule(r){14-17} 
\multicolumn{1}{c}{\textbf{}}                & \multicolumn{1}{c}{\textbf{1}} & \multicolumn{1}{c}{\textbf{5}} & \multicolumn{1}{c}{\textbf{10}} & \multicolumn{1}{c}{\textbf{20}} & \multicolumn{1}{c}{\textbf{30}} & \multicolumn{1}{c}{\textbf{40}} & \multicolumn{1}{c}{\textbf{50}} & \multicolumn{1}{c}{\textbf{60}} & \multicolumn{1}{c}{\textbf{70}} & \multicolumn{1}{c}{\textbf{80}} & \multicolumn{1}{c}{\textbf{90}} & \multicolumn{1}{c|}{\textbf{100}} & \multicolumn{1}{c}{\textbf{20}} & \multicolumn{1}{c}{\textbf{100}} & \multicolumn{1}{c}{\textbf{500}} & \multicolumn{1}{c}{\textbf{1000}} \\ \midrule
Crisis\textsubscript{Oltea} & 94.88                          & 95.18                          & 95.59                           & 95.67                           & 95.73                           & 95.65                           & 95.88                           & 95.77                           & 95.72                           & 95.83                           & 95.92                           & 95.87                            & 37.20                           & 70.27                            & 95.09                            & 95.20                             \\
\bottomrule
\end{tabular}
\caption{Full results of few-shot learning on Baseline (1), fine-tuning RoBERTa.}\label{tab:few_shot-roberta}
\end{table*}

\begin{table*}[h]
\centering
\tiny     
\begin{tabular}{@{}lrrrrrrrrrrrr|rrrr@{}}
\toprule
\multicolumn{1}{c}{}                         & \multicolumn{12}{c|}{\textbf{Percentage}}                                                                                                                                                                                                                                                                                                                                                                             & \multicolumn{4}{c}{\textbf{\# of Training Samples}}                                                                                       \\ \cmidrule(r){2-13} \cmidrule(r){14-17} 
\multicolumn{1}{c}{\textbf{}}                & \multicolumn{1}{c}{\textbf{1}} & \multicolumn{1}{c}{\textbf{5}} & \multicolumn{1}{c}{\textbf{10}} & \multicolumn{1}{c}{\textbf{20}} & \multicolumn{1}{c}{\textbf{30}} & \multicolumn{1}{c}{\textbf{40}} & \multicolumn{1}{c}{\textbf{50}} & \multicolumn{1}{c}{\textbf{60}} & \multicolumn{1}{c}{\textbf{70}} & \multicolumn{1}{c}{\textbf{80}} & \multicolumn{1}{c}{\textbf{90}} & \multicolumn{1}{c|}{\textbf{100}} & \multicolumn{1}{c}{\textbf{20}} & \multicolumn{1}{c}{\textbf{100}} & \multicolumn{1}{c}{\textbf{500}} & \multicolumn{1}{c}{\textbf{1000}} \\ \midrule
Crisis\textsubscript{Oltea} & 93.47 & 95.07 & 95.42 & 95.40 & 95.53 & 95.55 & 95.59 & 95.79 & 95.63 & 95.76 & 95.68 & 95.76 & 50.00     & 46.96    & 94.65    & 95.02    \\
Emo\textsubscript{Moham}    & 20.49 & 18.15 & 53.35 & 73.16 & 76.84 & 76.95 & 78.00 & 78.55 & 78.88 & 79.54 & 79.94 & 80.23 & 20.02     & 16.86    & 70.70    & 75.59    \\
Hate\textsubscript{Was}     & 28.22 & 51.43 & 53.03 & 54.95 & 55.62 & 55.54 & 56.26 & 56.46 & 56.39 & 56.26 & 56.91 & 57.32 & 29.25     & 28.22    & 51.29    & 53.60    \\
Hate\textsubscript{Dav}     & 28.86 & 68.38 & 73.29 & 75.37 & 76.60 & 76.12 & 77.32 & 76.39 & 76.77 & 77.07 & 76.90 & 76.93 & 31.20     & 30.34    & 57.32    & 67.51    \\
\bottomrule
\end{tabular}
\caption{Full results of few-shot learning on Baseline (11), fine-tuning BERTweet.}\label{tab:few_shot-bertweet}
\end{table*}

\begin{table*}[h]
\centering
\tiny     
\begin{tabular}{@{}lrrrrrrrrrrrr|rrrr@{}}
\toprule
\multicolumn{1}{c}{}                         & \multicolumn{12}{c|}{\textbf{Percentage}}                                                                                                                                                                                                                                                                                                                                                                             & \multicolumn{4}{c}{\textbf{\# of Training Samples}}                                                                                       \\ \cmidrule(r){2-13} \cmidrule(r){14-17} 
\multicolumn{1}{c}{\textbf{}}                & \multicolumn{1}{c}{\textbf{1}} & \multicolumn{1}{c}{\textbf{5}} & \multicolumn{1}{c}{\textbf{10}} & \multicolumn{1}{c}{\textbf{20}} & \multicolumn{1}{c}{\textbf{30}} & \multicolumn{1}{c}{\textbf{40}} & \multicolumn{1}{c}{\textbf{50}} & \multicolumn{1}{c}{\textbf{60}} & \multicolumn{1}{c}{\textbf{70}} & \multicolumn{1}{c}{\textbf{80}} & \multicolumn{1}{c}{\textbf{90}} & \multicolumn{1}{c|}{\textbf{100}} & \multicolumn{1}{c}{\textbf{20}} & \multicolumn{1}{c}{\textbf{100}} & \multicolumn{1}{c}{\textbf{500}} & \multicolumn{1}{c}{\textbf{1000}} \\ \midrule
Crisis\textsubscript{Oltea} & 94.21 & 94.94 & 95.28 & 95.53 & 95.69 & 95.72 & 95.76 & 95.81 & 95.89 & 95.96 & 95.86 & 95.94 & 61.80     & 90.88    & 94.31    & 94.63    \\
Emo\textsubscript{Moham}    & 24.31 & 53.06 & 75.65 & 77.15 & 78.46 & 78.53 & 78.77 & 79.68 & 80.17 & 79.75 & 81.00 & 81.05 & 23.47     & 41.68    & 76.72    & 78.35    \\
Hate\textsubscript{Was}     & 32.05 & 51.26 & 53.38 & 54.94 & 55.47 & 56.18 & 55.99 & 56.46 & 56.80 & 57.06 & 57.29 & 57.13 & 34.03     & 32.73    & 51.66    & 53.03    \\
Hate\textsubscript{Dav}     & 38.33 & 71.56 & 73.83 & 74.42 & 76.12 & 75.36 & 76.50 & 76.98 & 76.96 & 75.93 & 77.81 & 77.15 & 34.57     & 34.33    & 66.42    & 70.04    \\
\bottomrule
\end{tabular}
\caption{Full results of few-shot learning on SimCSE-Distant.}\label{tab:few_shot-simcse-distant}
\end{table*}

\begin{table*}[h]
\centering
\tiny   
\begin{tabular}{@{}lrrrrrrrrrrrr|rrrr@{}}
\toprule
\multicolumn{1}{c}{}                         & \multicolumn{12}{c|}{\textbf{Percentage}}                                                                                                                                                                                                                                                                                                                                                                             & \multicolumn{4}{c}{\textbf{\# of Training Samples}}                                                                                       \\ \cmidrule(r){2-13} \cmidrule(r){14-17} 
\multicolumn{1}{c}{\textbf{}}                & \multicolumn{1}{c}{\textbf{1}} & \multicolumn{1}{c}{\textbf{5}} & \multicolumn{1}{c}{\textbf{10}} & \multicolumn{1}{c}{\textbf{20}} & \multicolumn{1}{c}{\textbf{30}} & \multicolumn{1}{c}{\textbf{40}} & \multicolumn{1}{c}{\textbf{50}} & \multicolumn{1}{c}{\textbf{60}} & \multicolumn{1}{c}{\textbf{70}} & \multicolumn{1}{c}{\textbf{80}} & \multicolumn{1}{c}{\textbf{90}} & \multicolumn{1}{c|}{\textbf{100}} & \multicolumn{1}{c}{\textbf{20}} & \multicolumn{1}{c}{\textbf{100}} & \multicolumn{1}{c}{\textbf{500}} & \multicolumn{1}{c}{\textbf{1000}} \\ \midrule
Crisis\textsubscript{Oltea} & 94.25                          & 94.97                          & 95.33                           & 95.49                           & 95.55                           & 95.66                           & 95.75                           & 95.81                           & 95.85                           & 95.92                           & 95.82                           & 95.92                            & 54.77                           & 90.26                            & 94.09                            & 94.89                             \\
\bottomrule
\end{tabular}
\caption{Full results of few-shot learning on DCL.}\label{tab:few_shot-dcl}
\end{table*}

\begin{table*}[h]
\centering
\tiny     
\begin{tabular}{@{}lrrrrrrrrrrrr|rrrr@{}}
\toprule
\multicolumn{1}{c}{}                         & \multicolumn{12}{c|}{\textbf{Percentage}}                                                                                                                                                                                                                                                                                                                                                                             & \multicolumn{4}{c}{\textbf{\# of Training Samples}}                                                                                       \\ \cmidrule(r){2-13} \cmidrule(r){14-17} 
\multicolumn{1}{c}{\textbf{}}                & \multicolumn{1}{c}{\textbf{1}} & \multicolumn{1}{c}{\textbf{5}} & \multicolumn{1}{c}{\textbf{10}} & \multicolumn{1}{c}{\textbf{20}} & \multicolumn{1}{c}{\textbf{30}} & \multicolumn{1}{c}{\textbf{40}} & \multicolumn{1}{c}{\textbf{50}} & \multicolumn{1}{c}{\textbf{60}} & \multicolumn{1}{c}{\textbf{70}} & \multicolumn{1}{c}{\textbf{80}} & \multicolumn{1}{c}{\textbf{90}} & \multicolumn{1}{c|}{\textbf{100}} & \multicolumn{1}{c}{\textbf{20}} & \multicolumn{1}{c}{\textbf{100}} & \multicolumn{1}{c}{\textbf{500}} & \multicolumn{1}{c}{\textbf{1000}} \\ \midrule
Crisis\textsubscript{Oltea} & 94.88 & 95.26 & 95.61 & 95.59 & 95.65 & 95.75 & 95.82 & 95.72 & 95.85 & 95.88 & 96.04 & 95.94 & 67.01     & 93.24    & 94.87    & 95.10    \\
Emo\textsubscript{Moham}    & 30.07 & 66.09 & 76.41 & 77.93 & 79.08 & 78.51 & 79.80 & 80.42 & 80.69 & 79.35 & 80.96 & 81.05 & 23.74     & 55.31    & 77.21    & 79.26    \\
Hate\textsubscript{Was}     & 33.12 & 53.06 & 54.15 & 54.85 & 55.84 & 56.30 & 56.65 & 56.73 & 56.90 & 57.10 & 57.30 & 57.13 & 33.88     & 38.36    & 52.63    & 54.14    \\
Hate\textsubscript{Dav}     & 62.43 & 72.62 & 74.77 & 74.38 & 75.47 & 76.11 & 77.32 & 77.53 & 77.19 & 77.08 & 77.96 & 77.15 & 33.47     & 42.19    & 68.00    & 70.83    \\
\bottomrule
\end{tabular}
\caption{Full results of few-shot learning on InfoDCL-RoBERTa.}\label{tab:few_shot-infodcl-roberta}
\end{table*}

\begin{table*}[h]
\centering
\tiny     
\begin{tabular}{@{}lrrrrrrrrrrrr|rrcc@{}}
\toprule
\multicolumn{1}{c}{}                         & \multicolumn{12}{c|}{\textbf{Percentage}}                                                                                                                                                                                                                                                                                                                                                                             & \multicolumn{4}{c}{\textbf{\# of Training Samples}}                                                                                       \\ \cmidrule(r){2-13} \cmidrule(r){14-17} 
\multicolumn{1}{c}{\textbf{}}                & \multicolumn{1}{c}{\textbf{1}} & \multicolumn{1}{c}{\textbf{5}} & \multicolumn{1}{c}{\textbf{10}} & \multicolumn{1}{c}{\textbf{20}} & \multicolumn{1}{c}{\textbf{30}} & \multicolumn{1}{c}{\textbf{40}} & \multicolumn{1}{c}{\textbf{50}} & \multicolumn{1}{c}{\textbf{60}} & \multicolumn{1}{c}{\textbf{70}} & \multicolumn{1}{c}{\textbf{80}} & \multicolumn{1}{c}{\textbf{90}} & \multicolumn{1}{c|}{\textbf{100}} & \multicolumn{1}{c}{\textbf{20}} & \multicolumn{1}{c}{\textbf{100}} & \multicolumn{1}{c}{\textbf{500}} & \multicolumn{1}{c}{\textbf{1000}} \\ \midrule
Crisis\textsubscript{Oltea} & 94.09 & 95.07 & 95.29 & 95.55 & 95.70 & 95.60 & 95.83 & 95.79 & 95.86 & 95.84 & 95.84 & 95.84 & 57.68     & 89.00    & 94.13    & 94.79    \\
Emo\textsubscript{Moham}    & 29.53 & 34.42 & 67.80 & 74.42 & 77.04 & 77.55 & 77.83 & 79.56 & 80.06 & 80.66 & 80.04 & 81.96 & 25.21     & 30.90    & 73.64    & 76.25    \\
Hate\textsubscript{Was}     & 31.12 & 52.01 & 53.92 & 54.92 & 55.82 & 55.86 & 56.38 & 56.95 & 56.48 & 57.11 & 56.94 & 57.65 & 33.14     & 31.69    & 52.52    & 53.62    \\
Hate\textsubscript{Dav}     & 32.42 & 69.28 & 74.02 & 75.12 & 76.59 & 76.15 & 76.77 & 77.05 & 77.23 & 77.40 & 77.77 & 77.94 & 32.84     & 31.47    & 60.75    & 68.86    \\
\bottomrule
\end{tabular}
\caption{Full results of few-shot learning on InfoDCL-BERTweet.}\label{tab:few_shot-infodcl-bertweet}
\end{table*}

\end{document}